\pdfoutput=1

\documentclass[10pt,twocolumn,letterpaper]{article}

\usepackage[pagenumbers]{cvpr}      %

\usepackage{cuted} %

\usepackage{currfile} %

\usepackage{caption} %

\usepackage{algorithm}
\usepackage{algpseudocode}
\usepackage{tabularx} %
\newcolumntype{Y}{>{\centering\arraybackslash}X}
\newcommand{\vE}{\emph{Explicit}}
\newcommand{\vI}{\emph{Implicit}}

\usepackage{longtable}
\usepackage{booktabs}
\usepackage{listings}
\usepackage{tcolorbox}
\newcommand{\methodname}{PreferenceScene}

\definecolor{cvprblue}{rgb}{0.21,0.49,0.74}
\usepackage[pagebackref,breaklinks,colorlinks,allcolors=cvprblue]{hyperref}

\title{PreferenceScene: Personalized Indoor Scene Synthesis from Feedback}%

\author{Seongrae Noh \quad RyunHa Lee \quad HyeongYeop Kang$^{\dagger}$\\[0.8em]
Korea University}

\providecommand{\mm}[3]{\begin{tabular}[t]{@{}c@{}}#1\\[-3pt]{\tiny(#2--#3)}\end{tabular}}
\providecommand{\hdtwo}[2]{\begin{tabular}[b]{@{}c@{}}#1\\#2\end{tabular}}
\newcommand{\trMainFbsHolodeckE}{44.0$\pm$0.2}

\newcommand{\trMainFbsHolodeckI}{41.4$\pm$0.6}

\newcommand{\trMainFbsLayoutgptE}{27.9$\pm$1.0}

\newcommand{\trMainFbsLayoutgptI}{25.2$\pm$1.3}

\newcommand{\trMainFbsIdesignE}{44.3$\pm$1.1}

\newcommand{\trMainFbsIdesignI}{41.0$\pm$0.5}

\newcommand{\trMainFbsLayoutvlmE}{45.1$\pm$0.3}

\newcommand{\trMainFbsLayoutvlmI}{43.1$\pm$1.4}

\newcommand{\trMainFbsRThreelE}{43.9$\pm$1.0}

\newcommand{\trMainFbsRThreelI}{40.8$\pm$0.6}

\newcommand{\trMainOursE}{56.6$\pm$0.3}

\newcommand{\trMainOursI}{49.6$\pm$1.7}

\newcommand{\trMainFbHolodeckE}{43.0$\pm$0.1}

\newcommand{\trMainFbHolodeckI}{40.6$\pm$1.0}

\newcommand{\trMainFbLayoutgptE}{27.1$\pm$0.6}

\newcommand{\trMainFbLayoutgptI}{26.6$\pm$1.6}

\newcommand{\trMainFbIdesignE}{45.6$\pm$0.3}

\newcommand{\trMainFbIdesignI}{41.7$\pm$1.0}

\newcommand{\trMainFbLayoutvlmE}{45.3$\pm$0.2}

\newcommand{\trMainFbLayoutvlmI}{43.1$\pm$0.8}

\newcommand{\trMainFbRThreelE}{43.9$\pm$0.8}

\newcommand{\trMainFbRThreelI}{41.2$\pm$0.3}

\newcommand{\trMainOsHolodeckE}{40.8$\pm$0.8}

\newcommand{\trMainOsHolodeckI}{34.2$\pm$1.1}

\newcommand{\trMainOsLayoutgptE}{25.8$\pm$0.8}

\newcommand{\trMainOsLayoutgptI}{24.7$\pm$1.2}

\newcommand{\trMainOsIdesignE}{40.4$\pm$1.3}

\newcommand{\trMainOsIdesignI}{32.9$\pm$0.4}

\newcommand{\trMainOsLayoutvlmE}{43.4$\pm$0.4}

\newcommand{\trMainOsLayoutvlmI}{35.1$\pm$0.4}

\newcommand{\trMainOsRThreelE}{41.2$\pm$0.5}

\newcommand{\trMainOsRThreelI}{33.5$\pm$0.5}

\newcommand{\trMainOssHolodeckE}{41.3$\pm$1.0}

\newcommand{\trMainOssHolodeckI}{33.2$\pm$0.5}

\newcommand{\trMainOssLayoutgptE}{26.5$\pm$0.9}

\newcommand{\trMainOssLayoutgptI}{23.9$\pm$2.3}

\newcommand{\trMainOssIdesignE}{39.9$\pm$1.3}

\newcommand{\trMainOssIdesignI}{31.3$\pm$0.4}

\newcommand{\trMainOssLayoutvlmE}{41.5$\pm$0.9}

\newcommand{\trMainOssLayoutvlmI}{33.7$\pm$0.9}

\newcommand{\trMainOssRThreelE}{41.3$\pm$0.7}

\newcommand{\trMainOssRThreelI}{33.3$\pm$0.6}

\newcommand{\trMainOrHolodeckE}{45.3}

\newcommand{\trMainOrHolodeckI}{44.0}

\newcommand{\trMainOrLayoutgptE}{25.8}

\newcommand{\trMainOrLayoutgptI}{25.4}

\newcommand{\trMainOrIdesignE}{43.8}

\newcommand{\trMainOrIdesignI}{40.3}

\newcommand{\trMainOrLayoutvlmE}{47.8}

\newcommand{\trMainOrLayoutvlmI}{47.5}

\newcommand{\trMainOrRThreelE}{45.5}

\newcommand{\trMainOrRThreelI}{45.2}

\newcommand{\trPairFbsHolodeckED}{+12.6}

\newcommand{\trPairFbsHolodeckID}{+8.2}

\newcommand{\trPairFbsLayoutgptED}{+26.4}

\newcommand{\trPairFbsLayoutgptID}{+23.0}

\newcommand{\trPairFbsIdesignED}{+12.3}

\newcommand{\trPairFbsIdesignID}{+8.5}

\newcommand{\trPairFbsLayoutvlmED}{+11.5}

\newcommand{\trPairFbsLayoutvlmID}{+6.4}

\newcommand{\trPairFbsRThreelED}{+12.6}

\newcommand{\trPairFbsRThreelID}{+8.8}

\newcommand{\trPairFbHolodeckED}{+13.6}

\newcommand{\trPairFbHolodeckID}{+9.0}

\newcommand{\trPairFbLayoutgptED}{+27.1}

\newcommand{\trPairFbLayoutgptID}{+21.6}

\newcommand{\trPairFbIdesignED}{+11.0}

\newcommand{\trPairFbIdesignID}{+7.8}

\newcommand{\trPairFbLayoutvlmED}{+11.3}

\newcommand{\trPairFbLayoutvlmID}{+6.4}

\newcommand{\trPairFbRThreelED}{+12.7}

\newcommand{\trPairFbRThreelID}{+8.4}

\newcommand{\trPairOsHolodeckED}{+15.8}

\newcommand{\trPairOsHolodeckID}{+15.4}

\newcommand{\trPairOsLayoutgptED}{+28.4}

\newcommand{\trPairOsLayoutgptID}{+23.5}

\newcommand{\trPairOsIdesignED}{+16.2}

\newcommand{\trPairOsIdesignID}{+16.7}

\newcommand{\trPairOsLayoutvlmED}{+13.2}

\newcommand{\trPairOsLayoutvlmID}{+14.5}

\newcommand{\trPairOsRThreelED}{+15.4}

\newcommand{\trPairOsRThreelID}{+16.1}

\newcommand{\trPairOssHolodeckED}{+15.3}

\newcommand{\trPairOssHolodeckID}{+16.4}

\newcommand{\trPairOssLayoutgptED}{+27.8}

\newcommand{\trPairOssLayoutgptID}{+24.4}

\newcommand{\trPairOssIdesignED}{+16.7}

\newcommand{\trPairOssIdesignID}{+18.2}

\newcommand{\trPairOssLayoutvlmED}{+15.1}

\newcommand{\trPairOssLayoutvlmID}{+15.9}

\newcommand{\trPairOssRThreelED}{+15.3}

\newcommand{\trPairOssRThreelID}{+16.3}

\newcommand{\trPairOrHolodeckED}{+11.1}

\newcommand{\trPairOrHolodeckID}{+7.0}

\newcommand{\trPairOrLayoutgptED}{+29.2}

\newcommand{\trPairOrLayoutgptID}{+24.8}

\newcommand{\trPairOrIdesignED}{+12.7}

\newcommand{\trPairOrIdesignID}{+10.7}

\newcommand{\trPairOrLayoutvlmED}{+8.7}

\newcommand{\trPairOrLayoutvlmID}{+3.5}

\newcommand{\trPairOrRThreelED}{+11.0}

\newcommand{\trPairOrRThreelID}{+5.8}

\newcommand{\trGainFbsLayoutgptI}{+1.4}

\newcommand{\trGainFbsIdesignI}{+9.7}

\newcommand{\trGainFbLayoutgptE}{+1.2}

\newcommand{\trGainFbIdesignE}{+5.2}

\newcommand{\trCostFbsHolodeckLlmcallsMm}{\mm{19.5}{14.0}{20.0}}

\newcommand{\trCostFbsHolodeckTurnsMm}{\mm{361}{133}{1653}}

\newcommand{\trCostFbsLayoutgptLlmcallsMm}{\mm{4.0}{4.0}{4.0}}

\newcommand{\trCostFbsLayoutgptTurnsMm}{\mm{155}{59}{578}}

\newcommand{\trCostFbsIdesignLlmcallsMm}{\mm{85.8}{24.0}{339.0}}

\newcommand{\trCostFbsIdesignTurnsMm}{\mm{823}{170}{4611}}

\newcommand{\trCostFbsLayoutvlmLlmcallsMm}{\mm{184.9}{103.0}{332.0}}

\newcommand{\trCostFbsLayoutvlmTurnsMm}{\mm{753}{257}{3729}}

\newcommand{\trCostFbsRThreelLlmcallsMm}{\mm{59.7}{24.0}{92.0}}

\newcommand{\trCostFbsRThreelTurnsMm}{\mm{695}{163}{1618}}

\newcommand{\trCostEditHolodeckLlmcallsMm}{\mm{13.9}{11.0}{15.0}}

\newcommand{\trCostEditHolodeckTurnsMm}{\mm{358}{86}{1829}}

\newcommand{\trCostOursLlmcallsMm}{\mm{6.0}{6.0}{8.0}}
\newcommand{\trCostOursJevrequests}{25}

\newcommand{\trCostOursUsd}{0.0090}

\newcommand{\trCostOursTurnsMm}{\mm{269}{5.7}{562}}

\newcommand{\trPaidPaidE}{59.3}
\newcommand{\trPaidOursE}{60.5}

\newcommand{\trPaidPaidI}{52.0}
\newcommand{\trPaidOursI}{48.3}

\newcommand{\trAblFullE}{67.1$\pm$1.3}

\newcommand{\trAblCTwoDROneE}{+0.3}

\newcommand{\trAblCThreebDROneE}{$-$0.9}
\newcommand{\trAblFullI}{56.5$\pm$2.9}

\newcommand{\trAblCTwoDROneI}{$-$0.4}

\newcommand{\trAblCThreebDROneI}{$-$0.2}

\newcommand{\trTimFbsHolodeckScenesMm}{\mm{102}{54}{134}}

\newcommand{\trTimFbsLayoutgptScenesMm}{\mm{12}{9.6}{18}}

\newcommand{\trTimFbsIdesignScenesMm}{\mm{340}{131}{639}}

\newcommand{\trTimFbsLayoutvlmScenesMm}{\mm{361}{112}{1588}}

\newcommand{\trTimFbsRThreelScenesMm}{\mm{332}{185}{510}}

\newcommand{\trTimEditHolodeckScenesMm}{\mm{79}{43}{98}}

\newcommand{\trTimOursScenesMm}{\mm{43}{34}{65}}

\newcommand{\hsN}{20}
\newcommand{\hsAgeMean}{27.9}
\newcommand{\hsAgeMin}{21}
\newcommand{\hsAgeMax}{65}

\newcommand{\hsPerSys}{120}
\newcommand{\hsWinAllQOne}{79.7}

\newcommand{\hsWinHolodeckQOne}{74.2}
\newcommand{\hsCIHolodeckQOne}{65.7--81.2}
\newcommand{\hsPHolodeckQOne}{$<$0.0001}
\newcommand{\hsWinLayoutgptQOne}{91.7}
\newcommand{\hsCILayoutgptQOne}{85.3--95.4}
\newcommand{\hsPLayoutgptQOne}{$<$0.0001}
\newcommand{\hsWinIdesignQOne}{84.2}
\newcommand{\hsCIIdesignQOne}{76.6--89.6}
\newcommand{\hsPIdesignQOne}{$<$0.0001}
\newcommand{\hsWinLayoutvlmQOne}{79.2}
\newcommand{\hsCILayoutvlmQOne}{71.1--85.5}
\newcommand{\hsPLayoutvlmQOne}{$<$0.0001}
\newcommand{\hsWinRThreelQOne}{75.0}
\newcommand{\hsCIRThreelQOne}{66.6--81.9}
\newcommand{\hsPRThreelQOne}{$<$0.0001}
\newcommand{\hsWinHolodeckEditQOne}{74.2}
\newcommand{\hsCIHolodeckEditQOne}{65.7--81.2}
\newcommand{\hsPHolodeckEditQOne}{$<$0.0001}
\newcommand{\hsWinAllQTwo}{80.8}

\newcommand{\hsWinHolodeckQTwo}{73.3}
\newcommand{\hsCIHolodeckQTwo}{64.8--80.4}
\newcommand{\hsPHolodeckQTwo}{$<$0.0001}
\newcommand{\hsWinLayoutgptQTwo}{91.7}
\newcommand{\hsCILayoutgptQTwo}{85.3--95.4}
\newcommand{\hsPLayoutgptQTwo}{$<$0.0001}
\newcommand{\hsWinIdesignQTwo}{85.0}
\newcommand{\hsCIIdesignQTwo}{77.5--90.3}
\newcommand{\hsPIdesignQTwo}{$<$0.0001}
\newcommand{\hsWinLayoutvlmQTwo}{87.5}
\newcommand{\hsCILayoutvlmQTwo}{80.4--92.3}
\newcommand{\hsPLayoutvlmQTwo}{$<$0.0001}
\newcommand{\hsWinRThreelQTwo}{79.2}
\newcommand{\hsCIRThreelQTwo}{71.1--85.5}
\newcommand{\hsPRThreelQTwo}{$<$0.0001}
\newcommand{\hsWinHolodeckEditQTwo}{68.3}
\newcommand{\hsCIHolodeckEditQTwo}{59.6--76.0}
\newcommand{\hsPHolodeckEditQTwo}{$<$0.0001}

\newcommand{\hsTauReal}{0.17}
\newcommand{\hsTauUUReal}{0.53}
\newcommand{\hsTauULReal}{0.17}

\newcommand{\hsTauFunc}{0.33}
\newcommand{\hsTauUUFunc}{0.42}
\newcommand{\hsTauULFunc}{0.30}

\newcommand{\hsTauInstr}{0.36}
\newcommand{\hsTauUUInstr}{0.48}
\newcommand{\hsTauULInstr}{0.33}

\newcommand{\hsRatedScenes}{40}

\providecommand{\flag}[1]{\textcolor{red}{#1}}%

\newcommand{\sxJevTokMax}{455}
\newcommand{\sxJevTokMean}{213}
\newcommand{\sxJevTokMin}{40}

\newcommand{\anAblArmfcAllArmE}{63.5$\pm$2.5}  %
\newcommand{\anAblArmfcAllDE}{$-$3.6}  %
\newcommand{\anAblArmfcAllDstdE}{2.2}  %
\newcommand{\anAblArmfcAllCiE}{[$-$6.2, $-$1.1]}  %
\newcommand{\anAblArmfcTmissedDE}{$-$4.7}  %
\newcommand{\anAblArmfcTmissedDstdE}{1.2}  %
\newcommand{\anAblArmfcTmissedCiE}{[$-$9.2, $-$0.8]}  %
\newcommand{\anAblArmfcAllArmI}{49.5$\pm$3.0}  %
\newcommand{\anAblArmfcAllDI}{$-$7.0}  %
\newcommand{\anAblArmfcAllDstdI}{3.7}  %
\newcommand{\anAblArmfcAllCiI}{[$-$10.0, $-$4.1]}  %
\newcommand{\anAblArmfcTmissedDI}{$-$8.2}  %
\newcommand{\anAblArmfcTmissedDstdI}{5.8}  %
\newcommand{\anAblArmfcTmissedCiI}{[$-$12.1, $-$4.7]}  %
\newcommand{\anAblNoddAllArmE}{66.2$\pm$3.7}  %
\newcommand{\anAblNoddAllDE}{$-$0.9}  %
\newcommand{\anAblNoddAllDstdE}{3.7}  %
\newcommand{\anAblNoddAllCiE}{[$-$2.4, +0.6]}  %
\newcommand{\anAblNoddTmissedDE}{$-$1.7}  %
\newcommand{\anAblNoddTmissedDstdE}{4.4}  %
\newcommand{\anAblNoddTmissedCiE}{[$-$4.6, +1.3]}  %
\newcommand{\anAblNoddAllArmI}{55.5$\pm$1.7}  %
\newcommand{\anAblNoddAllDI}{$-$1.0}  %
\newcommand{\anAblNoddAllDstdI}{4.0}  %
\newcommand{\anAblNoddAllCiI}{[$-$3.4, +1.6]}  %
\newcommand{\anAblNoddTmissedDI}{$-$1.1}  %
\newcommand{\anAblNoddTmissedDstdI}{5.6}  %
\newcommand{\anAblNoddTmissedCiI}{[$-$4.3, +2.3]}  %
\newcommand{\anAblNodrwAllArmE}{67.2$\pm$1.0}  %
\newcommand{\anAblNodrwAllDE}{+0.1}  %
\newcommand{\anAblNodrwAllDstdE}{2.2}  %
\newcommand{\anAblNodrwAllCiE}{[$-$2.4, +2.7]}  %
\newcommand{\anAblNodrwTmissedDE}{+2.3}  %
\newcommand{\anAblNodrwTmissedDstdE}{2.5}  %
\newcommand{\anAblNodrwTmissedCiE}{[$-$1.3, +5.8]}  %
\newcommand{\anAblNodrwAllArmI}{57.0$\pm$2.2}  %
\newcommand{\anAblNodrwAllDI}{+0.5}  %
\newcommand{\anAblNodrwAllDstdI}{3.1}  %
\newcommand{\anAblNodrwAllCiI}{[$-$2.3, +3.2]}  %
\newcommand{\anAblNodrwTmissedDI}{+0.7}  %
\newcommand{\anAblNodrwTmissedDstdI}{5.1}  %
\newcommand{\anAblNodrwTmissedCiI}{[$-$3.0, +4.5]}  %
\newcommand{\eTwelveCHolodeckOursE}{58.5$\pm$0.5}
\newcommand{\eTwelveCHolodeckBaseE}{48.5$\pm$0.2}
\newcommand{\eTwelveCHolodeckDE}{+10.1}
\newcommand{\eTwelveCHolodeckDstdE}{0.4}
\newcommand{\eTwelveCHolodeckCiE}{[+7.6, +12.6]}

\newcommand{\eTwelveCIdesignBaseE}{49.2$\pm$1.1}
\newcommand{\eTwelveCIdesignDE}{+9.3}
\newcommand{\eTwelveCIdesignDstdE}{1.6}
\newcommand{\eTwelveCIdesignCiE}{[+6.3, +12.4]}

\newcommand{\eTwelveCLayoutvlmBaseE}{48.5$\pm$0.2}
\newcommand{\eTwelveCLayoutvlmDE}{+10.0}
\newcommand{\eTwelveCLayoutvlmDstdE}{0.7}
\newcommand{\eTwelveCLayoutvlmCiE}{[+7.2, +12.9]}

\newcommand{\eTwelveCRThreelBaseE}{49.5$\pm$1.1}
\newcommand{\eTwelveCRThreelDE}{+9.0}
\newcommand{\eTwelveCRThreelDstdE}{1.5}
\newcommand{\eTwelveCRThreelCiE}{[+6.2, +11.9]}

\newcommand{\eTwelveCEditHolodeckBaseE}{50.4$\pm$1.1}
\newcommand{\eTwelveCEditHolodeckDE}{+8.2}
\newcommand{\eTwelveCEditHolodeckDstdE}{1.3}
\newcommand{\eTwelveCEditHolodeckCiE}{[+5.6, +10.8]}
\newcommand{\eTwelveCRngLoE}{+8.2}
\newcommand{\eTwelveCRngHiE}{+10.1}
\newcommand{\eTwelveCHolodeckOursI}{51.7$\pm$1.8}
\newcommand{\eTwelveCHolodeckBaseI}{45.9$\pm$0.7}
\newcommand{\eTwelveCHolodeckDI}{+5.8}
\newcommand{\eTwelveCHolodeckDstdI}{2.3}
\newcommand{\eTwelveCHolodeckCiI}{[+3.2, +8.4]}

\newcommand{\eTwelveCIdesignBaseI}{45.7$\pm$0.6}
\newcommand{\eTwelveCIdesignDI}{+6.0}
\newcommand{\eTwelveCIdesignDstdI}{2.0}
\newcommand{\eTwelveCIdesignCiI}{[+2.9, +9.2]}

\newcommand{\eTwelveCLayoutvlmBaseI}{46.5$\pm$1.4}
\newcommand{\eTwelveCLayoutvlmDI}{+5.2}
\newcommand{\eTwelveCLayoutvlmDstdI}{2.0}
\newcommand{\eTwelveCLayoutvlmCiI}{[+2.5, +8.0]}

\newcommand{\eTwelveCRThreelBaseI}{45.5$\pm$0.6}
\newcommand{\eTwelveCRThreelDI}{+6.2}
\newcommand{\eTwelveCRThreelDstdI}{1.8}
\newcommand{\eTwelveCRThreelCiI}{[+3.4, +9.1]}

\newcommand{\eTwelveCEditHolodeckBaseI}{46.8$\pm$1.6}
\newcommand{\eTwelveCEditHolodeckDI}{+4.9}
\newcommand{\eTwelveCEditHolodeckDstdI}{0.7}
\newcommand{\eTwelveCEditHolodeckCiI}{[+2.3, +7.4]}
\newcommand{\eTwelveCRngLoI}{+4.9}
\newcommand{\eTwelveCRngHiI}{+6.2}
\newcommand{\eTwelveCNa}{76}
\newcommand{\eTwelveCNexcl}{24}

\newcommand{\eTwelveCNwin}{7}
\newcommand{\eTwelveCNrthree}{6}
\newcommand{\eTwelveCNinst}{120}

\newcommand{\eTwelveAllNa}{100}  %
\newcommand{\suppBootB}{10{,}000}  %
\newcommand{\eElevenAllHolodeckOrigE}{44.0$\pm$0.2}  %
\newcommand{\eElevenAllHolodeckOwnE}{50.6$\pm$0.7}  %
\newcommand{\eElevenAllHolodeckMethE}{58.1$\pm$0.5}  %
\newcommand{\eElevenAllHolodeckDOoE}{+6.6}  %
\newcommand{\eElevenAllHolodeckDOoStdE}{0.7}  %
\newcommand{\eElevenAllHolodeckCiOoE}{[+5.2, +7.9]}  %
\newcommand{\eElevenAllHolodeckDMoE}{+7.5}  %
\newcommand{\eElevenAllHolodeckDMoStdE}{0.9}  %
\newcommand{\eElevenAllHolodeckCiMoE}{[+5.5, +9.4]}  %
\newcommand{\eElevenAddHolodeckOrigE}{29.9$\pm$0.2}  %
\newcommand{\eElevenAddHolodeckOwnE}{46.9$\pm$1.1}  %
\newcommand{\eElevenAddHolodeckMethE}{58.4$\pm$1.5}  %
\newcommand{\eElevenAddHolodeckDOoE}{+17.0}  %
\newcommand{\eElevenAddHolodeckDOoStdE}{1.0}  %
\newcommand{\eElevenAddHolodeckCiOoE}{[+13.9, +20.2]}  %
\newcommand{\eElevenAddHolodeckDMoE}{+11.5}  %
\newcommand{\eElevenAddHolodeckDMoStdE}{1.3}  %
\newcommand{\eElevenAddHolodeckCiMoE}{[+8.1, +14.9]}  %
\newcommand{\eElevenAllIdesignOrigE}{44.3$\pm$1.1}  %
\newcommand{\eElevenAllIdesignOwnE}{50.9$\pm$1.6}  %
\newcommand{\eElevenAllIdesignMethE}{58.1$\pm$0.5}  %
\newcommand{\eElevenAllIdesignDOoE}{+6.6}  %
\newcommand{\eElevenAllIdesignDOoStdE}{0.7}  %
\newcommand{\eElevenAllIdesignCiOoE}{[+5.6, +7.7]}  %
\newcommand{\eElevenAllIdesignDMoE}{+7.1}  %
\newcommand{\eElevenAllIdesignDMoStdE}{1.1}  %
\newcommand{\eElevenAllIdesignCiMoE}{[+4.8, +9.6]}  %
\newcommand{\eElevenAddIdesignOrigE}{28.6$\pm$1.0}  %
\newcommand{\eElevenAddIdesignOwnE}{45.0$\pm$2.1}  %
\newcommand{\eElevenAddIdesignMethE}{58.4$\pm$1.5}  %
\newcommand{\eElevenAddIdesignDOoE}{+16.4}  %
\newcommand{\eElevenAddIdesignDOoStdE}{1.1}  %
\newcommand{\eElevenAddIdesignCiOoE}{[+13.7, +19.1]}  %
\newcommand{\eElevenAddIdesignDMoE}{+13.3}  %
\newcommand{\eElevenAddIdesignDMoStdE}{2.7}  %
\newcommand{\eElevenAddIdesignCiMoE}{[+9.8, +16.9]}  %
\newcommand{\eElevenAllLayoutvlmOrigE}{45.1$\pm$0.3}  %
\newcommand{\eElevenAllLayoutvlmOwnE}{51.2$\pm$1.1}  %
\newcommand{\eElevenAllLayoutvlmMethE}{58.1$\pm$0.5}  %
\newcommand{\eElevenAllLayoutvlmDOoE}{+6.0}  %
\newcommand{\eElevenAllLayoutvlmDOoStdE}{1.0}  %
\newcommand{\eElevenAllLayoutvlmCiOoE}{[+4.8, +7.3]}  %
\newcommand{\eElevenAllLayoutvlmDMoE}{+6.9}  %
\newcommand{\eElevenAllLayoutvlmDMoStdE}{1.4}  %
\newcommand{\eElevenAllLayoutvlmCiMoE}{[+4.8, +9.0]}  %
\newcommand{\eElevenAddLayoutvlmOrigE}{31.1$\pm$0.8}  %
\newcommand{\eElevenAddLayoutvlmOwnE}{47.3$\pm$0.8}  %
\newcommand{\eElevenAddLayoutvlmMethE}{58.4$\pm$1.5}  %
\newcommand{\eElevenAddLayoutvlmDOoE}{+16.2}  %
\newcommand{\eElevenAddLayoutvlmDOoStdE}{1.6}  %
\newcommand{\eElevenAddLayoutvlmCiOoE}{[+13.2, +19.0]}  %
\newcommand{\eElevenAddLayoutvlmDMoE}{+11.1}  %
\newcommand{\eElevenAddLayoutvlmDMoStdE}{2.1}  %
\newcommand{\eElevenAddLayoutvlmCiMoE}{[+7.9, +14.2]}  %
\newcommand{\eElevenAllRThreelOrigE}{43.9$\pm$1.0}  %
\newcommand{\eElevenAllRThreelOwnE}{51.2$\pm$0.6}  %
\newcommand{\eElevenAllRThreelMethE}{58.1$\pm$0.5}  %
\newcommand{\eElevenAllRThreelDOoE}{+7.3}  %
\newcommand{\eElevenAllRThreelDOoStdE}{0.9}  %
\newcommand{\eElevenAllRThreelCiOoE}{[+6.1, +8.4]}  %
\newcommand{\eElevenAllRThreelDMoE}{+6.9}  %
\newcommand{\eElevenAllRThreelDMoStdE}{0.4}  %
\newcommand{\eElevenAllRThreelCiMoE}{[+4.6, +9.1]}  %
\newcommand{\eElevenAddRThreelOrigE}{28.8$\pm$1.3}  %
\newcommand{\eElevenAddRThreelOwnE}{44.2$\pm$0.6}  %
\newcommand{\eElevenAddRThreelMethE}{58.2$\pm$1.2}  %
\newcommand{\eElevenAddRThreelDOoE}{+15.4}  %
\newcommand{\eElevenAddRThreelDOoStdE}{1.0}  %
\newcommand{\eElevenAddRThreelCiOoE}{[+13.0, +17.7]}  %
\newcommand{\eElevenAddRThreelDMoE}{+14.0}  %
\newcommand{\eElevenAddRThreelDMoStdE}{1.2}  %
\newcommand{\eElevenAddRThreelCiMoE}{[+10.8, +17.3]}  %
\newcommand{\eElevenAllEditHolodeckOrigE}{45.6$\pm$0.7}  %
\newcommand{\eElevenAllEditHolodeckOwnE}{52.5$\pm$1.1}  %
\newcommand{\eElevenAllEditHolodeckMethE}{58.1$\pm$0.5}  %
\newcommand{\eElevenAllEditHolodeckDOoE}{+6.9}  %
\newcommand{\eElevenAllEditHolodeckDOoStdE}{0.5}  %
\newcommand{\eElevenAllEditHolodeckCiOoE}{[+5.7, +8.2]}  %
\newcommand{\eElevenAllEditHolodeckDMoE}{+5.6}  %
\newcommand{\eElevenAllEditHolodeckDMoStdE}{1.3}  %
\newcommand{\eElevenAllEditHolodeckCiMoE}{[+3.6, +7.6]}  %
\newcommand{\eElevenAddEditHolodeckOrigE}{32.8$\pm$0.6}  %
\newcommand{\eElevenAddEditHolodeckOwnE}{50.8$\pm$1.4}  %
\newcommand{\eElevenAddEditHolodeckMethE}{58.4$\pm$1.5}  %
\newcommand{\eElevenAddEditHolodeckDOoE}{+18.0}  %
\newcommand{\eElevenAddEditHolodeckDOoStdE}{0.9}  %
\newcommand{\eElevenAddEditHolodeckCiOoE}{[+15.2, +20.9]}  %
\newcommand{\eElevenAddEditHolodeckDMoE}{+7.6}  %
\newcommand{\eElevenAddEditHolodeckDMoStdE}{0.9}  %
\newcommand{\eElevenAddEditHolodeckCiMoE}{[+4.6, +10.5]}  %
\newcommand{\eElevenRngAllDOoLoE}{+6.0}  %
\newcommand{\eElevenRngAllDOoHiE}{+7.3}  %
\newcommand{\eElevenRngAllDMoLoE}{+5.6}  %
\newcommand{\eElevenRngAllDMoHiE}{+7.5}  %
\newcommand{\eElevenRngAddDOoLoE}{+15.4}  %
\newcommand{\eElevenRngAddDOoHiE}{+18.0}  %
\newcommand{\eElevenRngAddDMoLoE}{+7.6}  %
\newcommand{\eElevenRngAddDMoHiE}{+14.0}  %
\newcommand{\eElevenRngRestDOoLoE}{$-$0.2}  %
\newcommand{\eElevenRngRestDOoHiE}{+0.9}  %
\newcommand{\eElevenAllHolodeckOrigI}{41.4$\pm$0.6}  %
\newcommand{\eElevenAllHolodeckOwnI}{46.0$\pm$0.7}  %
\newcommand{\eElevenAllHolodeckMethI}{49.7$\pm$1.9}  %
\newcommand{\eElevenAllHolodeckDOoI}{+4.6}  %
\newcommand{\eElevenAllHolodeckDOoStdI}{0.7}  %
\newcommand{\eElevenAllHolodeckCiOoI}{[+3.5, +5.8]}  %
\newcommand{\eElevenAllHolodeckDMoI}{+3.7}  %
\newcommand{\eElevenAllHolodeckDMoStdI}{2.3}  %
\newcommand{\eElevenAllHolodeckCiMoI}{[+1.7, +5.8]}  %
\newcommand{\eElevenAddHolodeckOrigI}{30.2$\pm$1.3}  %
\newcommand{\eElevenAddHolodeckOwnI}{42.8$\pm$0.6}  %
\newcommand{\eElevenAddHolodeckMethI}{49.0$\pm$2.0}  %
\newcommand{\eElevenAddHolodeckDOoI}{+12.6}  %
\newcommand{\eElevenAddHolodeckDOoStdI}{1.6}  %
\newcommand{\eElevenAddHolodeckCiOoI}{[+10.1, +15.1]}  %
\newcommand{\eElevenAddHolodeckDMoI}{+6.2}  %
\newcommand{\eElevenAddHolodeckDMoStdI}{2.3}  %
\newcommand{\eElevenAddHolodeckCiMoI}{[+2.8, +9.6]}  %
\newcommand{\eElevenAllIdesignOrigI}{41.0$\pm$0.5}  %
\newcommand{\eElevenAllIdesignOwnI}{45.4$\pm$0.6}  %
\newcommand{\eElevenAllIdesignMethI}{49.7$\pm$1.9}  %
\newcommand{\eElevenAllIdesignDOoI}{+4.3}  %
\newcommand{\eElevenAllIdesignDOoStdI}{0.2}  %
\newcommand{\eElevenAllIdesignCiOoI}{[+3.4, +5.3]}  %
\newcommand{\eElevenAllIdesignDMoI}{+4.4}  %
\newcommand{\eElevenAllIdesignDMoStdI}{2.0}  %
\newcommand{\eElevenAllIdesignCiMoI}{[+1.8, +6.9]}  %
\newcommand{\eElevenAddIdesignOrigI}{29.8$\pm$0.9}  %
\newcommand{\eElevenAddIdesignOwnI}{40.6$\pm$0.5}  %
\newcommand{\eElevenAddIdesignMethI}{49.0$\pm$2.0}  %
\newcommand{\eElevenAddIdesignDOoI}{+10.8}  %
\newcommand{\eElevenAddIdesignDOoStdI}{0.7}  %
\newcommand{\eElevenAddIdesignCiOoI}{[+8.6, +12.9]}  %
\newcommand{\eElevenAddIdesignDMoI}{+8.5}  %
\newcommand{\eElevenAddIdesignDMoStdI}{1.6}  %
\newcommand{\eElevenAddIdesignCiMoI}{[+5.1, +11.9]}  %
\newcommand{\eElevenAllLayoutvlmOrigI}{43.1$\pm$1.4}  %
\newcommand{\eElevenAllLayoutvlmOwnI}{47.1$\pm$1.1}  %
\newcommand{\eElevenAllLayoutvlmMethI}{49.7$\pm$1.9}  %
\newcommand{\eElevenAllLayoutvlmDOoI}{+3.9}  %
\newcommand{\eElevenAllLayoutvlmDOoStdI}{0.4}  %
\newcommand{\eElevenAllLayoutvlmCiOoI}{[+3.0, +4.8]}  %
\newcommand{\eElevenAllLayoutvlmDMoI}{+2.7}  %
\newcommand{\eElevenAllLayoutvlmDMoStdI}{2.3}  %
\newcommand{\eElevenAllLayoutvlmCiMoI}{[+0.4, +4.9]}  %
\newcommand{\eElevenAddLayoutvlmOrigI}{32.2$\pm$2.6}  %
\newcommand{\eElevenAddLayoutvlmOwnI}{42.7$\pm$1.7}  %
\newcommand{\eElevenAddLayoutvlmMethI}{49.0$\pm$2.0}  %
\newcommand{\eElevenAddLayoutvlmDOoI}{+10.5}  %
\newcommand{\eElevenAddLayoutvlmDOoStdI}{1.5}  %
\newcommand{\eElevenAddLayoutvlmCiOoI}{[+8.3, +12.7]}  %
\newcommand{\eElevenAddLayoutvlmDMoI}{+6.3}  %
\newcommand{\eElevenAddLayoutvlmDMoStdI}{3.6}  %
\newcommand{\eElevenAddLayoutvlmCiMoI}{[+3.0, +9.6]}  %
\newcommand{\eElevenAllRThreelOrigI}{40.8$\pm$0.6}  %
\newcommand{\eElevenAllRThreelOwnI}{45.1$\pm$0.9}  %
\newcommand{\eElevenAllRThreelMethI}{49.7$\pm$1.9}  %
\newcommand{\eElevenAllRThreelDOoI}{+4.3}  %
\newcommand{\eElevenAllRThreelDOoStdI}{1.0}  %
\newcommand{\eElevenAllRThreelCiOoI}{[+3.2, +5.3]}  %
\newcommand{\eElevenAllRThreelDMoI}{+4.7}  %
\newcommand{\eElevenAllRThreelDMoStdI}{2.5}  %
\newcommand{\eElevenAllRThreelCiMoI}{[+2.4, +6.9]}  %
\newcommand{\eElevenAddRThreelOrigI}{29.5$\pm$0.8}  %
\newcommand{\eElevenAddRThreelOwnI}{39.4$\pm$1.9}  %
\newcommand{\eElevenAddRThreelMethI}{48.7$\pm$2.2}  %
\newcommand{\eElevenAddRThreelDOoI}{+9.9}  %
\newcommand{\eElevenAddRThreelDOoStdI}{2.1}  %
\newcommand{\eElevenAddRThreelCiOoI}{[+7.8, +12.1]}  %
\newcommand{\eElevenAddRThreelDMoI}{+9.4}  %
\newcommand{\eElevenAddRThreelDMoStdI}{3.3}  %
\newcommand{\eElevenAddRThreelCiMoI}{[+6.1, +12.6]}  %
\newcommand{\eElevenAllEditHolodeckOrigI}{42.6$\pm$1.4}  %
\newcommand{\eElevenAllEditHolodeckOwnI}{46.7$\pm$1.7}  %
\newcommand{\eElevenAllEditHolodeckMethI}{49.7$\pm$1.9}  %
\newcommand{\eElevenAllEditHolodeckDOoI}{+4.1}  %
\newcommand{\eElevenAllEditHolodeckDOoStdI}{0.5}  %
\newcommand{\eElevenAllEditHolodeckCiOoI}{[+3.1, +5.2]}  %
\newcommand{\eElevenAllEditHolodeckDMoI}{+3.0}  %
\newcommand{\eElevenAllEditHolodeckDMoStdI}{1.7}  %
\newcommand{\eElevenAllEditHolodeckCiMoI}{[+0.9, +5.2]}  %
\newcommand{\eElevenAddEditHolodeckOrigI}{32.8$\pm$1.0}  %
\newcommand{\eElevenAddEditHolodeckOwnI}{44.5$\pm$0.9}  %
\newcommand{\eElevenAddEditHolodeckMethI}{49.0$\pm$2.0}  %
\newcommand{\eElevenAddEditHolodeckDOoI}{+11.7}  %
\newcommand{\eElevenAddEditHolodeckDOoStdI}{0.9}  %
\newcommand{\eElevenAddEditHolodeckCiOoI}{[+9.5, +14.1]}  %
\newcommand{\eElevenAddEditHolodeckDMoI}{+4.6}  %
\newcommand{\eElevenAddEditHolodeckDMoStdI}{1.7}  %
\newcommand{\eElevenAddEditHolodeckCiMoI}{[+1.0, +8.1]}  %
\newcommand{\eElevenRngAllDOoLoI}{+3.9}  %
\newcommand{\eElevenRngAllDOoHiI}{+4.6}  %
\newcommand{\eElevenRngAllDMoLoI}{+2.7}  %
\newcommand{\eElevenRngAllDMoHiI}{+4.7}  %
\newcommand{\eElevenRngAddDOoLoI}{+9.9}  %
\newcommand{\eElevenRngAddDOoHiI}{+12.6}  %
\newcommand{\eElevenRngAddDMoLoI}{+4.6}  %
\newcommand{\eElevenRngAddDMoHiI}{+9.4}  %
\newcommand{\eElevenRngRestDOoLoI}{$-$0.6}  %
\newcommand{\eElevenRngRestDOoHiI}{+0.4}  %
\newcommand{\eElevenNinst}{120}  %

\newcommand{\eOneGainOursI}{+12.1}

\newcommand{\eOneDidLoI}{+2.4}
\newcommand{\eOneDidHiI}{+4.6}

\newcommand{\eFourteenQuarterTurnTwoAllArmE}{64.6$\pm$2.6}  %
\newcommand{\eFourteenQuarterTurnTwoAllDE}{$-$0.3}  %
\newcommand{\eFourteenQuarterTurnTwoAllDStdE}{3.3}  %
\newcommand{\eFourteenQuarterTurnTwoAllCiE}{[$-$2.2, +1.5]}  %
\newcommand{\eFourteenQuarterTurnTwoWrongArmE}{22.9$\pm$25.3}  %
\newcommand{\eFourteenQuarterTurnTwoWrongDE}{$-$59.0}  %
\newcommand{\eFourteenQuarterTurnTwoWrongDStdE}{21.3}  %
\newcommand{\eFourteenQuarterTurnTwoWrongCiE}{[$-$83.3, $-$33.3]}  %
\newcommand{\eFourteenQuarterTurnThreeAllArmE}{63.7$\pm$2.7}  %
\newcommand{\eFourteenQuarterTurnThreeAllDE}{$-$2.2}  %
\newcommand{\eFourteenQuarterTurnThreeAllDStdE}{3.4}  %
\newcommand{\eFourteenQuarterTurnThreeAllCiE}{[$-$4.2, $-$0.1]}  %
\newcommand{\eFourteenQuarterTurnThreeWrongArmE}{25.0$\pm$16.9}  %
\newcommand{\eFourteenQuarterTurnThreeWrongDE}{$-$49.9}  %
\newcommand{\eFourteenQuarterTurnThreeWrongDStdE}{27.2}  %
\newcommand{\eFourteenQuarterTurnThreeWrongCiE}{[$-$68.1, $-$29.2]}  %
\newcommand{\eFourteenQuarterFinalAllArmE}{65.3$\pm$3.1}  %
\newcommand{\eFourteenQuarterFinalAllDE}{$-$2.0}  %
\newcommand{\eFourteenQuarterFinalAllDStdE}{2.9}  %
\newcommand{\eFourteenQuarterFinalAllCiE}{[$-$4.5, +0.5]}  %
\newcommand{\eFourteenQuarterFinalWrongArmE}{48.5$\pm$11.7}  %
\newcommand{\eFourteenQuarterFinalWrongDE}{$-$19.1}  %
\newcommand{\eFourteenQuarterFinalWrongDStdE}{3.3}  %
\newcommand{\eFourteenQuarterFinalWrongCiE}{[$-$27.8, $-$9.7]}  %
\newcommand{\eFourteenQuarterTurnTwoAllArmI}{48.0$\pm$2.0}  %
\newcommand{\eFourteenQuarterTurnTwoAllDI}{0.0}  %
\newcommand{\eFourteenQuarterTurnTwoAllDStdI}{1.0}  %
\newcommand{\eFourteenQuarterTurnTwoAllCiI}{[$-$3.5, +3.1]}  %
\newcommand{\eFourteenQuarterTurnTwoWrongArmI}{18.1$\pm$6.4}  %
\newcommand{\eFourteenQuarterTurnTwoWrongDI}{$-$32.6}  %
\newcommand{\eFourteenQuarterTurnTwoWrongDStdI}{33.4}  %
\newcommand{\eFourteenQuarterTurnTwoWrongCiI}{[$-$53.9, $-$7.6]}  %
\newcommand{\eFourteenQuarterTurnThreeAllArmI}{51.9$\pm$1.3}  %
\newcommand{\eFourteenQuarterTurnThreeAllDI}{$-$2.6}  %
\newcommand{\eFourteenQuarterTurnThreeAllDStdI}{4.2}  %
\newcommand{\eFourteenQuarterTurnThreeAllCiI}{[$-$5.8, +0.6]}  %
\newcommand{\eFourteenQuarterTurnThreeWrongArmI}{27.8$\pm$12.7}  %
\newcommand{\eFourteenQuarterTurnThreeWrongDI}{$-$28.2}  %
\newcommand{\eFourteenQuarterTurnThreeWrongDStdI}{28.6}  %
\newcommand{\eFourteenQuarterTurnThreeWrongCiI}{[$-$42.8, $-$15.3]}  %
\newcommand{\eFourteenQuarterFinalAllArmI}{53.1$\pm$0.7}  %
\newcommand{\eFourteenQuarterFinalAllDI}{$-$2.4}  %
\newcommand{\eFourteenQuarterFinalAllDStdI}{3.1}  %
\newcommand{\eFourteenQuarterFinalAllCiI}{[$-$6.3, +1.2]}  %
\newcommand{\eFourteenQuarterFinalWrongArmI}{40.7$\pm$11.7}  %
\newcommand{\eFourteenQuarterFinalWrongDI}{$-$19.9}  %
\newcommand{\eFourteenQuarterFinalWrongDStdI}{14.3}  %
\newcommand{\eFourteenQuarterFinalWrongCiI}{[$-$33.4, $-$3.4]}  %
\newcommand{\eFourteenHalfTurnTwoAllArmE}{62.3$\pm$2.8}  %
\newcommand{\eFourteenHalfTurnTwoAllDE}{$-$2.6}  %
\newcommand{\eFourteenHalfTurnTwoAllDStdE}{3.0}  %
\newcommand{\eFourteenHalfTurnTwoAllCiE}{[$-$4.7, $-$0.6]}  %
\newcommand{\eFourteenHalfTurnTwoWrongArmE}{23.8$\pm$3.4}  %
\newcommand{\eFourteenHalfTurnTwoWrongDE}{$-$51.8}  %
\newcommand{\eFourteenHalfTurnTwoWrongDStdE}{13.8}  %
\newcommand{\eFourteenHalfTurnTwoWrongCiE}{[$-$66.6, $-$33.8]}  %
\newcommand{\eFourteenHalfTurnThreeAllArmE}{62.3$\pm$2.6}  %
\newcommand{\eFourteenHalfTurnThreeAllDE}{$-$3.6}  %
\newcommand{\eFourteenHalfTurnThreeAllDStdE}{3.6}  %
\newcommand{\eFourteenHalfTurnThreeAllCiE}{[$-$6.7, $-$0.6]}  %
\newcommand{\eFourteenHalfTurnThreeWrongArmE}{36.0$\pm$8.3}  %
\newcommand{\eFourteenHalfTurnThreeWrongDE}{$-$34.8}  %
\newcommand{\eFourteenHalfTurnThreeWrongDStdE}{10.5}  %
\newcommand{\eFourteenHalfTurnThreeWrongCiE}{[$-$45.2, $-$24.5]}  %
\newcommand{\eFourteenHalfFinalAllArmE}{63.1$\pm$1.5}  %
\newcommand{\eFourteenHalfFinalAllDE}{$-$4.2}  %
\newcommand{\eFourteenHalfFinalAllDStdE}{0.3}  %
\newcommand{\eFourteenHalfFinalAllCiE}{[$-$7.3, $-$1.3]}  %
\newcommand{\eFourteenHalfFinalWrongArmE}{49.4$\pm$14.6}  %
\newcommand{\eFourteenHalfFinalWrongDE}{$-$21.9}  %
\newcommand{\eFourteenHalfFinalWrongDStdE}{7.8}  %
\newcommand{\eFourteenHalfFinalWrongCiE}{[$-$30.7, $-$12.8]}  %
\newcommand{\eFourteenHalfTurnTwoAllArmI}{48.1$\pm$1.4}  %
\newcommand{\eFourteenHalfTurnTwoAllDI}{+0.1}  %
\newcommand{\eFourteenHalfTurnTwoAllDStdI}{2.1}  %
\newcommand{\eFourteenHalfTurnTwoAllCiI}{[$-$2.5, +2.8]}  %
\newcommand{\eFourteenHalfTurnTwoWrongArmI}{18.8$\pm$8.6}  %
\newcommand{\eFourteenHalfTurnTwoWrongDI}{$-$29.5}  %
\newcommand{\eFourteenHalfTurnTwoWrongDStdI}{18.9}  %
\newcommand{\eFourteenHalfTurnTwoWrongCiI}{[$-$45.4, $-$14.4]}  %
\newcommand{\eFourteenHalfTurnThreeAllArmI}{50.8$\pm$2.8}  %
\newcommand{\eFourteenHalfTurnThreeAllDI}{$-$3.6}  %
\newcommand{\eFourteenHalfTurnThreeAllDStdI}{4.9}  %
\newcommand{\eFourteenHalfTurnThreeAllCiI}{[$-$6.5, $-$0.7]}  %
\newcommand{\eFourteenHalfTurnThreeWrongArmI}{28.1$\pm$4.3}  %
\newcommand{\eFourteenHalfTurnThreeWrongDI}{$-$22.1}  %
\newcommand{\eFourteenHalfTurnThreeWrongDStdI}{15.1}  %
\newcommand{\eFourteenHalfTurnThreeWrongCiI}{[$-$31.3, $-$12.9]}  %
\newcommand{\eFourteenHalfFinalAllArmI}{53.1$\pm$2.9}  %
\newcommand{\eFourteenHalfFinalAllDI}{$-$2.5}  %
\newcommand{\eFourteenHalfFinalAllDStdI}{3.6}  %
\newcommand{\eFourteenHalfFinalAllCiI}{[$-$5.6, +0.7]}  %
\newcommand{\eFourteenHalfFinalWrongArmI}{38.9$\pm$2.6}  %
\newcommand{\eFourteenHalfFinalWrongDI}{$-$19.3}  %
\newcommand{\eFourteenHalfFinalWrongDStdI}{6.7}  %
\newcommand{\eFourteenHalfFinalWrongCiI}{[$-$29.4, $-$8.3]}  %
\newcommand{\eFourteenRngTurnOneAllDLo}{$-$0.1}  %
\newcommand{\eFourteenRngFinalAllLossLo}{2.0}  %
\newcommand{\eFourteenRngFinalAllLossHi}{4.2}  %
\newcommand{\eFourteenRngTurnTwoWrongDLo}{$-$59.0}  %
\newcommand{\eFourteenRngTurnTwoWrongDHi}{$-$29.5}  %
\newcommand{\eFourteenRngFinalWrongDLo}{$-$21.9}  %
\newcommand{\eFourteenRngFinalWrongDHi}{$-$19.1}  %
\newcommand{\eFourteenRngFinalWrongShareLo}{6.0}  %
\newcommand{\eFourteenRngFinalWrongShareHi}{14.4}  %
\newcommand{\eFourteenCleanFinalAllE}{67.4}  %
\newcommand{\eFourteenCleanFinalAllI}{55.6}  %
\newcommand{\eFourteenQuarterTOneTurnThreeArmE}{41.0$\pm$30.4}  %
\newcommand{\eFourteenQuarterTOneTurnThreeDE}{$-$43.1}  %
\newcommand{\eFourteenQuarterTOneTurnThreeDStdE}{21.4}  %
\newcommand{\eFourteenQuarterTOneTurnThreeCiE}{[$-$70.8, $-$11.1]}  %
\newcommand{\eFourteenQuarterTOneFinalArmE}{66.0$\pm$9.4}  %
\newcommand{\eFourteenQuarterTOneFinalDE}{$-$12.5}  %
\newcommand{\eFourteenQuarterTOneFinalDStdE}{12.5}  %
\newcommand{\eFourteenQuarterTOneFinalCiE}{[$-$26.2, +2.1]}  %
\newcommand{\eFourteenQuarterTOneNIntE}{13}  %
\newcommand{\eFourteenQuarterTOneTurnThreeArmI}{35.4$\pm$13.0}  %
\newcommand{\eFourteenQuarterTOneTurnThreeDI}{$-$17.4}  %
\newcommand{\eFourteenQuarterTOneTurnThreeDStdI}{24.0}  %
\newcommand{\eFourteenQuarterTOneTurnThreeCiI}{[$-$35.6, +6.5]}  %
\newcommand{\eFourteenQuarterTOneFinalArmI}{42.4$\pm$8.4}  %
\newcommand{\eFourteenQuarterTOneFinalDI}{$-$16.0}  %
\newcommand{\eFourteenQuarterTOneFinalDStdI}{9.4}  %
\newcommand{\eFourteenQuarterTOneFinalCiI}{[$-$36.5, +11.1]}  %
\newcommand{\eFourteenQuarterTOneNIntI}{17}  %
\newcommand{\eFourteenQuarterTOneChgFinalDI}{+16.7}  %
\newcommand{\eFourteenQuarterTOneChgFinalCiI}{[$-$9.6, +46.7]}  %
\newcommand{\eFourteenHalfTOneTurnThreeArmE}{56.4$\pm$1.7}  %
\newcommand{\eFourteenHalfTOneTurnThreeDE}{$-$24.4}  %
\newcommand{\eFourteenHalfTOneTurnThreeDStdE}{6.9}  %
\newcommand{\eFourteenHalfTOneTurnThreeCiE}{[$-$40.6, $-$10.0]}  %
\newcommand{\eFourteenHalfTOneFinalArmE}{55.6$\pm$11.4}  %
\newcommand{\eFourteenHalfTOneFinalDE}{$-$20.2}  %
\newcommand{\eFourteenHalfTOneFinalDStdE}{13.4}  %
\newcommand{\eFourteenHalfTOneFinalCiE}{[$-$37.0, $-$3.5]}  %
\newcommand{\eFourteenHalfTOneNIntE}{39}  %
\newcommand{\eFourteenHalfTOneTurnThreeArmI}{40.3$\pm$16.7}  %
\newcommand{\eFourteenHalfTOneTurnThreeDI}{$-$12.7}  %
\newcommand{\eFourteenHalfTOneTurnThreeDStdI}{22.8}  %
\newcommand{\eFourteenHalfTOneTurnThreeCiI}{[$-$22.0, $-$4.4]}  %
\newcommand{\eFourteenHalfTOneFinalArmI}{45.1$\pm$28.5}  %
\newcommand{\eFourteenHalfTOneFinalDI}{$-$10.9}  %
\newcommand{\eFourteenHalfTOneFinalDStdI}{25.7}  %
\newcommand{\eFourteenHalfTOneFinalCiI}{[$-$23.9, +2.4]}  %
\newcommand{\eFourteenHalfTOneNIntI}{34}  %
\newcommand{\eFourteenRngTOneTurnTwoDLo}{$-$59.0}  %
\newcommand{\eFourteenRngTOneTurnTwoDHi}{$-$29.5}  %
\newcommand{\eFourteenRngTOneTurnThreeDLo}{$-$43.1}  %
\newcommand{\eFourteenRngTOneTurnThreeDHi}{$-$12.7}  %
\newcommand{\eFourteenRngTOneFinalDLo}{$-$20.2}  %
\newcommand{\eFourteenRngTOneFinalDHi}{$-$10.9}  %
\newcommand{\eFourteenRngTOneChgFinalDLo}{+16.7}  %
\newcommand{\eFourteenRngTOneChgFinalDHi}{+46.5}  %
\newcommand{\eFourteenRngTOneNIntLo}{13}  %
\newcommand{\eFourteenRngTOneNIntHi}{39}  %
\newcommand{\eFourteenNinst}{24}  %
\newcommand{\eFourteenNinter}{288}  %
\newcommand{\eFourteenNturnsRun}{216}  %
\newcommand{\eFourteenNoisyQuarterE}{56}  %
\newcommand{\eFourteenNoisyQuarterI}{44}  %
\newcommand{\eFourteenNoisyHalfE}{112}  %
\newcommand{\eFourteenNoisyHalfI}{108}  %
\newcommand{\eFourteenNoisyAll}{320}  %
\newcommand{\eFourteenChkKeptAll}{303}  %
\newcommand{\eFourteenChkKeptPct}{94.7}  %
\newcommand{\eFourteenOrigTrueAll}{132}  %
\newcommand{\eFourteenOrigNeitherAll}{188}  %

\newcommand{\ruN}{10}
\newcommand{\ruIncomplete}{3}
\newcommand{\ruRooms}{20}
\newcommand{\ruTrans}{40}
\newcommand{\ruBetter}{14}
\newcommand{\ruSame}{23}
\newcommand{\ruWorse}{3}
\newcommand{\ruSa}{5.2}
\newcommand{\ruSb}{5.5}
\newcommand{\ruSc}{5.8}
\newcommand{\ruRoomsUp}{10}

\newcommand{\eThirteenNineTurnOneCtrlE}{51.8$\pm$1.7}  %
\newcommand{\eThirteenNineTurnOneArmE}{52.8$\pm$2.1}  %
\newcommand{\eThirteenNineTurnOneDE}{+0.9}  %
\newcommand{\eThirteenNineTurnOneDStdE}{0.7}  %
\newcommand{\eThirteenNineTurnOneCiE}{[$-$1.4, +3.0]}  %
\newcommand{\eThirteenNineFinalCtrlE}{55.8$\pm$1.9}  %
\newcommand{\eThirteenNineFinalArmE}{57.2$\pm$1.1}  %
\newcommand{\eThirteenNineFinalDE}{+1.4}  %
\newcommand{\eThirteenNineFinalDStdE}{2.4}  %
\newcommand{\eThirteenNineFinalCiE}{[$-$1.4, +4.2]}  %
\newcommand{\eThirteenNineTurnOneCtrlI}{35.9$\pm$2.3}  %
\newcommand{\eThirteenNineTurnOneArmI}{37.0$\pm$2.0}  %
\newcommand{\eThirteenNineTurnOneDI}{+1.1}  %
\newcommand{\eThirteenNineTurnOneDStdI}{1.1}  %
\newcommand{\eThirteenNineTurnOneCiI}{[$-$1.1, +3.2]}  %
\newcommand{\eThirteenNineFinalCtrlI}{47.9$\pm$1.0}  %
\newcommand{\eThirteenNineFinalArmI}{46.3$\pm$0.3}  %
\newcommand{\eThirteenNineFinalDI}{$-$1.6}  %
\newcommand{\eThirteenNineFinalDStdI}{1.1}  %
\newcommand{\eThirteenNineFinalCiI}{[$-$5.1, +2.1]}  %
\newcommand{\eThirteenNineFinalMaxLoss}{5.1}  %
\newcommand{\eThirteenAblTurnOneCtrlE}{63.8$\pm$1.6}  %
\newcommand{\eThirteenAblTurnOneArmE}{64.7$\pm$0.9}  %
\newcommand{\eThirteenAblTurnOneDE}{+0.9}  %
\newcommand{\eThirteenAblTurnOneDStdE}{0.7}  %
\newcommand{\eThirteenAblTurnOneCiE}{[$-$2.6, +4.2]}  %
\newcommand{\eThirteenAblFinalCtrlE}{67.4$\pm$1.5}  %
\newcommand{\eThirteenAblFinalArmE}{65.8$\pm$1.9}  %
\newcommand{\eThirteenAblFinalDE}{$-$1.5}  %
\newcommand{\eThirteenAblFinalDStdE}{1.9}  %
\newcommand{\eThirteenAblFinalCiE}{[$-$4.7, +1.9]}  %
\newcommand{\eThirteenAblTurnOneCtrlI}{40.3$\pm$2.7}  %
\newcommand{\eThirteenAblTurnOneArmI}{40.2$\pm$0.7}  %
\newcommand{\eThirteenAblTurnOneDI}{$-$0.1}  %
\newcommand{\eThirteenAblTurnOneDStdI}{2.3}  %
\newcommand{\eThirteenAblTurnOneCiI}{[$-$2.3, +2.2]}  %
\newcommand{\eThirteenAblFinalCtrlI}{55.6$\pm$2.7}  %
\newcommand{\eThirteenAblFinalArmI}{55.8$\pm$0.9}  %
\newcommand{\eThirteenAblFinalDI}{+0.3}  %
\newcommand{\eThirteenAblFinalDStdI}{2.4}  %
\newcommand{\eThirteenAblFinalCiI}{[$-$3.9, +4.4]}  %
\newcommand{\eThirteenCallsMean}{1{,}681}  %
\newcommand{\eThirteenCallsMin}{574}  %
\newcommand{\eThirteenCallsMax}{2{,}972}  %

\newcommand{\revDOpenTokIn}{564.6}  %
\newcommand{\revDJevTokIn}{36.1}  %
\newcommand{\revDJevJevTokIn}{226.2}  %
\providecommand{\dci}[2]{\begin{tabular}[t]{@{}c@{}}#1\\[-2pt]{\scriptsize#2}\end{tabular}}
\providecommand{\tbd}{{\color{red}?}}
\providecommand{\orcc}[1]{\ifcsname trOrcOurs#1\endcsname\csname trOrcOurs#1\endcsname\else\tbd\fi}
\providecommand{\qualb}[1]{\ifcsname trQualOursbulb#1\endcsname\csname trQualOursbulb#1\endcsname\else\tbd\fi}

\lstdefinestyle{prompt}{basicstyle=\ttfamily\scriptsize, breaklines=true, breakatwhitespace=false,
  columns=fullflexible, keepspaces=true, frame=single, framesep=2pt, rulecolor=\color{black!40},
  xleftmargin=3pt, xrightmargin=3pt, showstringspaces=false, captionpos=t,
  abovecaptionskip=2pt, belowcaptionskip=3pt, aboveskip=6pt, belowskip=6pt,
  literate={—}{{---}}1 {–}{{--}}1 {’}{{\textquotesingle}}1 {“}{{\textquotedblleft}}1 {”}{{\textquotedblright}}1 {é}{{\'e}}1 {'}{{\textquotesingle}}1}
\lstdefinestyle{json}{style=prompt, backgroundcolor=\color{black!3}}
\renewcommand{\lstlistingname}{Listing}
\tcbset{colback=white, colframe=black!40, boxrule=0.4pt, arc=0pt, left=3pt, right=3pt, top=2pt, bottom=2pt,
  colbacktitle=black!8, coltitle=black}
\makeatletter
\newcommand{\supptoc}{\begingroup\def\numberline##1\hskip -1em.~{\hbox to\@tempdima{##1.\hfil}}\tableofcontents\endgroup}
\newcommand{\tocnonum}{\hbox to\@tempdima{\hfil}}
\makeatother

\hypersetup{pdftitle={PreferenceScene: Personalized Indoor Scene Synthesis from Feedback},pdfauthor={Seongrae Noh, RyunHa Lee, HyeongYeop Kang}}
\begin{document}
\addtocontents{toc}{\protect\setcounter{tocdepth}{-1}}
\maketitle
\begingroup\renewcommand{\thefootnote}{\fnsymbol{footnote}}\footnotetext[2]{Corresponding author.}\endgroup
\begin{strip}
  \centering
  \includegraphics[width=\textwidth]{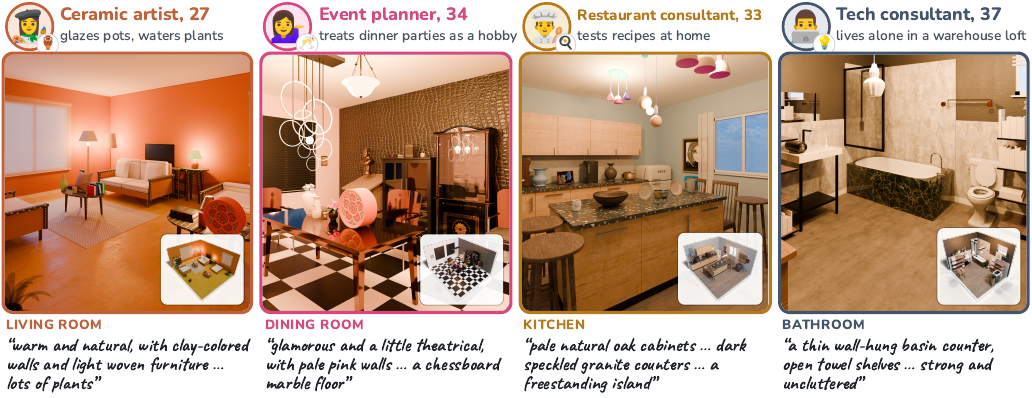}
  \captionof{figure}{PreferenceScene designs a room for the person who asks.
    Four simulated users of our test set, each with a summary of the persona, the 3D scene that PreferenceScene reached after three turns, and a quote from the instruction.
    Insets: the whole scene.}
  \label{fig:teaser}
\end{strip}

\begin{abstract}
Text-to-scene methods turn a short instruction into a furnished 3D room, but the room that a particular person wants rarely comes across fully in words: many preferences surface only when the person sees a concrete room and can point to what they like and dislike in it.
We therefore study personalized indoor scene synthesis under hidden user preferences: over several turns, a system shows a 3D scene, and the user, who sees only rendered images of the scene, replies with brief free-form feedback and a satisfaction score.
The system must therefore bring the whole room toward the user's preferences from feedback that comments on only part of it at a time.
It must also be able to change every part of the room, because a preference concerns not only single objects but the look and atmosphere that surfaces, lighting, textiles, furniture and their arrangement create together.
We present PreferenceScene, in which a large language model keeps an explicit estimate of the user's preferences for every design attribute of the room and, after each turn, infers the taste behind the feedback and revises the whole estimate, including the parts that the feedback does not mention.
A scene instantiation module builds every estimated value as its own part of the room, so that each of these parts can follow the user's preferences.
We evaluate on the test set of a benchmark with 320 instances in ten room types, with a simulated user and a render-based evaluator.
PreferenceScene matches the hidden preferences better than five published text-to-scene baselines adapted to the same interaction, and its match rises with every turn.
It even surpasses them when they are given the preferences directly, and a vision-language model rates its scenes as the most realistic and functional.
\end{abstract}

\section{Introduction}

Indoor scene synthesis generates furnished 3D rooms, and systems built on large language models (LLMs) now produce a plausible room from a short text instruction~\cite{feng2023layoutgpt,yang2024holodeck,celen2025idesign,sun2025layoutvlm,gu2026r3l}.
A room for a particular person, however, is hard to obtain from a single instruction.
A person who asks for a cozy bedroom may picture warm lamp light, linen bedding and pale oak floors but write only ``cozy''. The system fills in everything else, and the room it returns can match the words and still miss the person.
Users of text-to-image models meet the same gap: a text description conveys only part of the intended image, and users reach what they want by trial and error~\cite{hutchinson2022underspecification,liu2022design,hahn2025proactive}.
Instead of expecting one instruction to specify a room, a system that designs for a particular person should work with them over several turns.

Interaction does more than correct errors.
People often cannot list their preferences in advance, and many of them become clear only while they evaluate concrete options~\cite{slovic1995construction,lichtenstein2006construction}, which is why recommender and design tools let users react to examples instead of stating their requirements up front~\cite{viappiani2006preference,chen2012critiquing,schon1992designing}.
Shown a room, a person notices that the light is too cold or that the floor should be darker, preferences that a short instruction would not have stated.
We call such preferences \emph{hidden}: the system is not told them, and they appear only in the user's reactions to the scenes that it shows.
These reactions are also sparse.
A person comments on the few things that catch their eye, and asking for a judgment on every detail of a room would soon tire them, a known limit of interactive systems that rely on human judgments~\cite{takagi2001interactive,boutilier2002pomdp}.
A system must therefore bring the whole room toward the person's preferences from feedback that names only some of its parts.

The preferences concern the room as a whole.
The look and atmosphere of a room come from the combination of its surfaces, lighting, textiles, furniture and arrangement~\cite{ching2012interior}, and methods that select furniture styles or color schemes optimize compatibility over the whole scene rather than for each object~\cite{liu2015style,chen2015magic}.
A user's feedback moves freely among these parts, from the wall color to the bedding to the light of a lamp.
A personalized system must therefore be able to change each of these parts.
The five published text-to-scene baselines that we compare with~\cite{yang2024holodeck,feng2023layoutgpt,celen2025idesign,sun2025layoutvlm,gu2026r3l} select and place objects and at most pick wall and floor materials from a fixed library, so they cannot change the other parts.
Iterative and editing systems follow only what their instructions state~\cite{hu2024scenecraft,yang2025sceneweaver,zheng2025editroom,bucher2025respace,wang2026chat2layout}, and preference elicitation infers a hidden target from user responses only over items or images~\cite{sadigh2017active,handa2024bayesian,li2025eliciting,guo2018dialog,nabati2025pasta}, not over the design of a 3D room.

We formulate this setting as \emph{personalized indoor scene synthesis under hidden user preferences} (\cref{sec:problem}), in which the user sees only rendered images of each shown scene and replies with brief free-form feedback.
Our benchmark (\cref{sec:benchmark}) gives each simulated user a hidden design profile that the instruction states in part or not at all, as in elicitation studies with simulated users~\cite{gao2024aligning,hahn2025proactive,li2025eliciting}, and an evaluator checks which hidden values the final scene shows.

Our method, PreferenceScene, rests on one design principle: the system keeps its estimate of the user's preferences as an explicit \emph{plan} that assigns one value to every design attribute, and it instantiates every value as its own part of the scene.
Each part of the room is then a hypothesis about the user that feedback can confirm or overturn, and since every plan covers every attribute, the system reconsiders the whole room after each turn, including the parts that the feedback does not mention.
An LLM builds a space of candidate values for each attribute before the first turn and rewrites the plan from the feedback after every turn.

Against these systems, adapted to the same interaction, PreferenceScene achieves the highest preference match in both benchmark forms, also on the attributes that the feedback never mentions, and its match rises with every turn. \Cref{fig:teaser} shows final rooms of PreferenceScene for four users.
With feedback alone, it matches more hidden preferences than every published baseline that is given the preferences as text, and a vision-language model rates its scenes as the most realistic and functional.
In a user study, participants prefer its scenes to those of every baseline for both preferences and design (\cref{sec:exp_human}).

In summary, our contributions are as follows:
\begin{itemize}
  \item We present PreferenceScene, which models a user's preferences as an explicit plan over a space of candidate values for every design attribute and brings the whole room toward them from sparse free-form feedback (\cref{sec:method}).
  \item We give the plan control over the whole room with a scene instantiation module that sets surfaces, lighting, textiles and the materials and colors of furniture (\cref{sec:method_repr}).
  \item We formulate personalized indoor scene synthesis under hidden user preferences and build a benchmark of ten room types in two forms, with simulated users who hold hidden design profiles and see only rendered images (\cref{sec:problem,sec:benchmark}).
  \item We compare PreferenceScene with five published baselines adapted to the same interaction and test it with ablations, a second evaluator and a user study (\cref{sec:experiments}).
\end{itemize}

\section{Related work}
\label{sec:related}

Text-driven indoor scene synthesis uses an LLM to predict a layout~\cite{feng2023layoutgpt}, to propose constraints for an optimizer~\cite{yang2024holodeck,sun2025layoutvlm} or to reason about spatial relations~\cite{gu2026r3l}, and other methods combine such steps in multi-step pipelines~\cite{celen2025idesign,pun2026hsm} or iterate with their own critic~\cite{xia2026sage,pfaff2026scenesmith,yang2025sceneweaver}.
In these methods, the input text defines the target scene, and editing methods change a scene as instructed~\cite{zheng2025editroom,bucher2025respace,noh2026editasact}.
Among the methods that use preferences, I-Design~\cite{celen2025idesign} conditions on the preferences that the user writes in the instruction, and OptiScene~\cite{yang2025optiscene} and ReSpace~\cite{bucher2025respace} train one model for all users on preference pairs.
Interactive retrieval and preference elicitation infer a target that the user has in mind from feedback, comparisons or choices~\cite{guo2018dialog,sadigh2017active,koyama2020sequential,gao2024aligning,hahn2025proactive}, and dialogue-state tracking keeps an estimate of the user's goal over the turns~\cite{williams2013dialog}.
Our system infers the preferences that one user does not state from free-form feedback on rendered images of full 3D scenes, and keeps them as a plan over the design attributes of the room.
As in this line of work, we evaluate with simulated users that hold a hidden target~\cite{gao2024aligning,hahn2025proactive}, and our evaluator answers a multiple-choice question per attribute on the rendered images, like TIFA~\cite{hu2023tifa}. Because LLM evaluators favor their own generations~\cite{panickssery2024llm}, each role uses a different model family.
Supplementary \cref{S-sec:supp_related} extends this section.

\section{Problem definition}
\label{sec:problem}

In this task, a system interacts with a user over several turns and must deliver a final 3D scene that shows the user's hidden preferences, which the user does not state and reveals only through feedback on rendered images.

\begin{figure}[t]
  \centering
\includegraphics[width=\columnwidth]{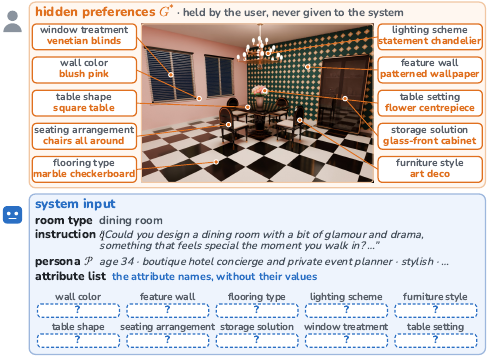}
  \caption{An instance (the user of \cref{fig:overview,fig:space}; instruction and persona shortened) and the reference scene built from $G^*$ (\cref{sec:benchmark}).
    Hidden values such as the blush pink walls and the marble checkerboard floor appear nowhere in the instruction or persona.}
  \label{fig:instance}
\end{figure}

\paragraph{Instance.}
An instance (\cref{fig:instance}) consists of the user's hidden preferences and the input that a system sees: a room type, an instruction $q$ in which the user asks for the room, and a persona $\mathcal{P}$ that describes the user.
Each room type has a schema of $M$ design attributes, \eg the flooring, and each attribute $a$ has a finite set $\mathcal{V}_a$ of mutually exclusive values, \eg wood-look planks or terrazzo.
The \emph{attribute list} is the schema without its values: the $M$ attribute names, each with a one-sentence description that names no value, which tell a system which design choices matter to the user but not which values the user wants.
The \emph{hidden preferences} $G^*$ assign one value $G^*[a]\in\mathcal{V}_a$ to every attribute $a$, in the same form as the plans of \cref{sec:method}.
The instruction and the persona can state some of these values and hint at others.

\paragraph{Interaction.}
A system receives the room type, $q$, $\mathcal{P}$ and the attribute list. It never receives the value sets $\mathcal{V}_a$ or $G^*$, and it does not update model parameters during the interaction.
The system then interacts with the user for $K$ turns. One such run on one instance is an \emph{interaction}.
In turn $k$, the system shows a 3D scene. The user sees its rendered images and returns natural-language feedback $\eta_k$ and a score $z_k\in\{1,\dots,10\}$ of how satisfied the user is with the room.
After turn $K$, the system outputs one more scene, the final scene $X$, which receives no feedback.

\paragraph{Evaluation.}
We measure how well the final scene $X$ shows $G^*$ by the \emph{perceived attribute match rate} (pAMR), the share of hidden values that an evaluator recovers from rendered images of $X$:
\begin{equation}
  \mathrm{pAMR}(X,G^*)=\frac{1}{M}\sum_{a}\mathbf{1}\big[r(X,a)=G^*[a]\big],
  \label{eq:pamr}
\end{equation}
where $r(X,a)\in\mathcal{V}_a$ is the answer of a vision-language model (VLM) to a multiple-choice question about attribute $a$ on the rendered images of $X$, with the values of $\mathcal{V}_a$ as options.
The indicator is averaged over two option orders (\cref{sec:exp_protocol}).
pAMR is computed on rendered images alone, so it applies to any system that outputs a scene, and we report it in percent, averaged over instances.

\begin{figure*}[t]
  \centering
\includegraphics[width=\textwidth]{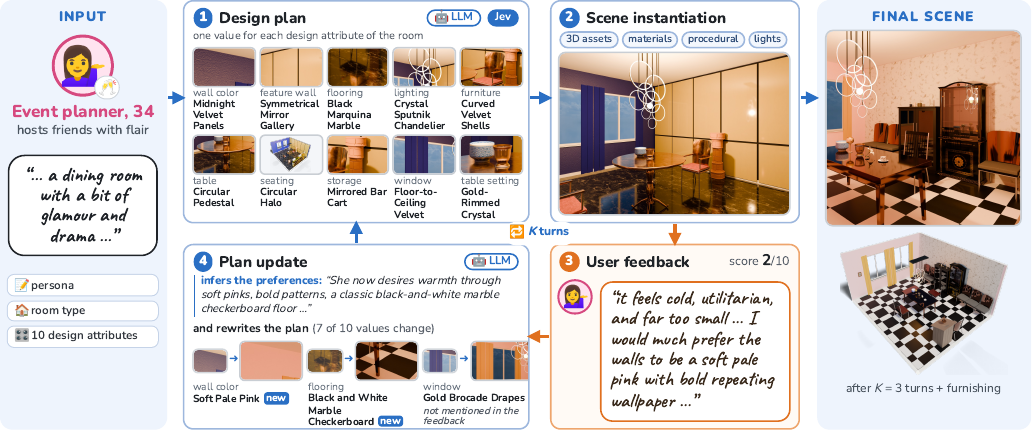}
  \caption{Overview of PreferenceScene on the first turn of a test interaction (second user of \cref{fig:teaser}, \vI{} form).
    (1) An LLM writes the design plan, one value for each design attribute.
    (2) Scene instantiation builds the plan as a 3D scene.
    (3) The user answers with feedback and a score.
    (4) The LLM infers the user's preferences and rewrites the plan. ``New'' marks a value outside the candidates.
    After $K$ turns, one more LLM call furnishes the final scene.}
  \label{fig:overview}
\end{figure*}

\section{Method}
\label{sec:method}

PreferenceScene keeps its current estimate of the user's preferences in an explicit form, so that every turn of feedback can change this estimate and every change shows in the room (\cref{fig:overview}).
The estimate is a \emph{plan}: the plan $\hat{S}_k$ of turn $k$ assigns each of the $M$ design attributes $a$ of the room one value $\hat{S}_k[a]$, a short text such as ``pale oak planks'' for the flooring.
Three components work on the plan.
Before the first turn, an LLM writes a space of candidate values for the attributes and chooses an initial plan from it (\cref{sec:method_cand}).
After every turn, the LLM revises the plan from the user's feedback (\cref{sec:method_update}).
Scene instantiation builds each plan as a 3D scene in which every value determines its own part of the room (\cref{sec:method_repr}).
We write $c$ for the system input of \cref{sec:problem}. No component reads the hidden preferences $G^*$ or the value sets $\mathcal{V}_a$.

\subsection{Preference space and initial plan}
\label{sec:method_cand}

The plan chooses its values from a set of options per attribute.
If the LLM rewrote every value each turn, a value's wording could change while the room stayed the same, and the plans of two turns would be hard to compare. A fixed list of options, however, cannot fit every user.
We therefore let the LLM write the options for each user once, before the first turn, and keep them for the whole interaction.

\begin{figure}[t]
  \centering
  \includegraphics[width=\columnwidth]{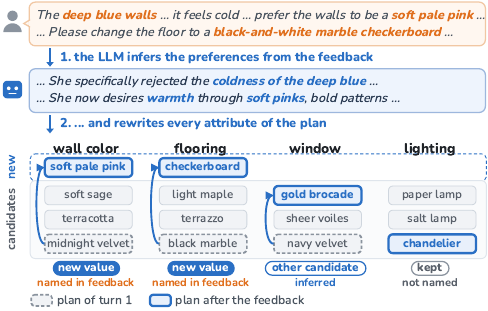}
  \caption{How the plan moves after the feedback of turn 1 (interaction of \cref{fig:overview}).
    The walls and the floor, which the feedback names, receive new values; the window, which it does not name, moves to a candidate that fits the inferred warmth.}
  \label{fig:space}
\end{figure}

Given the room type, $q$ and $\mathcal{P}$, one LLM call lists the visible design aspects of the room, such as the wall finish or the seating, and twelve candidate values for each aspect.
Six candidates are the choices that the LLM considers most likely for this person and six are other common choices, so that the space also contains options that the input does not suggest.
Each candidate has a short label, a one-sentence description of its look and, if it names a color, a hex code.
Because the LLM names the aspects in its own words, an aspect does not always correspond to one entry of the attribute list.
We therefore assign the aspects to the attributes with Jev~\cite{typesafe2026jev}, a decision model that receives a context and a set of multiple-choice questions and returns a probability for every option of each question.
The context is the room type, and each aspect receives one question, which quotes the aspect name, its description and its first three candidate labels. The options are the $M$ entries of the attribute list, each given by its name and description, and ``none''.
An aspect is assigned to its most probable attribute if that probability is at least 0.5, and further rules, given in supplementary \cref{S-sec:supp_cand}, ensure that every attribute receives at least one aspect.
The candidates of the aspects assigned to attribute $a$ form its candidate set $C_a$, and all candidate sets form the \emph{preference space} of the interaction (\cref{fig:space}).

A value of the plan is either a candidate from $C_a$ or a \emph{new value}, which the LLM writes in the same form as a candidate when no candidate fits what the feedback asks for.
A second LLM call reads $c$ and all candidate sets, writes a short analysis of the person's likely preferences and then chooses the initial plan $\hat{S}_1$.

\subsection{Plan update}
\label{sec:method_update}

The feedback of a turn mentions only some parts of the room, and it describes what the user saw, not what the plan specified.
A remark such as ``the room feels too dark'' can concern the wall color, the floor or the lamps, or a part that the scene shows differently from the plan.
The update therefore gives the LLM the whole history of the interaction, including what each shown scene contained.

After turn $k$, one LLM call writes the next plan,
\begin{equation}
  \hat{S}_{k+1}=\mathrm{LLM}\big(c,\,\{(a,C_a)\}_{a},\,(\hat{S}_{k'},D_{k'},\eta_{k'},z_{k'})_{k'=1}^{k}\big),
  \label{eq:update}
\end{equation}
where $\eta_{k'}$ and $z_{k'}$ are the feedback and score of turn $k'$, and $D_{k'}$ is the \emph{scene description} of turn $k'$: per attribute, a text stating what was built for its value (\cref{sec:method_repr}).

The LLM first infers the person's preferences from all feedback and scores and then writes a complete plan.
Because the plan must again assign a value to every attribute, the LLM decides for every part of the room whether to keep its value, to switch to another candidate or to write a new value, including the parts that the feedback does not mention.
A reaction to one part can therefore change others: a remark that the room feels cold, for example, can lead to warmer lamp light and a warmer wood at the same time.
After turn $K$, the same call writes the final plan $\hat{S}_{K+1}$, from which we build the final scene $X$.

\subsection{Scene instantiation}
\label{sec:method_repr}

Scene instantiation must let every value change its own part of the room.
A value is free text, so we first translate it into a \emph{specification}, a structured description of the part that the value determines: which objects it contains and how many, where they stand, and which finish, color or light they have.
Jev answers the questions that make up the specification.
Each attribute belongs to a component category, such as floor, seating or lighting, and each category has a fixed set of multiple-choice questions. Every question quotes the attribute, its description and the text of the value, and we take the most probable option of each question.
Colors come from the hex code of the value.
Examples and further explanation of specifications are given in supplementary \cref{S-sec:supp_questions,S-sec:supp_examples}.

We then build the scene from the specifications, based on the object retrieval and the constraint-based layout of Holodeck~\cite{yang2024holodeck}.
Objects are retrieved from Objaverse~\cite{deitke2023objaverse} assets by their size and by their visual and textual similarity to the text of the value, and floor and wall materials are retrieved from ambientCG~\cite{ambientcg}.
Our layout differs in where its constraints come from: the placement answers of the specification, such as the wall that a group of furniture stands against and the direction it faces, replace the constraints that an LLM writes. A search tries their combinations in the order of their probabilities and keeps the first layout that satisfies the hard constraints, such as no collisions.
Retrieval from a fixed asset pool cannot set the material or color of a piece of furniture, so we rebuild furniture procedurally inside the bounding box of the retrieved asset, with the shape, materials and colors that the specification sets.
Supplementary \cref{S-sec:supp_questions,S-sec:supp_retrieval,S-sec:supp_proc,S-sec:supp_place} give the categories and questions, the retrieval, the procedural library and the placement rules.

Only the final scene $X$ receives \emph{furnishing}, because a plan decides how the room looks but not every item that a furnished room contains.
One more LLM call adds further furniture, objects on furniture tops and pieces for the walls in the free space, without moving any object of the plan.
So that the additions do not contradict the plan, the same reply first lists, for each attribute, the object types that its value rules out, and pieces of these types are dropped.
An example of these lists and further explanation are given in supplementary \cref{S-sec:supp_furnish,S-sec:supp_examples}.

Finally, templates write the scene description $D_k$ of every attribute from what was built, \ie from the specification, the placement, the captions of the retrieved assets and their materials and colors, not from the words of the value, so that the description reports what the user could see.
The supplementary material gives details and examples of scene descriptions (\cref{S-sec:supp_descr,S-sec:supp_examples}), the prompts of all LLM calls (\cref{S-sec:supp_method}) and an algorithm listing (\cref{S-sec:supp_alg}).

\section{Benchmark}
\label{sec:benchmark}
\begin{figure*}[t]
  \centering
  \includegraphics[width=\textwidth]{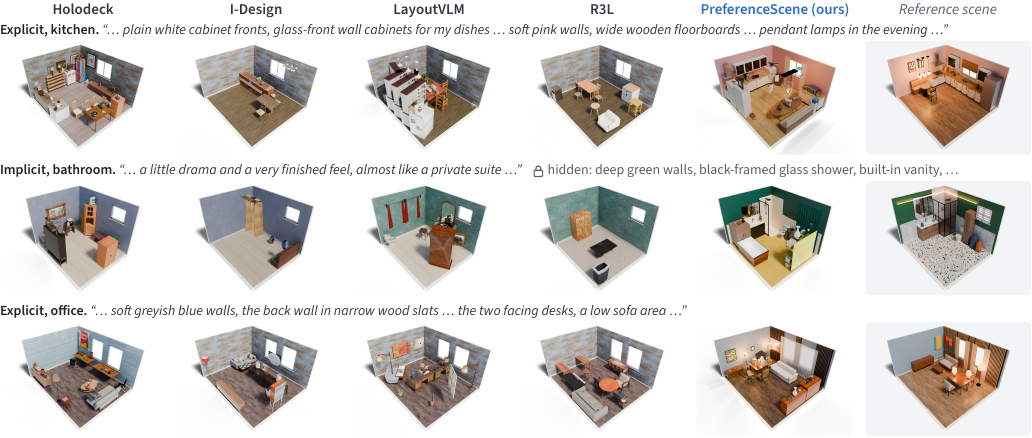}
  \caption{Final scenes after $K=3$ turns on three test instances.
    Each row gives the form, the room type and an excerpt of the instruction, and the \vI{} row also three hidden values that the instruction does not state.}
  \label{fig:qual}
\end{figure*}

Indoor scene datasets such as 3D-FRONT~\cite{fu20213dfront} contain rooms but no users, so we build a benchmark in which every instance has known hidden preferences and an input that states a controlled part of them.
It covers ten room types, each with a schema of $M=10$ attributes, and each attribute is one visible component of the room, \eg the wall paint, the floor or the bedding (100 attributes with 483 values in total).
Each room type has eight \emph{profiles}: a profile is one coherent design, written by an LLM as ten values of the schema, and it serves as the hidden preferences $G^*$ of its instances.
For every profile, GPT-5.4-mini~\cite{openai2026gpt54mini} writes two personas and, in two forms, their instructions, which gives 160 instances per form and 320 in total.
In the \vE{} form, the persona and the instruction state four to six of the ten hidden values, so that a system can read about half of the profile from its input; in the \vI{} form, they state none and only hint at the person's taste, so that the profile must be inferred from the interaction.
We split the instances by profile into a development subset of 40 instances per form, on which we developed the method, and a test subset of 120 instances per form, on which all results of \cref{sec:experiments} are measured.

The simulated user, Qwen3.8-27B~\cite{qwen2026qwen38}, holds its taste as a free-form \emph{taste narrative} of about 120 words that conveys every hidden value.
In every turn, it receives rendered images of the shown scene, the persona, the narrative and its earlier feedback and scores, and it writes three to five sentences of feedback and a satisfaction score from 1 to 10; it never sees the system's plan or the schema.
A rule-based construction, separate from our scene instantiation, draws every schema value directly. We revised the values that a vision-language model could not read back from its rendered images, and the same construction builds a \emph{reference scene} for every profile.
The evaluator of \cref{sec:exp_protocol} recovers 89.3\% of the hidden values from the reference scenes, against 21.4\% for a random guess.
Supplementary \cref{S-sec:supp_bench,S-sec:supp_full} give the schemas, the statistics, an example instance and the prompts.

\providecommand{\hdtwo}[2]{\begin{tabular}[b]{@{}c@{}}#1\\#2\end{tabular}}

\providecommand{\flag}[1]{\textcolor{red}{#1}}
\section{Experiments}
\label{sec:experiments}

\subsection{Setup}
\label{sec:exp_protocol}

\paragraph{Runs and models.}
Every interaction has $K=3$ turns.
Every system runs each instance four times, with seeds 0--3 for the simulated user and for the system's LLM, and we report the mean and standard deviation over the seeds.
GLM-4.6V-Flash~\cite{glm2025glm46v} is the evaluator, and Gemma-4-31B~\cite{gemmateam2026gemma4} is the LLM of every system.

\paragraph{Baselines.}
We compare with five published text-to-scene baselines: Holodeck~\cite{yang2024holodeck}, LayoutGPT~\cite{feng2023layoutgpt}, I-Design~\cite{celen2025idesign}, LayoutVLM~\cite{sun2025layoutvlm} and R3L~\cite{gu2026r3l}.
None of them takes feedback, so we run each one in the same interaction as our method.
In turn 1, a system generates a scene from the input $c$.
In every later turn, it generates a new scene from $c$ and the history of the interaction, \ie a description of each scene it showed, the user's feedback on it and the score.
We also build an editor from Holodeck, \emph{Holodeck-Edit}, which edits the floor, wall and object specification of its previous scene in every turn and keeps the assets of unchanged objects.
In the \emph{oracle} condition, as in PRELUDE~\cite{gao2024aligning}, a system receives the hidden preferences $G^*$ as text and generates one scene, which shows what it can realize when nothing has to be inferred.
Without feedback, Holodeck-Edit shows the scene of Holodeck, so its oracle value is that of Holodeck ($^{*}$ in \cref{tab:main}).
All systems use the same room shell (the walls, floor, doors and windows of each room type), the same pool of 6{,}793 Objaverse~\cite{deitke2023objaverse} assets and the same renderer (Blender Cycles~\cite{blender2025}) and cameras.
Supplementary \cref{S-sec:supp_base,S-sec:supp_variants} describe the adaptations, the handling of failed generations and further variants of the published baselines.
LayoutGPT has in-context exemplars for bedrooms and living rooms only, so it runs on the 24 test instances of these two room types in each form.

\paragraph{Metrics.}
Our main metric is pAMR (\cref{eq:pamr}), for which the evaluator answers the question of each attribute on two rendered images of the final scene, a cutaway view from above and an eye-level view.
Because the answers of a VLM depend on the option order, the evaluator answers each question twice, with the options in the given and in the reversed order, and pAMR averages the two answers.
For scene quality, we follow R3L~\cite{gu2026r3l}: \emph{Real.}, \emph{Func.} and \emph{Instr.} are the 1--10 scores of its evaluation prompt for realism, functionality and instruction following, answered by the evaluator.
\emph{Coll.} is the percentage of objects in collision under the mesh-intersection test of SceneEval~\cite{tam2026sceneeval}.
For cost, we report the number of LLM calls per interaction and the generation time per scene (\emph{Time}), \ie the wall time of a system's own steps divided by the $K+1$ scenes of an interaction.
We measure the time in a separate \emph{timing run} of eight interactions per system, one at a time on the same hardware.
In every table, the best value of each metric is in bold and the second best is underlined.

\subsection{Comparison with baselines}

\begin{table}[t]
  \centering
  \footnotesize
  \setlength{\tabcolsep}{6pt}
  \caption{Preference match: pAMR (\%, $\uparrow$) of the final scene, mean $\pm$ standard deviation over four seeds (Oracle: one seed; one scene from the \vE{} input and the hidden values).
    Bold: best; underlined: second best. $^{*}$Same scene as Holodeck. $^{\dagger}$Bedrooms and living rooms only.
}
  \label{tab:main}
  \begin{tabularx}{\linewidth}{@{}l*{3}{Y}@{}}
    \toprule
    Method & \vI{} & \vE{} & Oracle \\
    \midrule
    Holodeck~\cite{yang2024holodeck} & 41.4$\pm$0.6& 44.0$\pm$0.2& 45.3\\
    LayoutGPT$^{\dagger}$~\cite{feng2023layoutgpt} & 25.2$\pm$1.3& 27.9$\pm$1.0& 25.8\\
    I-Design~\cite{celen2025idesign} & 41.0$\pm$0.5& 44.3$\pm$1.1& 43.8\\
    LayoutVLM~\cite{sun2025layoutvlm} & \underline{43.1$\pm$1.4} & 45.1$\pm$0.3& \underline{47.8} \\
    R3L~\cite{gu2026r3l} & 40.8$\pm$0.6& 43.9$\pm$1.0& 45.5\\
    Holodeck-Edit & 42.6$\pm$1.4& \underline{45.6$\pm$0.7} & 45.3$^{*}$ \\
    \methodname{} (Ours) & \textbf{49.6$\pm$1.7} & \textbf{56.6$\pm$0.3} & \textbf{59.4} \\
    \bottomrule
  \end{tabularx}
\end{table}

\paragraph{Preference match.}
\methodname{} reaches 56.6 in the \vE{} form and 49.6 in the \vI{} form, the highest match in both (\cref{tab:main}).
The best published baseline, LayoutVLM, reaches 45.1 and 43.1, and Holodeck-Edit reaches 45.6 and 42.6.
\methodname{} is at least 11 points above every baseline in the \vE{} form and at least 6 points in the \vI{} form.

\paragraph{Qualitative results.}
In the \vE{} kitchen of \cref{fig:qual}, \methodname{} shows the pink walls, the wide wooden floorboards and the white cabinet fronts that the instruction asks for, as the reference scene does.
In the \vI{} bathroom, the hidden values include deep green walls, a black-framed glass shower and a built-in vanity, none of which the input states. \methodname{} shows all three after the feedback, while the published baselines show blue-grey or teal walls.
In the office, only \methodname{} shows the wall of wood slats and the amber lamp light that the instruction asks for.
Supplementary \cref{S-sec:supp_qual,S-sec:supp_limitations} show more results, failure cases and the limitations of the method.

\paragraph{Control over the whole room.}
Because the published baselines at most pick wall and floor materials from a fixed library, they cannot follow the feedback on the other parts of the room.
Even in the oracle condition, which gives them the hidden preferences, their scenes show at most 47.8\% of the hidden values (\cref{tab:main}).
\methodname{} shows 59.4\% in the oracle condition, and with feedback alone it exceeds the oracle value of every published baseline in both forms.
Supplementary \cref{S-sec:supp_inst_baselines,S-sec:supp_shared_attrs} report the match when the baselines use our scene instantiation and on the attributes that every system can change.

\paragraph{The role of feedback.}
The match of \methodname{} rises with every turn (curve in supplementary \cref{S-fig:supp_rounds}), from 51.9 to 56.6 in the \vE{} form and from 37.5 to 49.6 in the \vI{} form.
In the \vI{} form, where the feedback is the main evidence, the gain from turn 1 to the final scene, +12.1 points, is larger than that of every baseline (at most 9.7 points).
The feedback is also sparse: an LLM labeller (Ministral-3-14B) finds that the feedback of a turn names four to five of the ten attributes, and that about a third of the hidden attributes are never named in an interaction.
On these never-named attributes, \methodname{} matches 54.0\% and 43.2\% of the hidden values, against at most 44.4\% and 37.6\% for the baselines: it also brings unmentioned parts of the room toward the user's preferences.
Supplementary \cref{S-sec:supp_noisy} tests noisy feedback.

\begin{table}[t]
  \centering
  \footnotesize
  \setlength{\tabcolsep}{0.8pt}
  \caption{Quality of the final scenes of both forms (bold: best; underlined: second best), mean $\pm$ standard deviation over four seeds.
    Real., Func., Instr.: 1--10 evaluator scores. Coll.: \% of colliding objects. Time: seconds per scene in the timing run. $^{\dagger}$Bedrooms and living rooms only.
}
  \label{tab:quality}
  \begin{tabularx}{\linewidth}{@{}l*{5}{Y}@{}}
    \toprule
    Method & Real.\,$\uparrow$ & Func.\,$\uparrow$ & Instr.\,$\uparrow$ & Coll.\,$\downarrow$ & Time\,$\downarrow$ \\
    \midrule
    Holodeck & \underline{7.9$\pm$0.0} & \underline{7.2$\pm$0.0} & \underline{7.9$\pm$0.1} & 5.7$\pm$0.5 & 102 \\
    LayoutGPT$^{\dagger}$ & 7.0$\pm$0.1 & 6.6$\pm$0.1 & 6.5$\pm$0.1 & 31.2$\pm$2.1 & \textbf{12} \\
    I-Design & 7.6$\pm$0.0 & 6.9$\pm$0.0 & 7.5$\pm$0.1 & \underline{2.8$\pm$0.4} & 340 \\
    LayoutVLM & 7.6$\pm$0.0 & 6.9$\pm$0.0 & 7.6$\pm$0.0 & 41.9$\pm$0.8 & 361 \\
    R3L & 7.6$\pm$0.0 & 7.0$\pm$0.0 & 7.4$\pm$0.1 & \textbf{1.4$\pm$0.4} & 332 \\
    Holodeck-Edit & 7.8$\pm$0.1 & \underline{7.2$\pm$0.1} & 7.8$\pm$0.0 & 4.5$\pm$0.3 & 79 \\
    \methodname{} (Ours) & \textbf{8.2$\pm$0.0} & \textbf{7.5$\pm$0.0} & \textbf{8.4$\pm$0.0} & \underline{2.8$\pm$0.5} & \underline{43} \\
    \bottomrule
  \end{tabularx}
\end{table}

\paragraph{Scene quality and cost.}
In \cref{tab:quality}, \methodname{} has the highest realism, functionality and instruction scores (8.2, 7.5 and 8.4, against at most 7.9, 7.2 and 7.9).
Of its objects, 2.8\% are in collision, and none is out of bounds.
An interaction of \methodname{} makes six LLM calls and about 25 requests to Jev.
Its generation time per scene is the second lowest of the seven systems.

\paragraph{Robustness.}\label{sec:exp_robust}
We re-score all final scenes with a second evaluator from another model family, Ministral-3-14B~\cite{mistral2026ministral3}, and with a variant of the evaluator that may answer ``not visible''.
\methodname{} stays above every baseline under both.
With the second evaluator, it scores 52.5 (\vE{}) and 46.5 (\vI{}), against at most 44.9 and 43.0.
With the ``not visible'' option, it scores 47.8 and 40.9, against at most 35.8 and 33.4.
Supplementary \cref{S-sec:supp_llm,S-sec:supp_open_jev} also report the method with a hosted LLM in place of Gemma-4-31B and with an open model in place of Jev.

\subsection{Ablations}

\begin{table}[t]
  \centering
  \footnotesize
  \setlength{\tabcolsep}{4pt}
  \caption{Ablations on the dining-room and office test instances: pAMR (\%, $\uparrow$), mean $\pm$ standard deviation over four seeds. Bold: best; underlined: second best.}
  \label{tab:ablation}
  \begin{tabularx}{\linewidth}{@{}l*{2}{Y}@{}}
    \toprule
    Variant & \vE{} & \vI{} \\
    \midrule
    \methodname{} (full) & \textbf{67.1$\pm$1.3} & \textbf{56.5$\pm$2.9} \\
    \midrule
    No iteration (initial plan) & 64.0$\pm$1.4 & 40.6$\pm$2.2 \\
    w/o older turns & \underline{66.4$\pm$2.9} & \underline{50.1$\pm$2.9} \\
    w/o new values & 64.0$\pm$2.8 & 48.4$\pm$2.4 \\
    \bottomrule
  \end{tabularx}
\end{table}

\Cref{tab:ablation} removes one component at a time on the dining-room and office test instances (24 per form).
\emph{No iteration} scores the scene of the initial plan, so its distance to the full method is the whole effect of the feedback: about 3 points in the \vE{} form and 16 points in the \vI{} form.
\emph{Without older turns}, the plan update (\cref{eq:update}) reads only the latest turn of the history. The match then drops to 66.4 and 50.1: the LLM uses the feedback of all turns, not only the latest.
\emph{Without new values}, a value outside the candidate set $C_a$ is rejected and the attribute keeps its previous value.
The match then drops to 64.0 and 48.4, although only about one value in six in the full method's final plans is new (16.2\% and 16.6\%): the plan must be able to leave the candidates when the user asks for something none of them offers.
Supplementary \cref{S-sec:supp_more_ablations} reports further ablations, including variants without the explicit plan, without the scene descriptions in the history and without the rules of scene instantiation.

\subsection{User study}
\label{sec:exp_human}

Each of the 20 participants completed 36 trials, six for each baseline.
In every trial, they read the instruction, the persona and the taste narrative and saw the final scenes of \methodname{} and of one baseline for the same instance, without system names and in random order, as in SceneSmith~\cite{pfaff2026scenesmith}.
They chose the scene that better matches the person's preferences and the scene that is better designed overall, regardless of these preferences.
\begin{figure}[t]
  \centering
  \includegraphics[width=\linewidth]{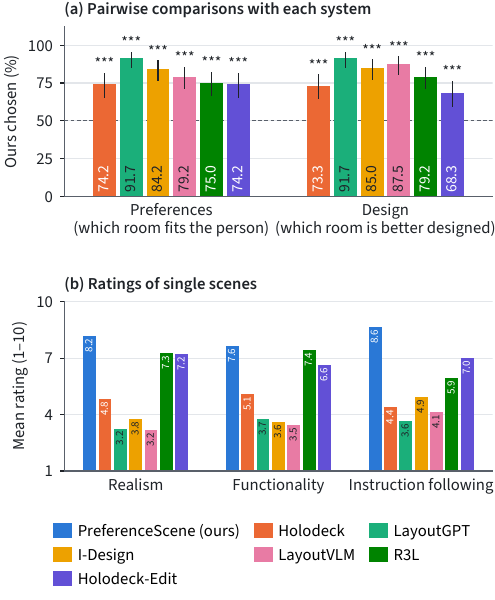}
  \caption{User study (20 participants). (a) Share of answers that chose \methodname{} over each system, with 95\% Wilson intervals; dashed: 50\%; * $p<0.05$, ** $p<0.01$, *** $p<0.001$ (corrected binomial test). (b) Mean ratings (1--10).}
  \label{fig:human}
\end{figure}
They chose \methodname{} for the preferences in 79.7\% of the 720 trials and for the design in 80.8\% (\cref{fig:human}a).
Against the individual baselines, these rates are 74.2--91.7\% and 68.3--91.7\%, and each differs from 50\% in a two-sided binomial test with the Benjamini-Hochberg correction over the 12 tests ($p<0.0001$).
Participants also rated 24 single scenes on a five-point scale for realism, functionality and instruction following, with the rubric, the block design and the mapping to 1--10 of R3L~\cite{gu2026r3l}.
Their mean ratings place \methodname{} first of the seven systems in realism, functionality and instruction following (\cref{fig:human}b).
As in R3L~\cite{gu2026r3l}, we compare the evaluator of \cref{tab:quality} with the participants: Kendall's $\tau_b$ between the mean rating of a scene and its evaluator score is 0.17, 0.33 and 0.36 for realism, functionality and instruction following, and 0.53, 0.42 and 0.48 between two random halves of the raters of each scene.
Supplementary \cref{S-sec:supp_human} gives the rates per system with Wilson intervals and the ratings next to the evaluator scores.
Supplementary \cref{S-sec:supp_real_users} reports interactions with real users.

\section{Conclusion}
We studied personalized indoor scene synthesis under hidden user preferences, where a system learns what a person wants for a room from the person's reactions to the rooms it shows.
PreferenceScene keeps its estimate of the user's preferences as an explicit plan over a space of candidate values for every design attribute, revises the whole plan from sparse free-form feedback, and instantiates every value as its own part of the room.
On our benchmark, it matches hidden preferences better than five published text-to-scene baselines that receive the same feedback, and better than these systems when they are given the preferences directly.
Participants of a user study prefer its scenes to those of all baselines, for both preferences and design.

\paragraph{Limitations and future work.}
We restrict our study to one room and one user.
Whole homes, households of several people and changing preferences remain future work.

\section*{LLM Usage Statement}
Claude (Anthropic), used through Claude Code, assisted the authors with the experiments and the writing of this paper.
The authors take full responsibility for its content.

{
    \small
    \bibliographystyle{ieeenat_fullname}
    \bibliography{main_refs_fixed,supp_refs}%
}

\clearpage
\crefname{lstlisting}{Listing}{Listings}
\Crefname{lstlisting}{Listing}{Listings}
\maketitlesupplementary
\appendix
\setcounter{table}{0}\setcounter{figure}{0}\setcounter{algorithm}{0}\setcounter{equation}{0}
\renewcommand{\thetable}{S\arabic{table}}\renewcommand{\theHtable}{S\arabic{table}}
\renewcommand{\thefigure}{S\arabic{figure}}\renewcommand{\theHfigure}{S\arabic{figure}}
\renewcommand{\thealgorithm}{S\arabic{algorithm}}\providecommand{\theHalgorithm}{}\renewcommand{\theHalgorithm}{S\arabic{algorithm}}
\renewcommand{\theequation}{S\arabic{equation}}\renewcommand{\theHequation}{S\arabic{equation}}
\addtocontents{toc}{\protect\setcounter{tocdepth}{1}}
\setcounter{tocdepth}{1}
\supptoc

\providecommand{\sbChance}{21.4}
\providecommand{\sbDevProfilesRoom}{2}
\providecommand{\sbDevProfilesRoomW}{two}
\providecommand{\sbGroupAcc}{13}
\providecommand{\sbGroupFurn}{26}
\providecommand{\sbGroupLight}{10}
\providecommand{\sbGroupSpat}{10}
\providecommand{\sbGroupSurf}{30}
\providecommand{\sbGroupText}{11}
\providecommand{\sbHistFive}{40}
\providecommand{\sbHistFour}{31}
\providecommand{\sbHistSix}{24}
\providecommand{\sbHistThree}{5}
\providecommand{\sbInstRoom}{16}
\providecommand{\sbJudgeMaxSide}{1024}
\providecommand{\sbJudgeMaxTokens}{2}
\providecommand{\sbJudgeQuestionsScene}{20}
\providecommand{\sbJudgeRepo}{\texttt{zai-org/GLM-4.6V-Flash}}
\providecommand{\sbJudgeSeed}{0}
\providecommand{\sbJudgeTemp}{0}
\providecommand{\sbNAttrs}{100}
\providecommand{\sbNAttrsRoom}{10}
\providecommand{\sbNBases}{3}
\providecommand{\sbNBasesW}{three}
\providecommand{\sbNDev}{40}
\providecommand{\sbNDevProfiles}{20}
\providecommand{\sbNInst}{320}
\providecommand{\sbNInstForm}{160}
\providecommand{\sbNOrders}{2}
\providecommand{\sbNOrdersW}{two}
\providecommand{\sbNProfiles}{80}
\providecommand{\sbNRefScenes}{80}
\providecommand{\sbNRooms}{10}
\providecommand{\sbNTest}{120}
\providecommand{\sbNTestBoth}{240}
\providecommand{\sbNTestProfiles}{60}
\providecommand{\sbNTurns}{3}
\providecommand{\sbNValues}{483}
\providecommand{\sbNarrAskMax}{140}
\providecommand{\sbNarrAskMin}{90}
\providecommand{\sbNarrMax}{137}
\providecommand{\sbNarrMean}{115.9}
\providecommand{\sbNarrMin}{94}
\providecommand{\sbPersonaFields}{8}
\providecommand{\sbPersonasProfile}{2}
\providecommand{\sbPersonasProfileW}{two}
\providecommand{\sbProfDistMean}{8.4}
\providecommand{\sbProfDistMin}{5}
\providecommand{\sbProfilesRoom}{8}
\providecommand{\sbRenderH}{768}
\providecommand{\sbRenderW}{1024}
\providecommand{\sbRoomValMax}{55}
\providecommand{\sbRoomValMin}{40}
\providecommand{\sbScoreMax}{10}
\providecommand{\sbScoreMin}{1}
\providecommand{\sbStatedAttrMax}{8}
\providecommand{\sbStatedAttrMin}{8}
\providecommand{\sbStatedFive}{60}
\providecommand{\sbStatedFour}{50}
\providecommand{\sbStatedImplicitMax}{0}
\providecommand{\sbStatedMax}{6}
\providecommand{\sbStatedMean}{5.0}
\providecommand{\sbStatedMin}{4}
\providecommand{\sbStatedSix}{50}
\providecommand{\sbTestInstRoom}{12}
\providecommand{\sbUserBands}{5}
\providecommand{\sbUserBandsW}{five}
\providecommand{\sbUserMaxChars}{1200}
\providecommand{\sbUserMaxSide}{1024}
\providecommand{\sbUserMaxTokens}{480}
\providecommand{\sbUserMinChars}{10}
\providecommand{\sbUserMinP}{0}
\providecommand{\sbUserPresence}{1.5}
\providecommand{\sbUserRepo}{\texttt{Qwen/Qwen3.8-27B-FP8}}
\providecommand{\sbUserTemp}{0.7}
\providecommand{\sbUserTopK}{20}
\providecommand{\sbUserTopP}{0.8}
\providecommand{\sbValMax}{6}
\providecommand{\sbValMean}{4.8}
\providecommand{\sbValMin}{3}
\providecommand{\sbValuesUsed}{460}
\providecommand{\sbVisAcc}{95.6}
\providecommand{\sbVisAll}{396}
\providecommand{\sbVisAnswers}{6}
\providecommand{\sbVisAnswersW}{six}
\providecommand{\sbVisPass}{454}
\providecommand{\sbVisPassMin}{5}
\providecommand{\sbVisPassMinW}{five}
\providecommand{\sbVisScenes}{1179}

\section{Benchmark details}
\label{S-sec:supp_bench}

This section gives the details of the benchmark of \cref{sec:benchmark} of the main paper: an example instance (\cref{S-sec:supp_example}), the attribute schemas (\cref{S-sec:supp_schemas}), the profiles and the inputs (\cref{S-sec:supp_profiles}), the simulated user (\cref{S-sec:supp_user}), the visibility check (\cref{S-sec:supp_visibility}), the reference scenes (\cref{S-sec:supp_reference}) and the evaluator (\cref{S-sec:supp_judge}).
The listings after the references give the prompts verbatim, with their template placeholders.

\subsection{Example instance}
\label{S-sec:supp_example}
\begin{figure*}[t]
  \centering
  \includegraphics[width=\textwidth]{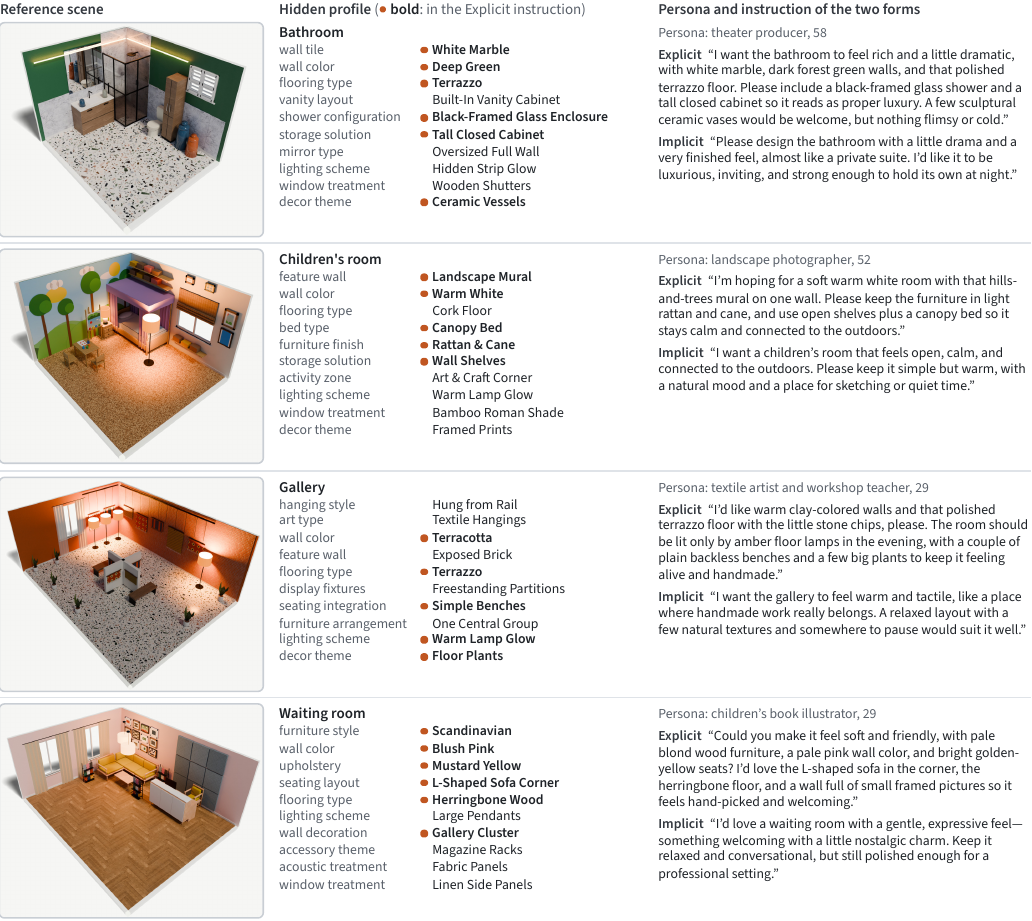}
  \caption{Example test profiles of the benchmark in four room types.
    Each row shows one profile: its reference scene, its ten hidden values and one persona with its \vE{} and \vI{} instructions.
    Bold values with a dot are stated in the \vE{} instruction.}
  \label{S-fig:bench_samples}
\end{figure*}

\Cref{S-fig:bench_example} shows one instance in both forms, \cref{S-fig:bench_samples} shows four test profiles with their reference scenes, and \cref{S-tab:bench_stats} gives the statistics of the benchmark.
\begin{figure}[t]
  \centering
  \scriptsize
  \setlength{\fboxsep}{3pt}
  \fbox{\begin{minipage}{0.95\columnwidth}
    \raggedright
    \textbf{Hidden profile} of a gallery ($\star$: stated in the \vE{} form):
    hanging style Hung from Rail, art type Black-and-White Photographs, wall color Gallery White$\star$, feature wall Wood Panelling, flooring type Large Format Tile$\star$, display fixtures Freestanding Partitions, seating integration Simple Benches$\star$, furniture arrangement Two Separate Groups$\star$, lighting scheme Picture Lights in the Dark$\star$, decor theme Floor Plants.\\[3pt]
    \textbf{Persona:} age 28, urban planner. ``He spends his days drawing, reviewing plans, and moving between meetings on foot or by bike.''\\[3pt]
    \textbf{\vE{} instruction:} ``I'd like bright white walls, large plain light tiles underfoot, and the furniture split into two groups with a clear aisle through the middle. Please keep the works lit by their own little lamps instead of daylight, with the benches plain and understated.''\\[3pt]
    \textbf{\vI{} instruction:} ``Please keep the gallery orderly and contemporary, with enough space that it feels mentally clear. I want it to be understated, thoughtful, and not overly decorated.''\\[3pt]
    \textbf{Taste narrative} (excerpt; given only to the simulated user): ``\dots{} The ceiling lights should be off, each work lit by its own small lamp, and several large potted plants should keep the space calm and alive.''
  \end{minipage}}
  \caption{One instance of the development subset in both forms. A system receives the persona and one of the two instructions. The simulated user receives the persona, the taste narrative and the rendered images.}
  \label{S-fig:bench_example}
\end{figure}

\subsection{Attribute schemas}
\label{S-sec:supp_schemas}
\paragraph{Attributes and values.}
The ten room types are the bathroom, the bedroom, the children's room, the dining room, the gallery, the kitchen, the library, the living room, the office and the waiting room.
Each room type has a schema of \sbNAttrsRoom{} attributes.
An attribute is one visible component of the room, and its values are mutually exclusive: a scene shows exactly one of them.
Every value has a name and a one-sentence description of what a viewer sees.
\Cref{S-lst:judge} shows the values of one attribute.
Every attribute has two texts.
Its \emph{definition} names the component that the attribute refers to and is quoted by the prompts of the benchmark construction and by the questions of the evaluator.
Its description in the \emph{attribute list} is one sentence that names no value, \eg ``The mirror type decides the mirror above the washbasin''.
This description is the only text about the attribute that a system receives (\cref{sec:problem} of the main paper).
\Cref{S-sec:supp_full} lists the attributes of every room type with their groups, their descriptions in the attribute list and the names of their values.

\paragraph{Statistics.}
The \sbNRooms{} schemas have \sbNAttrs{} attributes and \sbNValues{} values, \sbRoomValMin{} to \sbRoomValMax{} per room type.
An attribute has \sbValMin{} to \sbValMax{} values, \sbValMean{} on average, and \sbHistThree{}, \sbHistFour{}, \sbHistFive{} and \sbHistSix{} attributes have three, four, five and six values, respectively.
The attributes follow the groups of design elements in an interior design textbook~\cite{ching2012interior}: \sbGroupSurf{} attributes describe surfaces, \sbGroupLight{} lighting, \sbGroupFurn{} furniture, \sbGroupText{} textiles, \sbGroupAcc{} accessories and \sbGroupSpat{} the spatial arrangement.

\subsection{Hidden profiles and inputs}
\label{S-sec:supp_profiles}

\begin{table}[t]
  \centering
  \footnotesize
  \setlength{\tabcolsep}{3pt}
  \begin{tabularx}{\linewidth}{@{}l*{2}{Y}@{}}
    \toprule
    \multicolumn{3}{@{}l}{\emph{Shared by both forms}} \\
    Attributes & \multicolumn{2}{c}{$M=10$ in each of 10 room types; 100 in total} \\
    Values per attribute & \multicolumn{2}{c}{3--6 (4.8); 483 in total} \\
    Hidden profiles & \multicolumn{2}{c}{80 (8 per room type)} \\
    Instances per form & \multicolumn{2}{c}{160 (40 development / 120 test)} \\
    Taste narrative words & \multicolumn{2}{c}{94--137 (115.9)} \\
    \midrule
     & Explicit & Implicit \\
    Stated attributes & 4--6 (5.0) & 0 \\
    Instruction words & 37--76 (57.2) & 25--46 (33.3) \\
    Persona words & 126--380 (183.7) & 100--240 (161.4) \\
    \bottomrule
  \end{tabularx}
  \caption{Statistics of the benchmark. Each cell gives the range and, in parentheses, the mean. The word count of a persona excludes the name and the age.}
  \label{S-tab:bench_stats}
\end{table}

\paragraph{Profiles.}
Each room type has \sbProfilesRoom{} profiles, \sbNProfiles{} in total.
GPT-5.4-mini writes all profiles of a room type in a single call from its schema (\cref{S-lst:profile}).
For each profile, the model first chooses a design style and a target mood and then one value per attribute that fits both.
Any two profiles of a room type differ in at least \sbProfDistMin{} of the \sbNAttrsRoom{} attributes and in \sbProfDistMean{} on average.
The \sbNProfiles{} profiles use \sbValuesUsed{} of the \sbNValues{} values.

\paragraph{Personas.}
A persona has \sbPersonaFields{} fields: the name, the age, the occupation, the lifestyle, the personality traits, the living situation, the desires and the dislikes.
For each room type, one call of GPT-5.4-mini writes personas from the profiles (\cref{S-lst:persona}), and a second call assigns each persona to the profile that fits it best, so that every profile receives \sbPersonasProfileW{} personas.
Both calls also list profiles of kind ``mixed'', which combine the values of two profiles of the room type; these profiles and their personas are not part of the benchmark.
Each form therefore has \sbInstRoom{} instances per room type and \sbNInstForm{} in total.

\paragraph{Inputs of the two forms.}
The two forms share the profile, the taste narrative and the first six fields of the persona; they differ in the desires, the dislikes and the instruction (\cref{S-tab:bench_stats}).
In the \vE{} form, an instance states \sbStatedMin{} to \sbStatedMax{} attributes, and \sbStatedFour{}, \sbStatedFive{} and \sbStatedSix{} instances state four, five and six, respectively.
The stated attributes are balanced: every attribute is stated in \sbStatedAttrMin{} of the \sbInstRoom{} instances of its room type.
GPT-5.4-mini writes the desires, the dislikes and the instruction from the persona, the profile and the attributes to be stated (\cref{S-lst:explicit}).
The instruction states at least two of the stated values, and the desires or the dislikes state the others.
In the \vI{} form, the desires and the dislikes keep their text from the persona generation, and GPT-5.4-mini writes the instruction from the persona alone (\cref{S-lst:implicit}).
No attribute counts as stated.

\paragraph{Taste narrative.}
The simulated user is designed to know its taste as prose, not as schema values.
For every persona, GPT-5.4-mini therefore writes the taste narrative, in the persona's voice, from the persona and the \sbNAttrsRoom{} hidden values with their descriptions (\cref{S-lst:narrative}).
The prompt asks for \sbNarrAskMin{} to \sbNarrAskMax{} words, and the narratives have \sbNarrMin{} to \sbNarrMax{} words, \sbNarrMean{} on average.
The prompt asks the narrative to convey every hidden value and to show the overall style.
Only the simulated user reads the narrative.

\paragraph{Subsets.}
In every room type, \sbDevProfilesRoomW{} profiles form the development subset, which has \sbNDevProfiles{} profiles and \sbNDev{} instances per form.
The other \sbNTestProfiles{} profiles form the test subset, which has \sbNTest{} instances per form, \sbTestInstRoom{} per room type.

\subsection{Simulated user}
\label{S-sec:supp_user}
\paragraph{Model and inputs.}
The simulated user is Qwen3.8-27B (checkpoint \sbUserRepo), served with vLLM~\cite{kwon2023vllm}.
In every turn it receives one system message and one user message (\cref{S-lst:user}).
The system message holds the task description, the persona and the taste narrative.
The user message holds two rendered images of the shown scene, followed by the request for feedback.
The images are a cutaway view from above and an eye-level view, each of $\sbRenderW\times\sbRenderH$ pixels.
From the second turn on, the user message also lists the feedback and the score of every earlier turn and asks whether the parts of the room that the simulated user asked to change have improved.
The simulated user never receives the system's plan or the schema.

\paragraph{Output.}
The answer is a JSON object with two fields, and its format is enforced during decoding.
The first field is the feedback, a text of \sbUserMinChars{} to \sbUserMaxChars{} characters, and the prompt asks for three to five sentences.
The second field is the score, an integer from \sbScoreMin{} to \sbScoreMax{}, and the prompt describes the scale in \sbUserBandsW{} levels of two points each.
A system receives only the feedback and the score.

\paragraph{Sampling.}
The model samples with temperature \sbUserTemp{}, top-$p$ \sbUserTopP{}, top-$k$ \sbUserTopK{}, min-$p$ \sbUserMinP{} and presence penalty \sbUserPresence{}, in non-thinking mode, and writes at most \sbUserMaxTokens{} tokens.
The sampling seed of each answer is derived from the instance, the turn and the seed of the interaction.

\subsection{Visibility check}
\label{S-sec:supp_visibility}
The visibility check measures whether the values of the schemas can be read from rendered images.
For each room type we fix \sbNBasesW{} base plans, each of which assigns one value to every attribute.
For every base plan and every value, the rule-based construction of the benchmark builds the scene that has this value and, for the other attributes, the values of the base plan.
Every value is thus rendered in \sbNBasesW{} scenes, and the check has \sbVisScenes{} scenes in total.
Qwen3.8-27B receives the two rendered images of a scene and the multiple-choice question of \cref{S-sec:supp_judge} on the varied attribute, with the options in the given and in the reversed order, so that every value has \sbVisAnswersW{} answers.
We revised, merged or replaced values that the model named in less than 70\% of their answers and repeated the check on the revised schemas.
For the schemas of the benchmark, \sbVisAcc\% of the answers name the rendered value, and \sbVisPass{} of the \sbNValues{} values are named by at least \sbVisPassMinW{} of their \sbVisAnswersW{} answers.

\subsection{Reference scenes}
\label{S-sec:supp_reference}
For every profile, the rule-based construction of the benchmark builds one reference scene directly from the \sbNAttrsRoom{} hidden values, \sbNRefScenes{} scenes in total.
The construction differs from the scene instantiation of our method in how a value becomes a specification: a hand-written table per room type maps each schema value to fixed parameters, and no LLM or Jev call is made.
Our method never reads the schema values and obtains its specifications from Jev and from retrieval (\cref{S-sec:supp_real}).
Both build the room, check collisions and search the placement with the same code, draw objects from the same asset pool and render with the same lighting, white balance and cameras.
The instances of a profile share its reference scene in both forms.
\Cref{sec:benchmark} of the main paper reports the pAMR of the reference scenes over the \sbNTestBoth{} test instances of both forms.

\subsection{Evaluator and pAMR}
\label{S-sec:supp_judge}
\paragraph{Evaluator.}
The evaluator is GLM-4.6V-Flash (checkpoint \sbJudgeRepo), served with vLLM~\cite{kwon2023vllm}.
For a scene it answers one multiple-choice question per attribute (\cref{S-lst:judge}).
A question is one message that holds the two rendered images of the scene, followed by the question text.
The text names the attribute with its definition and lists the values of the attribute as lettered options, each with its description.
The answer is constrained to one option letter and is decoded at temperature \sbJudgeTemp{} in non-thinking mode.

\paragraph{pAMR.}
Every question is asked with the options in the given and in the reversed order, so a scene has \sbJudgeQuestionsScene{} answers.
pAMR (\cref{eq:pamr} of the main paper) is the share of these answers that name the hidden value.
A uniform random guess names the hidden value of an attribute with $n$ values with probability $1/n$.
The mean of $1/n$ over the \sbNAttrs{} attributes is \sbChance\%.
Because every room type has the same number of test instances, this mean is the pAMR of a random guess on the test subset.

\section{Method details}
\label{S-sec:supp_method}

This section gives the prompts, the assignment of design aspects to attributes, the settings of the LLM calls and an algorithm listing of \methodname{} (\cref{sec:method} of the main paper).

\subsection{Candidate values}
\label{S-sec:supp_cand}

One call of the LLM, Gemma-4-31B (\texttt{gemma-4-31b-it-qat}), writes the candidates before the first turn (\cref{S-lst:supp_cand}).
Its prompt contains the room type, the instruction $q$ and the eight fields of the persona $\mathcal{P}$, which every system receives, but not the attribute list.
Sampling uses temperature 1.0, top-$p$ 0.95 and top-$k$ 64, with thinking disabled and at most 14{,}000 output tokens.
The seed of the call is derived from the instance, the prompt and the seed of the interaction.
If the reply is not valid JSON, the call is not repeated; the aspects and candidates are read from the text by matching the field names of the requested format.

\paragraph{Assignment of design aspects to attributes.}
One request to Jev~\cite{typesafe2026jev} (\texttt{jev-1.13.0}) asks one multiple-choice question per design aspect $g_j$.
The options are the $M$ entries of the attribute list, each given as the attribute name and its description, and ``none''.
The question quotes the name of the aspect, the one-sentence description that the LLM wrote for it and its first three candidate labels.
Let $\lambda_j(a)$ be the probability that Jev returns for option $a$ of the question about $g_j$.
Aspect $g_j$ is assigned to the attribute $a=\arg\max_b\lambda_j(b)$ if $a$ is not ``none'' and $\lambda_j(a)\ge0.5$.
Each attribute $a$ then also receives the aspect $j^*=\arg\max_{j\notin A(a)}\lambda_j(a)$ if $\lambda_{j^*}(a)\ge0.3$, where $A(a)$ is the set of aspects already assigned to $a$.
Such an aspect can also belong to another attribute.
An attribute that has no aspect after these two rules receives the aspect with the highest $\lambda_j(a)$, if that probability is above 0.
The candidate set $C_a$ concatenates the candidates of the aspects in $A(a)$, at most 12 per aspect, with the most probable aspect first.
An attribute with fewer than two candidates is listed without candidates in the plan prompts (\cref{S-sec:supp_start}), and scene instantiation uses the default specification of its category (\cref{S-sec:supp_defaults}) unless the LLM writes a new value for it.

\subsection{Initial plan and plan update}
\label{S-sec:supp_start}

The calls for the initial plan and for every plan update send the LLM the same system message and a user message of the same form, which differs in the history, the task and one phrase of the answer format (\cref{S-lst:supp_start,S-lst:supp_update}).
The user message contains the input $c$ (\ie the room type, the instruction, the persona and every attribute name with its description) and the candidates of each attribute, grouped by aspect, each with its label, description and hex code.
The message states that an attribute whose candidates come from more than one aspect takes one value, and that an attribute without candidates keeps its default unless the LLM writes a new value.
From turn 2 on, the history lists, for every earlier turn, the value shown for each attribute with its scene description (\cref{S-sec:supp_descr}), the feedback and the score.
A new value from an earlier turn appears in the history with its label and description.
The history omits the analyses of the earlier replies.
The task of turn $k\le K$ asks for the plan of that turn, and the task after turn $K$ asks for the final plan, which receives no feedback.
Each call uses the sampling settings of the candidate call, at most 3{,}072 output tokens and a seed derived from the instance, the input, the seed of the interaction and the turn.

\paragraph{Reading a reply.}
A reply is valid if its JSON field ``plan'' gives, for every attribute, null, a non-empty text or an object with a non-empty label or description.
A text or an object label equal to a candidate label selects that candidate, unless the object gives a different description.
Null and texts such as ``default'' or ``none'' keep the default specification.
Any other text is a new value whose label and description are the text, and any other object is a new value with its label, description and hex code.
A text equal to the label of a new value from an earlier turn, or an object with the same label, description and hex code, selects that earlier value again, which keeps its specification.

\paragraph{Invalid replies.}
If a reply stops at the token limit, does not parse or is invalid, the request is repeated with a new derived seed, up to three requests per plan.
After three invalid replies, the first turn shows the first candidate of every attribute with candidates and the default for every other attribute, and a later turn repeats the previous plan.

\subsection{Furnishing the final scene}
\label{S-sec:supp_furnish}

The plan decides the look of the room, and scene instantiation places the pieces that the values name, so a room built from the plan alone contains few objects.
After the last turn, one more LLM call proposes further pieces for the final scene (\cref{S-lst:supp_furnish}).
Its prompt gives the values of the plan, the size of the room and the pieces that the room already contains.
It also lists the furniture tops that can carry objects, with their size, height and target number of objects, and the allowed object types, which are the type names that retrieval accepts.
The prompt asks for about ten floor pieces in the room, two to five objects on the top of every added piece with a top, at least six wall pieces and at least eight types among the added pieces.
It also asks that a table where people eat receive tableware and at most one centerpiece.
For an attribute of the decor, art or tabletop category, the reply first states whether its value asks for empty surfaces, names concrete pieces or does neither.
If the value asks for empty surfaces, neither objects on tops nor wall pieces are added.

\paragraph{Consistency with the plan.}
Before proposing pieces, the reply names, for every attribute, the kind of item that its value decides, such as the seats or the storage furniture, and lists the allowed types of that kind that the value does not name.
Every proposed piece of a listed type is dropped, together with the objects proposed for its top.
For this check, the prompt contains only the free-text values of the plan and the allowed type names, never the value sets $\mathcal{V}_a$ of the benchmark.
After these checks, a seat for a desk is added only with a desk and a television stand only with a television.\looseness=-1

\paragraph{Placement.}
The final plan is instantiated by the same procedure as the plans of the turns, and furnishing moves none of the resulting objects.
The added pieces are retrieved like all other objects (\cref{S-sec:supp_retrieval}) and placed in the remaining free space with the collision tests of the placement (\cref{S-sec:supp_place}).
Floor pieces are placed along free stretches of the walls and on free floor spots, objects on the free area of the tops, and wall pieces on free stretches of the walls.
A piece without a free place is not added.

\paragraph{Calls and replies.}
The furnishing call uses temperature 0, a JSON schema for the reply, a seed derived from the plan and at most 4{,}000 output tokens.
If the request fails four times, the final scene is delivered without additions.
The scenes shown in the turns, on which the user's feedback is based, are not furnished.
For the pAMR per turn (\cref{S-fig:supp_rounds,S-fig:supp_turns,S-tab:supp_open_jev,S-tab:supp_noisy}) and for the variant without iteration in \cref{tab:ablation} of the main paper, the scene of each turn is furnished as if it were delivered after~that~turn.

\subsection{Algorithm}
\label{S-sec:supp_alg}

\Cref{S-alg:turn} lists the steps of one interaction of \methodname{}.

\begin{algorithm}[t]
  \caption{One interaction of \methodname{}.}
  \label{S-alg:turn}
  \footnotesize
  \begin{algorithmic}[1]
    \Require room type, instruction $q$, persona $\mathcal{P}$, attribute list, number of turns $K$
    \State LLM call: 16 design aspects with 12 candidate values each, from the room type, $q$ and $\mathcal{P}$
    \State Jev assigns the aspects to the attributes, giving the candidate set $C_a$ of every attribute $a$
    \State LLM call: initial plan $\hat{S}_1$ from $c$ and $\{C_a\}$, with one value per attribute (a candidate, a new value or the default)
    \For{$k=1,\dots,K$}
      \State instantiate $\hat{S}_k$ as the scene $X_k$ (Jev answers the specification questions of every value shown for the first time) and render it
      \State receive the feedback $\eta_k$ and the score $z_k$
      \State write the scene descriptions $D_k$ of $X_k$
      \State LLM call: $\hat{S}_{k+1}$ from $c$, $\{C_a\}$ and $(\hat{S}_{k'},D_{k'},\eta_{k'},z_{k'})_{k'\le k}$ by \cref{eq:update} of the main paper
    \EndFor
    \State instantiate the final plan $\hat{S}_{K+1}$ as the final scene $X$
    \State LLM call: pieces to add to $X$ and, per attribute, the types that its value rules out (\cref{S-sec:supp_furnish}); drop the pieces of these types and place the rest without moving any object of $X$
  \end{algorithmic}
\end{algorithm}

\section{Scene instantiation details}
\label{S-sec:supp_real}

Scene instantiation sets every part of the room from the plan: the materials and colors of the walls and the floor; wall finishes such as panels, patterns and images; the lighting with its time of day, sources, power and color temperature; window coverings, rugs and fabrics; the furniture with its counts, sizes, materials and colors; the accessories; and the arrangement of the furniture.

\subsection{Component categories and questions}
\label{S-sec:supp_questions}

Each attribute belongs to one of 20 component categories.
Over the 100 attributes of the ten room types, the number of attributes per category is 10 for paint, floor, wall finish and light, 9 for decor, 7 for window, storage and seating, 5 for arrangement and style, 4 for fabric, 3 for fixtures and tables, 2 for beds, finish and art, and 1 for activity, tabletop, display and counter.
Each category defines the fields of the specification that the answers to its questions may set.
Every question about a candidate starts with ``A design for a(n) \{room\} describes its \{attribute\} (\{attribute description\}) under the heading `\{aspect\}' as: `\{label\}: \{description\}'.''
For a new value, the heading is the attribute name.
Most questions end with ``If the \{subject\} does not say, choose the option that fits it best.''
The time-of-day question allows dusk or night only when the text of the value names or implies that time, and lamp light alone implies no time.
\Cref{S-tab:supp_qplan} lists the questions per category, and \cref{S-tab:supp_options} the option sets with their sources.
All questions of the values first shown in a turn are sent to Jev in requests of at most 40 questions.
Each answer is the most probable option, with ties broken in favor of the first option.
The probabilities of the placement answers are kept for the placement search (\cref{S-sec:supp_place}).

\begin{table}[t]
  \centering
  \footnotesize
  \begin{tabularx}{\linewidth}{@{}>{\raggedright\arraybackslash\hsize=.77\hsize}X>{\raggedright\arraybackslash\hsize=1.23\hsize}Xr@{}}
    \toprule
    Category & Questions & Count \\
    \midrule
    paint, fabric, finish, style, counter & none (retrieval only) & 0 \\
    wall finish & cover, pattern & 2 \\
    floor & cover, pattern, rug & 3 \\
    window & form, opacity, length, rise & 4 \\
    light & time, source, color temperature, power, lamps, daylight, count of 4 lamp types, cove light, picture light & 12 \\
    arrangement & the 7 placement questions, per group, mix and, where they apply, procedural parameters & 9--12 \\
    object categories & a count for each pool type, size, same or mixed pieces, supporting surface and, where they apply, the 7 placement questions, per group for seating and procedural parameters & 93--105 \\
    \bottomrule
  \end{tabularx}
  \caption{Questions per value by component category. Pool types are the 90 type names of the asset pool.}
  \label{S-tab:supp_qplan}
\end{table}

\begin{table}[t]
  \centering
  \footnotesize
  \begin{tabularx}{\linewidth}{@{}>{\raggedright\arraybackslash\hsize=.85\hsize}X>{\raggedright\arraybackslash\hsize=1.3\hsize}X>{\raggedright\arraybackslash\hsize=.85\hsize}X@{}}
    \toprule
    Question & Options & Source \\
    \midrule
    count per type & none, one, two, three or four, five or more, some & pool type names; bins of this work \\
    size & small, standard, large ($\times$0.8, 1.0, 1.25 of the median pool size) & this work \\
    anchor, grouping, along, symmetry, facing, spacing & back, left or right wall, the wall with the largest window, or the middle; edge or middle; 0.2/0.5/0.8; yes/no; four facings; three spacings & Holodeck constraints~\cite{yang2024holodeck}; bins of this work \\
    groups, per group, mix & 1--3 groups; 1, 2, 3--4, 5--9, 10+; same or mixed & quantity and variance of Holodeck~\cite{yang2024holodeck}; bins of this work \\
    color temperature & 2700--6500\,K, six values & ANSI C78.377 nominal values \\
    time, source & day, dusk, night; daylight, electric & ProcTHOR~\cite{deitke2022procthor} \\
    power, lamps, daylight & three bins; yes/no; three bins & this work \\
    window covering & none, curtain, blind, shade, shutter; sheer, light, blackout; sill, floor; fold, roll & this work \\
    wall cover, pattern & paint, material, relief, panels, pattern, image; nine pattern kinds & this work \\
    procedural parameters & 14 room-specific sets in nine room types & parameters of the procedural parts \\
    \bottomrule
  \end{tabularx}
  \caption{Option sets of the specification questions and their sources.}
  \label{S-tab:supp_options}
\end{table}

\paragraph{Count questions.}
A count question reads ``How many \{type plural\} does the \{subject\} name for the \{attribute\}?''
The question also lists the five nearest pool types by Sentence-BERT~\cite{reimers2019sentencebert} similarity, and states that a piece must not be counted when one of these types names it better.
It further states that a word that gives only a shape, a material, a use or a part of another object names no object.
If every count answer of a value of an object category is ``none'', a second request asks which pool type is the main piece of the described attribute, and that type counts as one piece.

\subsection{Retrieval}
\label{S-sec:supp_retrieval}

\paragraph{Objects.}
We use the retrieval of Holodeck~\cite{yang2024holodeck} over the 6{,}793 Objaverse~\cite{deitke2023objaverse} assets that the baselines use.
The query is ``a 3D model of \{type\}, \{text\}'', where the text is the label and the description of the value.
The score of an asset is $100\max_{v}\cos(\mathrm{CLIP}_{\mathrm{text}},\mathrm{CLIP}_{\mathrm{image},v})+\mathrm{SBERT}(\text{text})\cdot\mathrm{SBERT}(\text{caption})-10\,\overline{|d-d'|}$.
The maximum is taken over three views $v$ of the asset, and $\overline{|d-d'|}$ is the mean absolute difference in meters between the sorted dimensions of the asset and of the target.
The target size is the median pool size of the type times the size option.
The encoders are OpenCLIP~\cite{ilharco2021openclip} ViT-L/14 trained on LAION-2B and Sentence-BERT all-mpnet-base-v2, as in Holodeck.
Unlike in Holodeck, retrieval is restricted to the pool type that the count answer of the specification names.
Assets below Holodeck's CLIP score threshold are ranked last instead of dropped.
The assets are selected by a deterministic top-$k$ instead of being sampled from the top ten, and ties are broken deterministically.
Holodeck's room-size and floor-placement filters are not used.
When a furniture piece is instantiated, its query also contains the texts of the style and furniture-finish attributes and, for upholstered types, of the fabric attribute, and assets used elsewhere in the scene are excluded.

\paragraph{Surfaces and colors.}
For a material query, each material is scored by $\cos(\mathrm{CLIP}_{\mathrm{text}},\mathrm{CLIP}_{\mathrm{swatch}})$ plus the string similarity between the query and the material name, as in Holodeck.
The floor and wall-finish categories search all 236 ambientCG~\cite{ambientcg} materials, and the other categories search only the materials for their kind of surface.
A color is the hex code of the value if the LLM wrote one, and otherwise the nearest of the 148 named colors of Matplotlib~\cite{hunter2007matplotlib} by CLIP text similarity, as in Holodeck.
A retrieved material is tinted toward the color in HSV (linear RGB), with the hue shift clipped to $\pm0.15$ of the hue circle and the scale factors of saturation and value clipped to $[0.3,1.6]$ and $[0.5,1.5]$.
The material of a kitchen counter is chosen again from the text of the counter value, by the CLIP term alone over all 236 materials, and a metallic material is rendered as metal.
Because a chromatic wall color with CIELAB $L^*$ below 25 shows no visible hue in the rendered images, it is replaced by another color that its text names, if that color has a hue within $60^\circ$ of the wall color and is lighter by more than 10 in $L^*$.

\paragraph{Image covers.}
A wall-finish specification with the cover ``image'' takes the canvas texture of the painting, among the 111 paintings of the asset pool, whose CLIP similarity to the text of the value is highest.
The canvas is center-cropped to the aspect ratio of the wall.
No image is generated.

\subsection{Procedural furniture library}
\label{S-sec:supp_proc}

Furniture of 27 pool types, in the six groups seating, chairs, tables, storage, shelving and beds, is rebuilt by our procedural library inside the bounding box of the retrieved asset, at its position and yaw (\cref{sec:method_repr} of the main paper).
Jev answers questions about the parameters of the library.
The material of each part, such as the frame, the top, the fronts or the upholstery, is retrieved from ambientCG with the text of the attribute, by the CLIP term alone within the kind of surface and the material families of the part.
The hex codes of the specification set the colors of the parts.
A color that the CLIP color search infers for a text without a color name is not applied to fabric parts.
On wood and metal parts, a color whose hue the tint cannot reach is dropped, because a clipped hue shift would show a third color that neither the material nor the value has.
Leather upholstery keeps the roughness map of its material, and a part whose material is metallic is rendered as metal.
A procedural piece whose mesh differs from its bounding box by more than 5\,mm along an axis is refitted, and the retrieved asset is kept if no refit succeeds.
Procedural parts also provide the object types that the pool lacks: kitchen counter runs with a sink and a hob, the island block, wall cabinets and the range hood; the shower tray, screen and bench of the bathroom; built-in wardrobe runs in the bedroom and the children's room; plinths and free panels in the gallery; screens and built-in shelving in the library; and built-in benches in the dining room and the kitchen.
Window coverings, wall slats, raised panels, flat pattern surfaces, rugs, cove light strips and picture lights are procedural parts shared across room types.

\subsection{Placement}
\label{S-sec:supp_place}
\paragraph{Difference from Holodeck.}
In Holodeck~\cite{yang2024holodeck}, an LLM writes constraints between pairs of objects, and a depth-first search on a grid places one object at a time for 30\,s and keeps the highest-scoring layout.
Our soft constraints are the placement answers of a specification, which describe a group of furniture.
The search enumerates combinations of these answers in the order of their probabilities, lays out each combination by fixed rules and keeps the first feasible layout.
Wall pieces, objects on furniture tops and ceiling lights are placed by fixed rules instead of Holodeck's LLM-written height constraints and random placement on supports.

Each room type defines required pieces, which are added when no value names their type, and default seats, which are added when no seating value names a seat.
It also defines the type of the central piece of a furniture group and neutral placement answers, which start the search and are its fallback.
In the living room, these are a coffee table and a side table, a sofa and two armchairs, the coffee table, and one symmetric group in the middle of the room.
The floor furniture follows the specification of the arrangement attribute, if any, otherwise that of the seating attribute, otherwise the neutral answers.
The search ranks the options of the anchor, grouping, number of groups, facing, symmetry, spacing and position along the wall by the answer probabilities of Jev.
It tries the three most probable options of each parameter in at most 400 trials, ordered by the soft score $\sum\log(p+10^{-3})$, and it varies last the parameters whose answer is confident ($p>0.5$).
The hard constraints keep every piece inside the room and exclude collisions.
To keep the door, windows and curtains clear and the eye-level view open, they also exclude pieces from the clearance zones of the door (1.0\,m deep and 0.2\,m wider than the door) and of the windows (0.35\,m deep above the sill) and from the band of the curtains, and floor furniture from within 1.2\,m of the eye-level camera.
A piece that stands against a wall is placed on a wall run, \ie a free stretch of wall at least 0.1\,m from a door, 0.22\,m from a window and 0.05\,m from the end of the wall.
If every trial fails and some trial fails only because floor pieces find no wall run, the walls are too short for the requested pieces.
The search then removes the unplaced pieces of the trial with the fewest of them and runs again before it falls back to the neutral answers.
For pieces at least 1.2\,m tall, the walls that the cutaway view does not hide are tried first.
Tables and the pieces of the activity, display and storage attributes are then placed according to their own placement answers, and small objects are placed in ceiling rows, on walls, on supports or on free floor spots.
An arrangement specification places at least the number of pieces per group times the number of groups, with at most 18 on walls and 12 on the floor.
To fill the wall, a dense wall hang of ten or more pictures requests 24 pictures of mixed sizes, which are packed over the height band of the wall.
When the floor placement asks for $k\ge2$ groups in the middle of the room, each group receives its own low table, the groups are spread over 0.8 of the room width, and a work table such as a reception desk is placed at a wall.
A floor layout with $k\ge2$ groups and no group size requests $k$ times the named seats, and seats that face each other are placed on both long sides of the low table of their group.
A symmetric seating, display or storage specification with one named piece requests a matched pair.
Because a one-wall kitchen needs a long free wall, its run moves to the visible wall with the longest free stretch if its own wall has a free stretch shorter than half the wall length and the other stretch is at least 0.5\,m longer.

\subsection{Lighting and windows}
\label{S-sec:supp_light}

Every room is lit by the Poly Haven~\cite{polyhaven} environment map ``reading room''.
At dusk and at night, the environment light and the daylight are reduced, the lamp power doubles, every window covering is closed, and the ceiling light is dimmed to 40\,W if the lighting text names lamps or hidden lights but no ceiling lamp.
Lamps are lit at dusk and at night, and at any time when the lighting text says that lamps are lit, names fixtures or hidden lights, or contains a glow word, such as candlelight, or a fixture word, such as pendant.
A lighting text describes an evening interior, \ie a lamp-lit room, if its main source is electric, its daylight bin is not high, it names no daylight and it names fixtures that are not only ceiling lamps, and if it contains an evening cue: dusk or night as the time, a lamp or glow word, or named hidden lights.
Such a text receives the evening lighting: no daylight, closed window coverings, lamps at double power and at the color temperature of the text, and a ceiling light of at least 200\,W.
If a lighting text says that lamps are lit, contains a glow word or describes an evening interior but lists no fixture and no hidden light, two lamps of each lamp type that it names are added, or else two floor lamps and two table lamps.
Because the point light of a lamp is invisible to the camera, each lit lamp from the asset pool receives an emitting sphere of radius 5\,cm below its light source, colored by the hex code or the last color word of the lighting text.
The rendered images have no camera white balance (\cref{S-sec:supp_render}), so a warm ceiling light would tint the surfaces orange; a ceiling light of color temperature $T<T_0=5000$\,K is therefore rendered at the temperature $T'$ with $1/T'-1/T_0=0.4\,(1/T-1/T_0)$.
Fabric window coverings take a retrieved fabric.

\subsection{Instantiation rules}
\label{S-sec:supp_rules}

The variant without the instantiation rules (\cref{S-sec:supp_more_ablations}) removes the following rules, which the named subsections state~in~full.
\begin{itemize}
  \item Lighting (\cref{S-sec:supp_light}): lamps lit by named fixtures, hidden lights and glow or fixture words; the evening lighting; the added lamps; the emitting spheres; and the shift of ceiling light below 5000\,K toward 5000\,K.
  \item Materials (\cref{S-sec:supp_proc,S-sec:supp_retrieval}): the choice of the material of each procedural part by the CLIP term alone and the second choice of the kitchen counter material from its text; the exclusion of inferred colors from fabric parts and of unreachable hues from wood and metal parts; the roughness map of leather; and the rendering of metallic materials as metal.
  \item Object counts and placement (\cref{S-sec:supp_place}): the counts of pictures in a dense wall hang and of tables, seats and matched pieces; the spreading of seat groups with a work table at a wall; and the move of a one-wall kitchen run.
  \item Colors (\cref{S-sec:supp_retrieval}): the replacement of dark chromatic wall colors.
\end{itemize}

\subsection{Default specifications}
\label{S-sec:supp_defaults}

An attribute that the plan leaves open, or a field that no answer sets, takes a fixed default that depends only on its category.
The defaults are the paint color \#e9e5dd, the ambientCG floor material WoodFloor051, painted walls, the fabric color \#b9b2a6, a 3000\,K light of medium power with lamps lit and daylight, a light floor-length curtain in \#ece8e0, and the neutral placement answers of the room type (\cref{S-sec:supp_place}).

\subsection{Scene descriptions}
\label{S-sec:supp_descr}

After a plan is instantiated, templates describe each attribute from its specification, the placement that the search chose, the captions of the retrieved assets, the names and tags of the ambientCG materials, and color names.
For a lighting attribute, the description states the color temperature, the power, the time of day, the fixtures, the lit lamps and whether daylight enters.
The description never copies the words of the value.

\subsection{Examples of specifications, scene descriptions and furnishing lists}
\label{S-sec:supp_examples}

\paragraph{Specifications.}
\Cref{S-lst:ex_spec} shows the specifications of three values of the plan of turn 1 in the interaction of \cref{fig:overview} of the main paper (a dining room, \vI{} form, seed 1).
A specification states, for one value, what scene instantiation builds.
The field \texttt{category} is the component category of the attribute (\cref{S-sec:supp_questions}), and \texttt{label} and \texttt{look} are the text of the value.
The field \texttt{colors} holds the hex codes of the value, and \texttt{surfaces} holds the ambientCG material retrieved for a surface and whether the surface is one material or a pattern.
The field \texttt{objects} lists the object types of the value with their number, the surface they stand on and a size in centimeters.
The field \texttt{placement} holds the placement answers: the wall that the group stands against or the middle of the room (\texttt{anchor}), the position along that wall, the grouping, the number of groups (\texttt{k}), symmetry, facing and spacing (\cref{S-sec:supp_place}).
The field \texttt{light} holds the color temperature in kelvin, the power, whether lamps are lit, the daylight, the time of day and the light fixtures.
All fields except \texttt{label}, \texttt{look} and \texttt{colors} come from the answers of Jev or from retrieval.

\paragraph{Scene descriptions.}
\Cref{S-lst:ex_descr} shows the scene descriptions $D_1$ of the same three attributes, which the LLM reads in the update after turn 1 (\cref{eq:update} of the main paper).
Templates write them from what was built: for the floor, the ambientCG material with its tags and the color; for the lighting, the settings of the light; for the seating, the retrieved pieces with the materials of their parts, their size and their placement.
The flooring value asks for black marble with white veins, and its description reports the retrieved material, Marble 006, with tags such as black, dark and grey.
The LLM therefore reads what the user saw and not what the value asks for.

\paragraph{Furnishing lists.}
\Cref{S-lst:ex_avoid} shows part of the reply of the LLM call that furnishes the final scene of the second user of \cref{fig:teaser} of the main paper, a dining room in the \vE{} form with seed 3; \cref{fig:overview} of the main paper shows the \vI{} form of the same user.
For every attribute, the reply first names what its value governs and lists the allowed object types that the value rules out (\cref{S-sec:supp_furnish}).
The storage value ``Black Lacquer Vitrine'' rules out other storage furniture, so the sideboard that the same reply proposes is dropped, together with the four objects proposed for its top.
The lighting value rules out no type.

\subsection{Rendering}
\label{S-sec:supp_render}

Blender Cycles~\cite{blender2025} renders a cutaway view from above and an eye-level view at $1024\times768$ pixels with at most 128 samples per pixel.
The scenes of every system are rendered with the same camera settings: the same background color seen by camera rays, no exposure change and no camera~white~balance.

\section{Experimental details}
\label{S-sec:supp_exp}

\subsection{Models and runs}
\label{S-sec:supp_models}

\Cref{S-tab:roles} lists the model of each role.
In the main evaluation, each role uses a different model family, and \methodname{} never calls the models that simulate the user, evaluate the scenes or generate the benchmark texts.
Every system runs each of the 120 test instances of each form with seeds 0--3.
The seed sets the sampling seed of every call to the simulated user and to the LLM of the system.
All interactions have $K=3$ turns: a system shows three scenes, receives the feedback and the score for each, and outputs its final scene, which receives no feedback.
The scenes of all systems are rendered by the same renderer from the same two cameras (\cref{S-sec:supp_render}).

\begin{table}[t]
  \centering
  \footnotesize
  \setlength{\tabcolsep}{3pt}
  \begin{tabularx}{\linewidth}{@{}lX@{}}
    \toprule
    Role & Model \\
    \midrule
    Benchmark texts & GPT-5.4-mini \\
    Simulated user & Qwen3.8-27B (open weights) \\
    Evaluator & GLM-4.6V-Flash (open weights) \\
    LLM of every system & Gemma-4-31B (open weights) \\
    Multiple-choice decisions & Jev (\texttt{jev-1.13.0})~\cite{typesafe2026jev} \\
    \midrule
    Second evaluator & Ministral-3-14B \\
    Hosted LLM & GPT-6.1 Terra \\
    \bottomrule
  \end{tabularx}
  \caption{Models and their roles. The multiple-choice decisions are made inside \methodname{}. The last two rows are used in \cref{S-sec:supp_judges,S-sec:supp_llm}, respectively. Ministral-3-14B is also the labeller of the attributes named in the feedback (\cref{S-sec:supp_named}).}
  \label{S-tab:roles}
\end{table}

\subsection{Adaptation of the baselines}
\label{S-sec:supp_base}

We run Holodeck~\cite{yang2024holodeck}, LayoutGPT~\cite{feng2023layoutgpt}, I-Design~\cite{celen2025idesign}, LayoutVLM~\cite{sun2025layoutvlm} and R3L~\cite{gu2026r3l} from their released code and adapt them as follows.

\paragraph{Common changes.}
Gemma-4-31B replaces the LLM of every baseline, with the sampling temperature of the baseline's released configuration: 1.0 for Holodeck and R3L, and 0.7 for LayoutGPT, I-Design and LayoutVLM, except for one agent of I-Design, which uses 0.
Every baseline uses the room shell of the instance's room type, which consists of the floor polygon, wall height, door and window, and takes its objects from the pool of 6{,}793 Objaverse~\cite{deitke2023objaverse} assets.
We convert the output of every baseline to our scene format and render it with our renderer.
Because none of the baselines sets the lighting, the renderer uses one fixed lighting setup for their scenes: a ceiling light of 200\,W at 3500\,K, a window light of 250\,W at 6000\,K and an environment map.\looseness=-1

\paragraph{What the systems set from the text.}
Holodeck sets the objects, their placement, and the wall and floor materials, which it selects from a fixed library.
LayoutGPT, I-Design, LayoutVLM and R3L set the objects and their placement.
None of these five baselines decides the lighting, the window coverings or the textiles of a room.
Lamps, curtains and textiles appear in their scenes only as assets of the pool that they place.

\paragraph{Input text.}
In turn 1, a system receives the room type, the instruction, the persona and the attribute list, which is added as the line ``Design attributes of this room (attribute: what it determines):'' followed by one line per attribute with its description.
In every later turn and for the final scene, the text also contains one block per earlier turn with a description of the scene shown, the feedback and the score.
The description, generated from the system's output, lists the wall and floor materials and the object names with their counts.
In the oracle condition, the text of turn 1 is followed by one line per attribute with the name and the description of its hidden value.

\paragraph{Holodeck.}
Holodeck selects wall and floor materials by text descriptions.
We map each selected material to one of our 236 ambientCG~\cite{ambientcg} materials by CLIP similarity to Holodeck's material text, and tint the mapped material toward the color of Holodeck's material, within the tint bounds of the mapped material and with a hue shift of at most $\pm0.25$ of the hue circle; a hue outside this limit is not shifted.

\paragraph{LayoutGPT.}
LayoutGPT has in-context exemplars for bedrooms and living rooms only, so it runs on the 24 test instances of these two room types in each form.
LayoutGPT outputs object boxes without materials.
Each box takes the pool asset of its category that fits its size, and the walls and the floor have fixed materials.

\paragraph{I-Design.}
I-Design chooses no wall or floor materials and takes the walls and the floor of the Holodeck scene of the same input and seed.
Before I-Design builds its scene graph, we remove the entries of an LLM reply that name a wall, the ceiling or the middle of the room as an object.
We limit its backtracking solver to 300\,s and each call of its edge extraction to 60\,s.
A generation that exceeds either limit counts as failed.

\paragraph{LayoutVLM.}
LayoutVLM takes the walls and the floor of the Holodeck scene of the same input and seed.
A generation whose constraint program cannot be solved counts as failed.\looseness=-1

\paragraph{R3L.}
R3L receives the room shell as its room, places floor objects, and takes the walls and the floor of the Holodeck scene of the same input and seed.

\paragraph{Holodeck-Edit.}
In turn 1, Holodeck-Edit shows the scene of Holodeck.
After turn 1, one LLM call per turn receives the input text of the published baselines, the floor design, the wall design and the object list of the last scene shown, and the request to keep every part that the feedback does not ask to change.
The furniture of unchanged entries and the small objects on it keep their assets, and Holodeck's solver places all objects again.

\paragraph{Failed generations.}
Requests that a system repeats inside one generation, such as the repeated LLM requests of \methodname{} (\cref{S-sec:supp_start,S-sec:supp_furnish}), are part of the procedure of the system.
We do not run a failed generation again.
The system keeps its previous scene.
If its first generation fails, it places the selected assets at random, as in LayoutVLM~\cite{sun2025layoutvlm}, or shows the empty room if no asset was selected.
The same rule applies to \methodname{}, which keeps the previous plan and scene when a turn fails.
A generation that an infrastructure fault, such as a lost connection, interrupts does not count as failed and is restarted with the same seed.
\Cref{S-tab:supp_success} gives the success rate of every system.

\subsection{Systems that we do not run}
\label{S-sec:supp_notrun}

We do not run SceneSmith~\cite{pfaff2026scenesmith} and SceneWeaver~\cite{yang2025sceneweaver}, because their papers report about 3.4\,h and 64\,min per scene, respectively, and one interaction requires $K+1=4$ generations.
Methods trained on 3D-FRONT~\cite{fu20213dfront}, such as ATISS~\cite{paschalidou2021atiss}, DiffuScene~\cite{tang2024diffuscene} and InstructScene~\cite{lin2024instructscene}, cover at most four of our ten room types.

\subsection{Metrics}
\label{S-sec:supp_metrics}

\paragraph{Preference match.}
pAMR (\cref{eq:pamr} of the main paper) is computed by the evaluator as described in \cref{S-sec:supp_judge}.

\paragraph{Scene quality.}
The evaluator scores \emph{Real.}, \emph{Func.} and \emph{Instr.} with the unchanged evaluation prompt of R3L~\cite{gu2026r3l} on the top-down and side views that R3L gives its evaluator.
The instruction field of this prompt holds the text of turn 1 without the attribute list: the room type, the instruction and the persona.
The prompt contains no system name, plan or feedback.
As in R3L, each scene is scored five times and the scores are averaged.
These five queries sample at temperature 1.0 and top-$p$ 1.0.
\emph{OOB} is the out-of-bound ratio of R3L, the percentage of objects whose footprint lies outside the room polygon by more than $0.0003$\,m$^2$.
\emph{Coll.} is the percentage of objects that collide with another object under the mesh-intersection test of SceneEval~\cite{tam2026sceneeval}: two objects collide if their meshes intersect and still intersect after one of them is moved 5\,mm away from the contact points.
Floor coverings such as rugs are not tested, as SceneSmith~\cite{pfaff2026scenesmith} excludes carpets.
The light emitters that scene instantiation places inside lit lamps (\cref{S-sec:supp_light}) are part of their lamp and are not counted as objects in \emph{Coll.}

\paragraph{Cost.}
The LLM calls and tokens of an interaction are counted over all its $K+1$ generations.
For \methodname{}, they include the furnishing call of the final scene (\cref{S-sec:supp_furnish}).
The generation time per scene is the wall time of a system's own steps, including its model calls and the building of the scene but not the rendering, divided by the $K+1$ scenes of an interaction.
For \methodname{}, these steps include the furnishing of the final scene.
The generation time per turn is the wall time from the feedback of turn $k$ to the rendered images of the next scene, averaged over $k=1,\dots,K$.
It thus includes the rendering and excludes the generation of the scene of turn~1.
A scene equal to a scene rendered before is not rendered again, and its images are read from a cache.
Neither time includes the simulated user.
The timing run executes the steps of every system, one interaction at a time, on one host with two Intel Xeon Gold 6526Y CPUs (64 threads).
The LLM runs on one Gemma-4-31B server with 4-bit weights, vLLM 0.32.0 and one NVIDIA H100 GPU, and the rendering on NVIDIA RTX 4090 and H100 GPUs.

\paragraph{Statistics.}
Every metric is first averaged over the test interactions of each seed.
The tables give the mean and the standard deviation over the four seeds, or over the seeds that their captions name.
The oracle rows and the row with the hosted LLM use seed 0 only.
A difference between \methodname{} and another row is computed over the pairs of instance and seed that both rows contain.\looseness=-1

\section{Additional results}
\label{S-sec:supp_results}

\subsection{Match per turn}
\label{S-sec:supp_turns}
\begin{figure}[t]
  \centering
  \includegraphics[width=0.88\columnwidth]{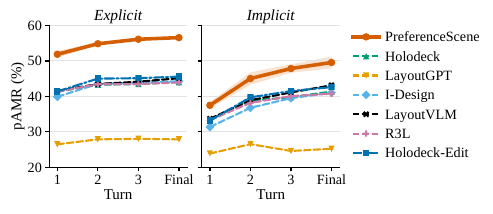}
  \caption{pAMR of the scene of each turn and of the final scene for \methodname{} with the scene of every turn furnished (\cref{S-sec:supp_furnish}), the published baselines with feedback and Holodeck-Edit.
    Lines: mean over four seeds; band: standard deviation of \methodname{}. LayoutGPT covers only bedrooms and living rooms.}
  \label{S-fig:supp_rounds}
\end{figure}

\Cref{S-fig:supp_rounds} shows the pAMR of the scene of each turn and of the final scene for all systems, and the main paper gives the values of \methodname{}.
In the \vI{} form, where the feedback is the main evidence, the gain of \methodname{} from the scene of turn 1 to the final scene (\eOneGainOursI{} points) exceeds the gain of every baseline on the same interactions: the difference between the two gains lies between \eOneDidLoI{} and \eOneDidHiI{} points, and every interval lies above 0.
Each difference is computed on the triples of instance, seed and attribute that both systems have, averaged over the triples of each seed and over seeds 0--3, with a 95\% percentile bootstrap interval over instances (\suppBootB{} resamples).\looseness=-1

\subsection{Results per room type}
\label{S-sec:supp_rooms}

On the test instances (12 per room type and form), \methodname{} has the highest pAMR of the final scene, averaged over four seeds, in all ten room types in the \vE{} form and in nine of the ten in the \vI{} form.

\subsection{Attributes named in the feedback}
\label{S-sec:supp_named}

For the counts in the paragraph ``The role of feedback'' of the main paper, an LLM labeller, Ministral-3-14B (\texttt{Ministral-3-14B-Instruct-2512}, temperature 0), reads the feedback of every turn of the interactions of \cref{tab:main} of the main paper.
It does not see the scene.
For each of the ten attributes of the room type, given with its options, it answers whether any part of the feedback concerns the attribute, as praise, a complaint, a wish or a question, and it quotes the shortest part of the feedback that supports the answer.
An attribute is named in a turn when the labeller answers yes, and it is never named in an interaction when no turn of the interaction names it.

\subsection{Further variants of the baselines}
\label{S-sec:supp_variants}

\Cref{S-tab:supp_variants} reports the published baselines in three further variants: with the feedback history but without the attribute list, with the attribute list but without feedback, and with neither.
A variant without feedback generates one scene from its input of turn 1.
The feedback history raises the pAMR of every published baseline, with and without the attribute list.
Over the ten rows with feedback of a form, the difference between a row and the same variant without feedback on the same interactions ranges from \trGainFbLayoutgptE{} to \trGainFbIdesignE{} points in the \vE{} form and from \trGainFbsLayoutgptI{} to \trGainFbsIdesignI{} points in the \vI{} form.
The attribute list changes the pAMR by less than 2 points.
\methodname{} is above every variant of every published baseline in both forms.
In the oracle condition, a system receives the hidden preferences as text in addition to its input of turn 1 and generates one scene.
The \vE{} values of this condition repeat the oracle column of \cref{tab:main} of the main paper.
Each $\Delta$ uses the interactions that both rows have: seed 0 for the oracle rows, and bedrooms and living rooms for LayoutGPT.

\newcommand{\varrow}[3]{#1 & \csname trMain#2#3E\endcsname & \csname trPair#2#3ED\endcsname & \csname trMain#2#3I\endcsname & \csname trPair#2#3ID\endcsname \\}
\begin{table}[t]
  \centering
  \footnotesize
  \setlength{\tabcolsep}{3.5pt}
  \begin{tabularx}{\linewidth}{@{}l*{4}{Y}@{}}
    \toprule
    & \multicolumn{2}{c}{\vE{}} & \multicolumn{2}{c}{\vI{}} \\
    \cmidrule(lr){2-3}\cmidrule(l){4-5}
    Method & pAMR & $\Delta$ & pAMR & $\Delta$ \\
    \midrule
    \multicolumn{5}{@{}l}{\emph{Attribute list and feedback (as in the main paper)}} \\
    Holodeck & \trMainFbsHolodeckE & \trPairFbsHolodeckED & \trMainFbsHolodeckI & \trPairFbsHolodeckID \\
    LayoutGPT & \trMainFbsLayoutgptE & \trPairFbsLayoutgptED & \trMainFbsLayoutgptI & \trPairFbsLayoutgptID \\
    I-Design & \trMainFbsIdesignE & \trPairFbsIdesignED & \trMainFbsIdesignI & \trPairFbsIdesignID \\
    LayoutVLM & \underline{\trMainFbsLayoutvlmE} & \trPairFbsLayoutvlmED & \underline{\trMainFbsLayoutvlmI} & \trPairFbsLayoutvlmID \\
    R3L & \trMainFbsRThreelE & \trPairFbsRThreelED & \trMainFbsRThreelI & \trPairFbsRThreelID \\
    \midrule
    \multicolumn{5}{@{}l}{\emph{Feedback, without the attribute list}} \\
    Holodeck & \trMainFbHolodeckE & \trPairFbHolodeckED & \trMainFbHolodeckI & \trPairFbHolodeckID \\
    LayoutGPT & \trMainFbLayoutgptE & \trPairFbLayoutgptED & \trMainFbLayoutgptI & \trPairFbLayoutgptID \\
    I-Design & \textbf{\trMainFbIdesignE} & \trPairFbIdesignED & \underline{\trMainFbIdesignI} & \trPairFbIdesignID \\
    LayoutVLM & \underline{\trMainFbLayoutvlmE} & \trPairFbLayoutvlmED & \textbf{\trMainFbLayoutvlmI} & \trPairFbLayoutvlmID \\
    R3L & \trMainFbRThreelE & \trPairFbRThreelED & \trMainFbRThreelI & \trPairFbRThreelID \\
    \midrule
    \multicolumn{5}{@{}l}{\emph{Attribute list, without feedback}} \\
    Holodeck & \underline{\trMainOssHolodeckE} & \trPairOssHolodeckED & \trMainOssHolodeckI & \trPairOssHolodeckID \\
    LayoutGPT & \trMainOssLayoutgptE & \trPairOssLayoutgptED & \trMainOssLayoutgptI & \trPairOssLayoutgptID \\
    I-Design & \trMainOssIdesignE & \trPairOssIdesignED & \trMainOssIdesignI & \trPairOssIdesignID \\
    LayoutVLM & \textbf{\trMainOssLayoutvlmE} & \trPairOssLayoutvlmED & \textbf{\trMainOssLayoutvlmI} & \trPairOssLayoutvlmID \\
    R3L & \underline{\trMainOssRThreelE} & \trPairOssRThreelED & \underline{\trMainOssRThreelI} & \trPairOssRThreelID \\
    \midrule
    \multicolumn{5}{@{}l}{\emph{Without the attribute list and without feedback}} \\
    Holodeck & \trMainOsHolodeckE & \trPairOsHolodeckED & \underline{\trMainOsHolodeckI} & \trPairOsHolodeckID \\
    LayoutGPT & \trMainOsLayoutgptE & \trPairOsLayoutgptED & \trMainOsLayoutgptI & \trPairOsLayoutgptID \\
    I-Design & \trMainOsIdesignE & \trPairOsIdesignED & \trMainOsIdesignI & \trPairOsIdesignID \\
    LayoutVLM & \textbf{\trMainOsLayoutvlmE} & \trPairOsLayoutvlmED & \textbf{\trMainOsLayoutvlmI} & \trPairOsLayoutvlmID \\
    R3L & \underline{\trMainOsRThreelE} & \trPairOsRThreelED & \trMainOsRThreelI & \trPairOsRThreelID \\
    \midrule
    \multicolumn{5}{@{}l}{\emph{Oracle: given the hidden preferences (one seed)}} \\
    Holodeck & \trMainOrHolodeckE & \trPairOrHolodeckED & \trMainOrHolodeckI & \trPairOrHolodeckID \\
    LayoutGPT & \trMainOrLayoutgptE & \trPairOrLayoutgptED & \trMainOrLayoutgptI & \trPairOrLayoutgptID \\
    I-Design & \trMainOrIdesignE & \trPairOrIdesignED & \trMainOrIdesignI & \trPairOrIdesignID \\
    LayoutVLM & \underline{\trMainOrLayoutvlmE} & \trPairOrLayoutvlmED & \underline{\trMainOrLayoutvlmI} & \trPairOrLayoutvlmID \\
    R3L & \trMainOrRThreelE & \trPairOrRThreelED & \trMainOrRThreelI & \trPairOrRThreelID \\
    \methodname{} & \textbf{\orcc{E}} & -- & \textbf{\orcc{I}} & -- \\
    \midrule
    \methodname{} (Ours) & \textbf{\trMainOursE} & -- & \textbf{\trMainOursI} & -- \\
    \bottomrule
  \end{tabularx}
  \caption{pAMR (\%, $\uparrow$) of the published baselines with and without the attribute list and feedback, and in the oracle condition; mean $\pm$ standard deviation over four seeds (one seed for the oracle rows).
    $\Delta$: \methodname{} with feedback minus the row on the same interactions, in percentage points. Bold: best; underlined: second best; both are marked within each block, and the last row is ranked with the first block.}
  \label{S-tab:supp_variants}
\end{table}

\subsection{Success rates of the generations}
\label{S-sec:supp_success}

Following OptiScene~\cite{yang2025optiscene}, \cref{S-tab:supp_success} reports the share of successful generations of every system with Gemma-4-31B.
Most failed generations of I-Design exceed the time limits of \cref{S-sec:supp_base}, and every failed generation of LayoutVLM has an unsolvable constraint program.

\begin{table}[t]
  \centering
  \footnotesize
  \begin{tabular}{@{}lcc@{}}
\toprule
System & \vE{} & \vI{} \\
\midrule
Holodeck & 98.1 & 98.5 \\
LayoutGPT & \textbf{100.0} & \textbf{100.0} \\
I-Design & 66.6 & 66.3 \\
LayoutVLM & 74.5 & 75.8 \\
R3L & 91.9 & 90.6 \\
Holodeck-Edit & \underline{99.3} & \underline{99.4} \\
\midrule
\methodname{} (Ours) & \textbf{100.0} & \textbf{100.0} \\
\bottomrule
\end{tabular}

  \caption{Successful generations (\%, $\uparrow$) among the four generations of every interaction on the test instances with four seeds. Bold: best; underlined: second best.}
  \label{S-tab:supp_success}
\end{table}

\subsection{Light emitters in the object counts}
\label{S-sec:supp_quality}

Scene instantiation lights a lamp of the asset pool with an emitting sphere of radius 5\,cm below the light source of the lamp (\cref{S-sec:supp_light}).
The sphere is the light source of the lamp and not a separate object, so \cref{tab:quality} of the main paper counts it as part of its lamp in \emph{Coll.} (\cref{S-sec:supp_metrics}).

\subsection{Cost}
\label{S-sec:supp_cost}

\Cref{S-tab:supp_cost} gives the mean cost of one interaction and its range over the interactions.
In addition to its six LLM calls, an interaction of \methodname{} sends on average \trCostOursJevrequests{} requests to Jev with \sxJevTokMean{}k input tokens (\sxJevTokMin{}k--\sxJevTokMax{}k), which cost \trCostOursUsd{}~USD.
The generation time per scene comes from the timing run of the main paper, which executes 8 interactions per system one at a time, with the same settings and servers for all systems.
The generation time per turn is measured in the test interactions and, for \methodname{}, excludes the furnishing call, which runs once after the last turn.

\begin{table*}[t]
  \centering
  \footnotesize
  \setlength{\tabcolsep}{6pt}
  \begin{tabularx}{\textwidth}{@{}l*{5}{Y}@{}}
\toprule
Method & \hdtwo{LLM}{calls\,$\downarrow$} & \hdtwo{LLM input}{tokens ($10^3$)\,$\downarrow$} & \hdtwo{LLM output}{tokens ($10^3$)\,$\downarrow$} & \hdtwo{Generation time}{per scene (s)\,$\downarrow$} & \hdtwo{Generation time}{per turn (s)\,$\downarrow$} \\
\midrule
    Holodeck & \trCostFbsHolodeckLlmcallsMm & \mm{28.9}{14.8}{35.0} & \mm{11.5}{3.6}{19.1} & \trTimFbsHolodeckScenesMm & \trCostFbsHolodeckTurnsMm \\
    LayoutGPT & \textbf{\trCostFbsLayoutgptLlmcallsMm} & \underline{\mm{18.6}{15.2}{22.0}} & \textbf{\mm{1.3}{0.3}{2.0}} & \textbf{\trTimFbsLayoutgptScenesMm} & \textbf{\trCostFbsLayoutgptTurnsMm} \\
    I-Design & \trCostFbsIdesignLlmcallsMm & \mm{123.8}{36.7}{474.0} & \mm{26.1}{5.1}{84.7} & \trTimFbsIdesignScenesMm & \trCostFbsIdesignTurnsMm \\
    LayoutVLM & \trCostFbsLayoutvlmLlmcallsMm & \mm{454.0}{250.2}{783.5} & \underline{\mm{7.9}{3.0}{17.6}} & \trTimFbsLayoutvlmScenesMm & \trCostFbsLayoutvlmTurnsMm \\
    R3L & \trCostFbsRThreelLlmcallsMm & \mm{120.8}{33.1}{287.4} & \mm{28.4}{3.3}{74.8} & \trTimFbsRThreelScenesMm & \trCostFbsRThreelTurnsMm \\
    Holodeck-Edit & \trCostEditHolodeckLlmcallsMm & \textbf{\mm{17.9}{8.5}{22.8}} & \mm{8.5}{1.4}{14.1} & \trTimEditHolodeckScenesMm & \trCostEditHolodeckTurnsMm \\
    \methodname{} (Ours) & \underline{\trCostOursLlmcallsMm} & \mm{35.1}{24.2}{51.7} & \mm{9.2}{7.5}{14.5} & \underline{\trTimOursScenesMm} & \underline{\trCostOursTurnsMm} \\
\bottomrule
\end{tabularx}

  \caption{Cost per interaction ($K=3$, four scenes) on the test instances of both forms, as mean (minimum--maximum) over the interactions. The generation time per scene is measured in the timing run of the main paper (eight interactions per system), and the generation time per turn in the test interactions.
    Tokens are those of Gemma-4-31B. Bold: best; underlined: second best.}
  \label{S-tab:supp_cost}
\end{table*}

\subsection{Evaluator variations}
\label{S-sec:supp_judges}

We re-score the final scenes of all interactions in two ways (\cref{sec:exp_robust} of the main paper).
The second evaluator, Ministral-3-14B, answers the same questions on the same rendered images.
The ``not visible'' variant adds the option ``Not Visible in the Images'' to every question of the evaluator, and this answer counts as no match.
\Cref{S-tab:supp_judges} shows that \methodname{} has the highest pAMR in both forms under both variations.

\begin{table*}[t]
  \centering
  \footnotesize
  \setlength{\tabcolsep}{6pt}
  \begin{tabularx}{\textwidth}{@{}l*{6}{Y}@{}}
\toprule
 & \multicolumn{3}{c}{\vE{}} & \multicolumn{3}{c}{\vI{}} \\
\cmidrule(lr){2-4}\cmidrule(l){5-7}
Method & Evaluator & \hdtwo{Second}{evaluator} & \hdtwo{``Not}{visible''} & Evaluator & \hdtwo{Second}{evaluator} & \hdtwo{``Not}{visible''} \\
\midrule
    Holodeck & 44.0$\pm$0.2 & 42.9$\pm$1.1 & 35.6$\pm$0.9 & 41.4$\pm$0.6 & 40.8$\pm$1.1 & \underline{33.4$\pm$0.6} \\
    LayoutGPT & 27.9$\pm$1.0 & 26.4$\pm$1.9 & 22.2$\pm$1.0 & 25.2$\pm$1.3 & 25.5$\pm$1.2 & 19.9$\pm$1.3 \\
    I-Design & 44.3$\pm$1.1 & 43.8$\pm$1.7 & 34.0$\pm$1.3 & 41.0$\pm$0.5 & 41.0$\pm$0.6 & 30.6$\pm$0.4 \\
    LayoutVLM & 45.1$\pm$0.3 & 44.0$\pm$0.3 & 33.0$\pm$0.4 & \underline{43.1$\pm$1.4} & \underline{43.0$\pm$1.3} & 32.2$\pm$1.0 \\
    R3L & 43.9$\pm$1.0 & 42.4$\pm$1.1 & 33.6$\pm$1.0 & 40.8$\pm$0.6 & 39.9$\pm$1.0 & 31.2$\pm$0.3 \\
    Holodeck-Edit & \underline{45.6$\pm$0.7} & \underline{44.9$\pm$0.5} & \underline{35.8$\pm$1.0} & 42.6$\pm$1.4 & 41.9$\pm$1.2 & 32.3$\pm$1.4 \\
    \methodname{} (Ours) & \textbf{56.6$\pm$0.3} & \textbf{52.5$\pm$1.0} & \textbf{47.8$\pm$0.6} & \textbf{49.6$\pm$1.7} & \textbf{46.5$\pm$0.8} & \textbf{40.9$\pm$1.1} \\
\bottomrule
\end{tabularx}

  \caption{pAMR (\%, $\uparrow$) of the final scene under the evaluator of the main paper (GLM-4.6V-Flash), the second evaluator (Ministral-3-14B) and the evaluator with a ``not visible'' option; mean $\pm$ standard deviation over four seeds. Bold: best; underlined: second best.}
  \label{S-tab:supp_judges}
\end{table*}

\subsection{A hosted LLM in place of Gemma-4-31B}
\label{S-sec:supp_llm}

\Cref{S-tab:supp_llm} compares \methodname{} with GPT-6.1 Terra and with Gemma-4-31B as its LLM on 30 test instances per form, drawn at random, with one interaction per instance at seed 0 (\cref{S-sec:supp_models}).
\methodname{} with GPT-6.1 Terra runs without the furnishing step (\cref{S-sec:supp_furnish}), so both rows score the final scene without it.
With GPT-6.1 Terra, the pAMR is \trPaidPaidE{}\% in the \vE{} form and \trPaidPaidI{}\% in the \vI{} form, against \trPaidOursE{}\% and \trPaidOursI{}\% with Gemma-4-31B.

\begin{table}[t]
  \centering
  \footnotesize
  \setlength{\tabcolsep}{6pt}
  \begin{tabularx}{\linewidth}{@{}l*{2}{Y}@{}}
    \toprule
    LLM of \methodname{} & \vE{} & \vI{} \\
    \midrule
    GPT-6.1 Terra & \trPaidPaidE & \textbf{\trPaidPaidI} \\
    Gemma-4-31B & \textbf{\trPaidOursE} & \trPaidOursI \\
    \bottomrule
  \end{tabularx}
  \caption{pAMR (\%, $\uparrow$) of \methodname{} with two LLMs on the same 30 test interactions per form (seed 0), scored on the final scenes without the furnishing step. Bold: best.}
  \label{S-tab:supp_llm}
\end{table}

\subsection{Ablation details}
\label{S-sec:supp_ablation}

\Cref{tab:ablation} of the main paper gives the pAMR of the final scenes of the ablation variants.
At turn 1, we compute the difference between a variant and the full method on the same interactions and on the scenes shown in the interaction, which are not furnished.
In percentage points, this difference is \trAblCTwoDROneE{} in the \vE{} form and \trAblCTwoDROneI{} in the \vI{} form without older turns in the history, and \trAblCThreebDROneE{} and \trAblCThreebDROneI{} without new values.
The variants thus differ from the full method by less than 1 point at turn 1, so the differences in the final scenes come from the plan updates.

\subsection{Baselines with our scene instantiation}
\label{S-sec:supp_inst_baselines}
Our scene instantiation sets the color of each wall, the light sources and the window coverings from values of their own, whereas the baselines set one material for all walls, render every scene under one fixed lighting setup and place window coverings at most as stock assets (\cref{S-sec:supp_base}).
We apply it to the final scenes of each baseline, with the values that this baseline decided itself for these parts, and compare these scenes with the original scenes of the baseline and with those of \methodname{}.

\begin{table*}[t]
  \centering
  \footnotesize
  \setlength{\tabcolsep}{3pt}
  \newcolumntype{A}[1]{>{\hsize=#1\hsize}Y}
  \begin{tabularx}{\textwidth}{@{}l*{2}{A{0.83}A{0.83}A{0.83}A{1.255}A{1.255}}@{}}
    \toprule
    & \multicolumn{5}{c}{\vE{}} & \multicolumn{5}{c}{\vI{}} \\
    \cmidrule(lr){2-6}\cmidrule(l){7-11}
    Baseline & Original & Own & Ours & \hdtwo{Own}{$-$ original} & \hdtwo{Ours}{$-$ own} & Original & Own & Ours & \hdtwo{Own}{$-$ original} & \hdtwo{Ours}{$-$ own} \\
    \midrule
    \multicolumn{11}{@{}l}{\emph{All attributes}} \\
    Holodeck & \eElevenAllHolodeckOrigE & \eElevenAllHolodeckOwnE & \eElevenAllHolodeckMethE & \dci{\eElevenAllHolodeckDOoE$\pm$\eElevenAllHolodeckDOoStdE}{\eElevenAllHolodeckCiOoE} & \dci{\eElevenAllHolodeckDMoE$\pm$\eElevenAllHolodeckDMoStdE}{\eElevenAllHolodeckCiMoE} & \eElevenAllHolodeckOrigI & \eElevenAllHolodeckOwnI & \eElevenAllHolodeckMethI & \dci{\eElevenAllHolodeckDOoI$\pm$\eElevenAllHolodeckDOoStdI}{\eElevenAllHolodeckCiOoI} & \dci{\eElevenAllHolodeckDMoI$\pm$\eElevenAllHolodeckDMoStdI}{\eElevenAllHolodeckCiMoI} \\
    I-Design & \eElevenAllIdesignOrigE & \eElevenAllIdesignOwnE & \eElevenAllIdesignMethE & \dci{\eElevenAllIdesignDOoE$\pm$\eElevenAllIdesignDOoStdE}{\eElevenAllIdesignCiOoE} & \dci{\eElevenAllIdesignDMoE$\pm$\eElevenAllIdesignDMoStdE}{\eElevenAllIdesignCiMoE} & \eElevenAllIdesignOrigI & \eElevenAllIdesignOwnI & \eElevenAllIdesignMethI & \dci{\eElevenAllIdesignDOoI$\pm$\eElevenAllIdesignDOoStdI}{\eElevenAllIdesignCiOoI} & \dci{\eElevenAllIdesignDMoI$\pm$\eElevenAllIdesignDMoStdI}{\eElevenAllIdesignCiMoI} \\
    LayoutVLM & \eElevenAllLayoutvlmOrigE & \eElevenAllLayoutvlmOwnE & \eElevenAllLayoutvlmMethE & \dci{\eElevenAllLayoutvlmDOoE$\pm$\eElevenAllLayoutvlmDOoStdE}{\eElevenAllLayoutvlmCiOoE} & \dci{\eElevenAllLayoutvlmDMoE$\pm$\eElevenAllLayoutvlmDMoStdE}{\eElevenAllLayoutvlmCiMoE} & \eElevenAllLayoutvlmOrigI & \eElevenAllLayoutvlmOwnI & \eElevenAllLayoutvlmMethI & \dci{\eElevenAllLayoutvlmDOoI$\pm$\eElevenAllLayoutvlmDOoStdI}{\eElevenAllLayoutvlmCiOoI} & \dci{\eElevenAllLayoutvlmDMoI$\pm$\eElevenAllLayoutvlmDMoStdI}{\eElevenAllLayoutvlmCiMoI} \\
    R3L & \eElevenAllRThreelOrigE & \eElevenAllRThreelOwnE & \eElevenAllRThreelMethE & \dci{\eElevenAllRThreelDOoE$\pm$\eElevenAllRThreelDOoStdE}{\eElevenAllRThreelCiOoE} & \dci{\eElevenAllRThreelDMoE$\pm$\eElevenAllRThreelDMoStdE}{\eElevenAllRThreelCiMoE} & \eElevenAllRThreelOrigI & \eElevenAllRThreelOwnI & \eElevenAllRThreelMethI & \dci{\eElevenAllRThreelDOoI$\pm$\eElevenAllRThreelDOoStdI}{\eElevenAllRThreelCiOoI} & \dci{\eElevenAllRThreelDMoI$\pm$\eElevenAllRThreelDMoStdI}{\eElevenAllRThreelCiMoI} \\
    Holodeck-Edit & \eElevenAllEditHolodeckOrigE & \eElevenAllEditHolodeckOwnE & \eElevenAllEditHolodeckMethE & \dci{\eElevenAllEditHolodeckDOoE$\pm$\eElevenAllEditHolodeckDOoStdE}{\eElevenAllEditHolodeckCiOoE} & \dci{\eElevenAllEditHolodeckDMoE$\pm$\eElevenAllEditHolodeckDMoStdE}{\eElevenAllEditHolodeckCiMoE} & \eElevenAllEditHolodeckOrigI & \eElevenAllEditHolodeckOwnI & \eElevenAllEditHolodeckMethI & \dci{\eElevenAllEditHolodeckDOoI$\pm$\eElevenAllEditHolodeckDOoStdI}{\eElevenAllEditHolodeckCiOoI} & \dci{\eElevenAllEditHolodeckDMoI$\pm$\eElevenAllEditHolodeckDMoStdI}{\eElevenAllEditHolodeckCiMoI} \\
    \midrule
    \multicolumn{11}{@{}l}{\emph{Added attributes}} \\
    Holodeck & \eElevenAddHolodeckOrigE & \eElevenAddHolodeckOwnE & \eElevenAddHolodeckMethE & \dci{\eElevenAddHolodeckDOoE$\pm$\eElevenAddHolodeckDOoStdE}{\eElevenAddHolodeckCiOoE} & \dci{\eElevenAddHolodeckDMoE$\pm$\eElevenAddHolodeckDMoStdE}{\eElevenAddHolodeckCiMoE} & \eElevenAddHolodeckOrigI & \eElevenAddHolodeckOwnI & \eElevenAddHolodeckMethI & \dci{\eElevenAddHolodeckDOoI$\pm$\eElevenAddHolodeckDOoStdI}{\eElevenAddHolodeckCiOoI} & \dci{\eElevenAddHolodeckDMoI$\pm$\eElevenAddHolodeckDMoStdI}{\eElevenAddHolodeckCiMoI} \\
    I-Design & \eElevenAddIdesignOrigE & \eElevenAddIdesignOwnE & \eElevenAddIdesignMethE & \dci{\eElevenAddIdesignDOoE$\pm$\eElevenAddIdesignDOoStdE}{\eElevenAddIdesignCiOoE} & \dci{\eElevenAddIdesignDMoE$\pm$\eElevenAddIdesignDMoStdE}{\eElevenAddIdesignCiMoE} & \eElevenAddIdesignOrigI & \eElevenAddIdesignOwnI & \eElevenAddIdesignMethI & \dci{\eElevenAddIdesignDOoI$\pm$\eElevenAddIdesignDOoStdI}{\eElevenAddIdesignCiOoI} & \dci{\eElevenAddIdesignDMoI$\pm$\eElevenAddIdesignDMoStdI}{\eElevenAddIdesignCiMoI} \\
    LayoutVLM & \eElevenAddLayoutvlmOrigE & \eElevenAddLayoutvlmOwnE & \eElevenAddLayoutvlmMethE & \dci{\eElevenAddLayoutvlmDOoE$\pm$\eElevenAddLayoutvlmDOoStdE}{\eElevenAddLayoutvlmCiOoE} & \dci{\eElevenAddLayoutvlmDMoE$\pm$\eElevenAddLayoutvlmDMoStdE}{\eElevenAddLayoutvlmCiMoE} & \eElevenAddLayoutvlmOrigI & \eElevenAddLayoutvlmOwnI & \eElevenAddLayoutvlmMethI & \dci{\eElevenAddLayoutvlmDOoI$\pm$\eElevenAddLayoutvlmDOoStdI}{\eElevenAddLayoutvlmCiOoI} & \dci{\eElevenAddLayoutvlmDMoI$\pm$\eElevenAddLayoutvlmDMoStdI}{\eElevenAddLayoutvlmCiMoI} \\
    R3L & \eElevenAddRThreelOrigE & \eElevenAddRThreelOwnE & \eElevenAddRThreelMethE & \dci{\eElevenAddRThreelDOoE$\pm$\eElevenAddRThreelDOoStdE}{\eElevenAddRThreelCiOoE} & \dci{\eElevenAddRThreelDMoE$\pm$\eElevenAddRThreelDMoStdE}{\eElevenAddRThreelCiMoE} & \eElevenAddRThreelOrigI & \eElevenAddRThreelOwnI & \eElevenAddRThreelMethI & \dci{\eElevenAddRThreelDOoI$\pm$\eElevenAddRThreelDOoStdI}{\eElevenAddRThreelCiOoI} & \dci{\eElevenAddRThreelDMoI$\pm$\eElevenAddRThreelDMoStdI}{\eElevenAddRThreelCiMoI} \\
    Holodeck-Edit & \eElevenAddEditHolodeckOrigE & \eElevenAddEditHolodeckOwnE & \eElevenAddEditHolodeckMethE & \dci{\eElevenAddEditHolodeckDOoE$\pm$\eElevenAddEditHolodeckDOoStdE}{\eElevenAddEditHolodeckCiOoE} & \dci{\eElevenAddEditHolodeckDMoE$\pm$\eElevenAddEditHolodeckDMoStdE}{\eElevenAddEditHolodeckCiMoE} & \eElevenAddEditHolodeckOrigI & \eElevenAddEditHolodeckOwnI & \eElevenAddEditHolodeckMethI & \dci{\eElevenAddEditHolodeckDOoI$\pm$\eElevenAddEditHolodeckDOoStdI}{\eElevenAddEditHolodeckCiOoI} & \dci{\eElevenAddEditHolodeckDMoI$\pm$\eElevenAddEditHolodeckDMoStdI}{\eElevenAddEditHolodeckCiMoI} \\
    \bottomrule
  \end{tabularx}
  \caption{pAMR (\%, $\uparrow$) of each baseline's final scene (Original), of the same scene with our scene instantiation of the added attributes (Own) and of the final scene of \methodname{} before its furnishing step (Ours), and the differences in percentage points, on all test instances of \cref{tab:main} of the main paper (\eElevenNinst{} per form) with seeds 0--3.
    Own uses the values that the baseline decided itself and default values otherwise.
    Mean $\pm$ standard deviation over four seeds, with the 95\% bootstrap interval over instances below each difference.
    Added attributes: the wall color and the lighting, and in the room types that have them the feature wall, the window treatment, the wall tile, the backsplash, the upper cabinets and the acoustic treatment.
    For R3L, the added attributes also include the mirror (bathroom), the table setting (dining room), the art type and the hanging style (gallery), the decor theme (kitchen) and the wall decoration (waiting room), so in this block the Ours value of R3L differs from that of the other baselines.}
  \label{S-tab:supp_inst_baselines}
\end{table*}

\paragraph{Setup.}
We use the final scenes of \cref{tab:main} of the main paper that each baseline generated on all test instances (\eElevenNinst{} per form) with seeds 0--3.
We call \emph{added attributes} the attributes of the walls, the lighting and the window coverings, and the attributes whose objects the baseline cannot place.
Into each final scene, we build one value of each added attribute.
For every baseline, these are the wall color and the lighting in every room type, the feature wall and the window treatment in the room types that have these attributes, the wall tile in the bathroom, the backsplash and the upper cabinets in the kitchen, and the acoustic treatment in the waiting room.
For R3L, they also include the mirror in the bathroom, the table setting in the dining room, the art type and the hanging style in the gallery, the decor theme in the kitchen and the wall decoration in the waiting room, which R3L cannot change.
We do not add the attributes of the floor or of the objects that the baseline places, such as the furniture style, the storage and the decor.

Each value is the one that the baseline decided itself in the texts of its final generation, and fixed keyword rules extract it without a model call.
These texts are the wall design, the descriptions of the objects that the baseline selected and, if its LLM writes one, the design strategy; for I-Design, LayoutVLM and R3L, the wall design is that of the Holodeck generation that builds their walls (\cref{S-sec:supp_base}).
An added attribute that these texts do not decide takes the default specification of its category (\cref{S-sec:supp_defaults}).
The color of a wall, a feature wall or a window covering is the first of the 148 named colors of Matplotlib that the text of its value names, with its lightness word.
If the text names none of these colors, or another color or tone word modifies the name or is joined to it, Gemma-4-31B, the LLM of all systems, reads this text alone and answers one of these colors or none.
If the answer is none, the value takes the default color or, for a feature wall, the color closest to the text by CLIP similarity (\cref{S-sec:supp_retrieval}).
No model in this construction reads the feedback.
The scene instantiation of \methodname{} translates each value into a specification and builds it with the builder, the heights and the placement that \methodname{} uses for this attribute (\cref{sec:method_repr} of the main paper and \cref{S-sec:supp_real}).
The lighting value replaces the fixed lighting of the baseline scenes (\cref{S-sec:supp_base}), adds the light fixtures that the value describes and lights the lamps of the baseline (\cref{S-sec:supp_light}).
The table setting is placed on the dining table of the baseline or, when the scene has none or the table is full, on other furniture, and the backsplash and the upper cabinets follow the counters of the baseline or, in a kitchen without counters, the longest free stretch of wall.
Apart from these changes, no object of the baseline is moved, replaced, recolored or removed.
As in the scenes of \methodname{}, every added object is checked for collisions with the objects of the baseline and is not placed if it collides; a floor-length curtain that collides is tried once more at sill length.

\paragraph{Comparison.}
We render the scenes with the baselines' own values and score them with the evaluator (\cref{S-sec:supp_judge}) like every other final scene.
On the same instances and seeds, we compare the final scene of the baseline (Original), the same scene with its own values (Own) and the final scene of the interaction of \methodname{} (Ours).
The baselines have no furnishing step (\cref{S-sec:supp_furnish}), so Ours is the final scene before this step.
We compute each difference on the triples of instance, seed and attribute and average it over the triples of each seed.
We report the mean $\pm$ standard deviation over the four seeds with a 95\% percentile bootstrap interval over instances (\suppBootB{} resamples).

\paragraph{Result.}
\Cref{S-tab:supp_inst_baselines} gives the pAMR on all attributes and on the added attributes.
On all attributes, the scene with the baseline's own values minus the original lies between \eElevenRngAllDOoLoE{} and \eElevenRngAllDOoHiE{} points in the \vE{} form and between \eElevenRngAllDOoLoI{} and \eElevenRngAllDOoHiI{} points in the \vI{} form, and every interval lies above 0.
On the added attributes, this difference lies between \eElevenRngAddDOoLoE{} and \eElevenRngAddDOoHiE{} points in the \vE{} form and between \eElevenRngAddDOoLoI{} and \eElevenRngAddDOoHiI{} points in the \vI{} form, and every interval lies above 0.
On the attributes that are not added, which the replaced lighting can still affect, this difference lies between \eElevenRngRestDOoLoE{} and \eElevenRngRestDOoHiE{} points in the \vE{} form and between \eElevenRngRestDOoLoI{} and \eElevenRngRestDOoHiI{} points in the \vI{} form, and no interval lies below 0.

On the added attributes, \methodname{} minus the scene with the baseline's own values lies between \eElevenRngAddDMoLoE{} and \eElevenRngAddDMoHiE{} points in the \vE{} form and between \eElevenRngAddDMoLoI{} and \eElevenRngAddDMoHiI{} points in the \vI{} form, and every interval lies above 0.
On all attributes, this difference lies between \eElevenRngAllDMoLoE{} and \eElevenRngAllDMoHiE{} points in the \vE{} form and between \eElevenRngAllDMoLoI{} and \eElevenRngAllDMoHiI{} points in the \vI{} form, and every interval lies above 0.
With the same scene instantiation, \methodname{} matches more hidden values than every baseline in both forms.

\begin{table*}[t]
  \centering
  \footnotesize
  \setlength{\tabcolsep}{4pt}
  \begin{tabularx}{\textwidth}{@{}l*{4}{Y}@{}}
    \toprule
    & \multicolumn{2}{c}{\vE{}} & \multicolumn{2}{c}{\vI{}} \\
    \cmidrule(lr){2-3}\cmidrule(l){4-5}
    System & pAMR & \methodname{} $-$ system & pAMR & \methodname{} $-$ system \\
    \midrule
    Holodeck & \eTwelveCHolodeckBaseE & \dci{\eTwelveCHolodeckDE$\pm$\eTwelveCHolodeckDstdE}{\eTwelveCHolodeckCiE} & \eTwelveCHolodeckBaseI & \dci{\eTwelveCHolodeckDI$\pm$\eTwelveCHolodeckDstdI}{\eTwelveCHolodeckCiI} \\
    I-Design & \eTwelveCIdesignBaseE & \dci{\eTwelveCIdesignDE$\pm$\eTwelveCIdesignDstdE}{\eTwelveCIdesignCiE} & \eTwelveCIdesignBaseI & \dci{\eTwelveCIdesignDI$\pm$\eTwelveCIdesignDstdI}{\eTwelveCIdesignCiI} \\
    LayoutVLM & \eTwelveCLayoutvlmBaseE & \dci{\eTwelveCLayoutvlmDE$\pm$\eTwelveCLayoutvlmDstdE}{\eTwelveCLayoutvlmCiE} & \eTwelveCLayoutvlmBaseI & \dci{\eTwelveCLayoutvlmDI$\pm$\eTwelveCLayoutvlmDstdI}{\eTwelveCLayoutvlmCiI} \\
    R3L & \eTwelveCRThreelBaseE & \dci{\eTwelveCRThreelDE$\pm$\eTwelveCRThreelDstdE}{\eTwelveCRThreelCiE} & \eTwelveCRThreelBaseI & \dci{\eTwelveCRThreelDI$\pm$\eTwelveCRThreelDstdI}{\eTwelveCRThreelCiI} \\
    Holodeck-Edit & \eTwelveCEditHolodeckBaseE & \dci{\eTwelveCEditHolodeckDE$\pm$\eTwelveCEditHolodeckDstdE}{\eTwelveCEditHolodeckCiE} & \eTwelveCEditHolodeckBaseI & \dci{\eTwelveCEditHolodeckDI$\pm$\eTwelveCEditHolodeckDstdI}{\eTwelveCEditHolodeckCiI} \\
    \midrule
    \methodname{} (ours) & \textbf{\eTwelveCHolodeckOursE} & -- & \textbf{\eTwelveCHolodeckOursI} & -- \\
    \bottomrule
  \end{tabularx}
  \caption{pAMR (\%, $\uparrow$) on the \eTwelveCNa{} attributes that every system other than LayoutGPT can change, and \methodname{} minus each baseline in percentage points.
    Scenes: the final scenes of the interactions of \cref{tab:main} of the main paper (\eTwelveCNinst{} test instances per form, seeds 0--3).
    Mean $\pm$ standard deviation over four seeds, with the 95\% bootstrap interval over instances below each difference.}
  \label{S-tab:supp_shared_attrs}
\end{table*}

\subsection{Match on the attributes\texorpdfstring{\protect\\}{} that every system can change}
\label{S-sec:supp_shared_attrs}

The published baselines select and place objects and pick the materials of the walls and the floor (Sec.~6.2 of the main paper).
We compare \methodname{} with them on the attributes whose part of the room every system sets in these ways.

\paragraph{Attributes.}
For every room type, attribute and system, we read the code of the system with our adaptation (\cref{S-sec:supp_base}) and its final scenes.
An attribute is part of the comparison when each of the six systems other than LayoutGPT sets the part of the room that the attribute describes from its input, through the objects that it places or the material of the walls or the floor, and when this part differs between the final scenes of the system for the room type.
This holds for \eTwelveCNa{} of the \eTwelveAllNa{} attributes, which cover all ten room types.
The other \eTwelveCNexcl{} attributes are the lighting of the ten room types, whose values describe lit light sources, whereas every baseline scene is rendered under one fixed lighting setup (\cref{S-sec:supp_base}); the window treatment of the \eTwelveCNwin{} room types that have one, which Holodeck, R3L and Holodeck-Edit do not place; the upper cabinets of the kitchen, for which no asset of the shared pool can be mounted on a wall; and \eTwelveCNrthree{} attributes of wall pieces and decorations, which R3L does not place because it places floor objects only: the mirror, the table setting, the art type, the hanging style, the decor theme of the kitchen and the wall decoration.
LayoutGPT runs only on bedrooms and living rooms and places object boxes without materials in a room whose walls and floor have fixed materials (\cref{S-sec:supp_base}), so it is not part of this comparison.

\paragraph{Comparison.}
We use the final scenes of the interactions of \cref{tab:main} of the main paper and compute the difference, \methodname{} minus the baseline, on the triples of instance, seed and attribute, averaged over the triples of each seed.
We report the mean $\pm$ standard deviation over the four seeds with a 95\% percentile bootstrap interval over instances (\suppBootB{} resamples).

\paragraph{Result.}
\Cref{S-tab:supp_shared_attrs} gives the pAMR on these attributes.
\methodname{} has the highest pAMR in both forms.
Its difference to the baselines lies between \eTwelveCRngLoE{} and \eTwelveCRngHiE{} points in the \vE{} form and between \eTwelveCRngLoI{} and \eTwelveCRngHiI{} points in the \vI{} form, and every interval lies above 0.

\subsection{An open model in place of Jev}
\label{S-sec:supp_open_jev}
\methodname{} sends two kinds of multiple-choice questions to Jev~\cite{typesafe2026jev}, a hosted decision model that is paid per request (\cref{S-sec:supp_cost}): the questions that assign the design aspects to the attributes before turn 1 (\cref{S-sec:supp_cand}) and the specification questions of scene instantiation in every turn (\cref{S-sec:supp_questions}).
To test whether \methodname{} needs this hosted model, we answer both kinds of questions with an open model and compare the match with that of \methodname{} with Jev on the same instances and seeds (\cref{S-tab:supp_open_jev}).

\begin{table*}[t]
  \centering
  \footnotesize
  \setlength{\tabcolsep}{4pt}
  \begin{tabularx}{\textwidth}{@{}l*{6}{Y}@{}}
    \toprule
    & \multicolumn{3}{c}{\vE{}} & \multicolumn{3}{c}{\vI{}} \\
    \cmidrule(lr){2-4}\cmidrule(l){5-7}
    Scene & Jev & Open model & $\Delta$ & Jev & Open model & $\Delta$ \\
    \midrule
    \multicolumn{7}{@{}l}{\emph{Test instances of \cref{S-sec:supp_llm} (30 per form)}} \\
    Turn 1 & \eThirteenNineTurnOneCtrlE & \eThirteenNineTurnOneArmE & \dci{\eThirteenNineTurnOneDE$\pm$\eThirteenNineTurnOneDStdE}{\eThirteenNineTurnOneCiE} & \eThirteenNineTurnOneCtrlI & \eThirteenNineTurnOneArmI & \dci{\eThirteenNineTurnOneDI$\pm$\eThirteenNineTurnOneDStdI}{\eThirteenNineTurnOneCiI} \\
    Final scene & \eThirteenNineFinalCtrlE & \eThirteenNineFinalArmE & \dci{\eThirteenNineFinalDE$\pm$\eThirteenNineFinalDStdE}{\eThirteenNineFinalCiE} & \eThirteenNineFinalCtrlI & \eThirteenNineFinalArmI & \dci{\eThirteenNineFinalDI$\pm$\eThirteenNineFinalDStdI}{\eThirteenNineFinalCiI} \\
    \midrule
    \multicolumn{7}{@{}l}{\emph{Dining-room and office test instances (24 per form)}} \\
    Turn 1 & \eThirteenAblTurnOneCtrlE & \eThirteenAblTurnOneArmE & \dci{\eThirteenAblTurnOneDE$\pm$\eThirteenAblTurnOneDStdE}{\eThirteenAblTurnOneCiE} & \eThirteenAblTurnOneCtrlI & \eThirteenAblTurnOneArmI & \dci{\eThirteenAblTurnOneDI$\pm$\eThirteenAblTurnOneDStdI}{\eThirteenAblTurnOneCiI} \\
    Final scene & \eThirteenAblFinalCtrlE & \eThirteenAblFinalArmE & \dci{\eThirteenAblFinalDE$\pm$\eThirteenAblFinalDStdE}{\eThirteenAblFinalCiE} & \eThirteenAblFinalCtrlI & \eThirteenAblFinalArmI & \dci{\eThirteenAblFinalDI$\pm$\eThirteenAblFinalDStdI}{\eThirteenAblFinalCiI} \\
    \bottomrule
  \end{tabularx}
  \caption{\methodname{} with Jev and with Gemma-4-31B in place of Jev (open model), both on two subsets of the test instances with seeds 0--2, with all scenes scored after the furnishing step: pAMR (\%, $\uparrow$) and the difference between the open model and Jev ($\Delta$, percentage points); mean $\pm$ standard deviation over three seeds, with the 95\% bootstrap interval over instances below each difference.}
  \label{S-tab:supp_open_jev}
\end{table*}

\paragraph{Setup.}
The open model is Gemma-4-31B, the open-weight LLM of \methodname{} (\cref{S-tab:roles}), so the LLM and the decision model are the same model.
A request to Jev contains one context and several multiple-choice questions.
The open model receives each question of a Jev request in a separate call, which gives the context of the request and the question with its options labeled A, B, C, \ldots\ and asks for the letter of the best option.
For each option, we take the probability that the first token of the reply is its letter and divide it by the sum of these probabilities over the shown options.
Every question is asked with its options in the given and in the reversed order, and the two distributions are averaged.
A question with more than 26 options, more than there are letters, is split into questions on groups of options and one question on the most probable options of every group, and the distributions of these questions are combined into one distribution over all options by the maximum-likelihood fit of Luce's choice model.
The open model thus returns a probability for every option of every question, as Jev does.
The questions, the options and every rule that turns these probabilities into an assignment or a specification, including the thresholds of the assignment (\cref{S-sec:supp_cand}), are unchanged.
\methodname{} uses no calibration fitted to the probabilities of Jev, so no parameter is refitted for the~open~model.\looseness=1

\paragraph{Comparison.}
We run \methodname{} with the open model with $K=3$ and seeds 0--2 on two sets of test instances: the 30 instances per form of \cref{S-sec:supp_llm} (90 interactions per form) and the 24 dining-room and office instances per form of the ablations (72 interactions per form).
The control is \methodname{} with Jev on the same instances and seeds, \ie the interactions of \cref{tab:main} of the main paper.
Both configurations start from the same reply of the candidate call, the LLM call that writes the candidate values before turn 1 (\cref{S-sec:supp_cand}), and therefore have the same candidate values.
The two configurations can differ from the assignment onward.
As in \cref{S-fig:supp_rounds}, every scene is rendered and scored by the evaluator (\cref{S-sec:supp_judge}) after the furnishing step (\cref{S-sec:supp_furnish}).
Each difference is computed on the triples of instance, seed and attribute.
It is averaged over the triples of each seed and reported as the mean $\pm$ standard deviation over the three seeds, with a 95\% percentile bootstrap interval over instances (\suppBootB{} resamples).

\paragraph{Result.}
\Cref{S-tab:supp_open_jev} gives the match.
On the final scenes of the test instances of \cref{S-sec:supp_llm}, the difference between the open model and Jev is \eThirteenNineFinalDE{} points in the \vE{} form (95\% interval \eThirteenNineFinalCiE{}) and \eThirteenNineFinalDI{} points in the \vI{} form (\eThirteenNineFinalCiI{}).
On the final scenes of the dining-room and office test instances, the difference is \eThirteenAblFinalDE{} points in the \vE{} form (\mbox{\eThirteenAblFinalCiE{}}) and \eThirteenAblFinalDI{} points in the \vI{} form (\eThirteenAblFinalCiI{}); both intervals contain 0.
At turn 1, the first scene after the assignment, and in the final scene, every interval in \cref{S-tab:supp_open_jev} contains 0 on both sets of instances, and the intervals of turns 2 and 3 also contain 0.
With the open model in place of Jev, no difference in match is detected: the four intervals of the final scenes contain 0 and exclude a loss of more than \eThirteenNineFinalMaxLoss{} points.
The open model answers the multiple-choice questions of an interaction with \eThirteenCallsMean{} calls to Gemma-4-31B on average (\eThirteenCallsMin{}--\eThirteenCallsMax{}), because it receives each question in a separate call, whereas \methodname{} with Jev sends about \trCostOursJevrequests{} requests to Jev per interaction on the test instances (\cref{S-sec:supp_cost}).
On the interactions of \cref{S-tab:supp_open_jev}, an interaction with the open model sends no paid request and on average \revDOpenTokIn{}k input tokens to Gemma-4-31B, whereas one with Jev sends \revDJevTokIn{}k input tokens to Gemma-4-31B and \revDJevJevTokIn{}k to Jev.

\subsection{Noisy feedback}
\label{S-sec:supp_noisy}
\begin{table*}[t]
  \centering
  \footnotesize
  \setlength{\tabcolsep}{3pt}
  \newcolumntype{A}[1]{>{\hsize=#1\hsize}Y}
  \begin{tabularx}{\textwidth}{@{}l*{4}{A{0.86}A{1.14}}@{}}
    \toprule
    & \multicolumn{4}{c}{\vE{}} & \multicolumn{4}{c}{\vI{}} \\
    \cmidrule(lr){2-5}\cmidrule(l){6-9}
    & \multicolumn{2}{c}{All attributes} & \multicolumn{2}{c}{Wrongly stated attributes} & \multicolumn{2}{c}{All attributes} & \multicolumn{2}{c}{Wrongly stated attributes} \\
    \cmidrule(lr){2-3}\cmidrule(lr){4-5}\cmidrule(lr){6-7}\cmidrule(l){8-9}
    Scene & pAMR\,$\uparrow$ & $\Delta$ & pAMR\,$\uparrow$ & $\Delta$ & pAMR\,$\uparrow$ & $\Delta$ & pAMR\,$\uparrow$ & $\Delta$ \\
    \midrule
    \multicolumn{9}{@{}l}{\emph{Noisy feedback with $p=0.25$}} \\
    Turn 2 & \eFourteenQuarterTurnTwoAllArmE & \dci{\eFourteenQuarterTurnTwoAllDE$\pm$\eFourteenQuarterTurnTwoAllDStdE}{\eFourteenQuarterTurnTwoAllCiE} & \eFourteenQuarterTurnTwoWrongArmE & \dci{\eFourteenQuarterTurnTwoWrongDE$\pm$\eFourteenQuarterTurnTwoWrongDStdE}{\eFourteenQuarterTurnTwoWrongCiE} & \eFourteenQuarterTurnTwoAllArmI & \dci{\eFourteenQuarterTurnTwoAllDI$\pm$\eFourteenQuarterTurnTwoAllDStdI}{\eFourteenQuarterTurnTwoAllCiI} & \eFourteenQuarterTurnTwoWrongArmI & \dci{\eFourteenQuarterTurnTwoWrongDI$\pm$\eFourteenQuarterTurnTwoWrongDStdI}{\eFourteenQuarterTurnTwoWrongCiI} \\
    Turn 3 & \eFourteenQuarterTurnThreeAllArmE & \dci{\eFourteenQuarterTurnThreeAllDE$\pm$\eFourteenQuarterTurnThreeAllDStdE}{\eFourteenQuarterTurnThreeAllCiE} & \eFourteenQuarterTurnThreeWrongArmE & \dci{\eFourteenQuarterTurnThreeWrongDE$\pm$\eFourteenQuarterTurnThreeWrongDStdE}{\eFourteenQuarterTurnThreeWrongCiE} & \eFourteenQuarterTurnThreeAllArmI & \dci{\eFourteenQuarterTurnThreeAllDI$\pm$\eFourteenQuarterTurnThreeAllDStdI}{\eFourteenQuarterTurnThreeAllCiI} & \eFourteenQuarterTurnThreeWrongArmI & \dci{\eFourteenQuarterTurnThreeWrongDI$\pm$\eFourteenQuarterTurnThreeWrongDStdI}{\eFourteenQuarterTurnThreeWrongCiI} \\
    Final scene & \eFourteenQuarterFinalAllArmE & \dci{\eFourteenQuarterFinalAllDE$\pm$\eFourteenQuarterFinalAllDStdE}{\eFourteenQuarterFinalAllCiE} & \eFourteenQuarterFinalWrongArmE & \dci{\eFourteenQuarterFinalWrongDE$\pm$\eFourteenQuarterFinalWrongDStdE}{\eFourteenQuarterFinalWrongCiE} & \eFourteenQuarterFinalAllArmI & \dci{\eFourteenQuarterFinalAllDI$\pm$\eFourteenQuarterFinalAllDStdI}{\eFourteenQuarterFinalAllCiI} & \eFourteenQuarterFinalWrongArmI & \dci{\eFourteenQuarterFinalWrongDI$\pm$\eFourteenQuarterFinalWrongDStdI}{\eFourteenQuarterFinalWrongCiI} \\
    \midrule
    \multicolumn{9}{@{}l}{\emph{Noisy feedback with $p=0.5$}} \\
    Turn 2 & \eFourteenHalfTurnTwoAllArmE & \dci{\eFourteenHalfTurnTwoAllDE$\pm$\eFourteenHalfTurnTwoAllDStdE}{\eFourteenHalfTurnTwoAllCiE} & \eFourteenHalfTurnTwoWrongArmE & \dci{\eFourteenHalfTurnTwoWrongDE$\pm$\eFourteenHalfTurnTwoWrongDStdE}{\eFourteenHalfTurnTwoWrongCiE} & \eFourteenHalfTurnTwoAllArmI & \dci{\eFourteenHalfTurnTwoAllDI$\pm$\eFourteenHalfTurnTwoAllDStdI}{\eFourteenHalfTurnTwoAllCiI} & \eFourteenHalfTurnTwoWrongArmI & \dci{\eFourteenHalfTurnTwoWrongDI$\pm$\eFourteenHalfTurnTwoWrongDStdI}{\eFourteenHalfTurnTwoWrongCiI} \\
    Turn 3 & \eFourteenHalfTurnThreeAllArmE & \dci{\eFourteenHalfTurnThreeAllDE$\pm$\eFourteenHalfTurnThreeAllDStdE}{\eFourteenHalfTurnThreeAllCiE} & \eFourteenHalfTurnThreeWrongArmE & \dci{\eFourteenHalfTurnThreeWrongDE$\pm$\eFourteenHalfTurnThreeWrongDStdE}{\eFourteenHalfTurnThreeWrongCiE} & \eFourteenHalfTurnThreeAllArmI & \dci{\eFourteenHalfTurnThreeAllDI$\pm$\eFourteenHalfTurnThreeAllDStdI}{\eFourteenHalfTurnThreeAllCiI} & \eFourteenHalfTurnThreeWrongArmI & \dci{\eFourteenHalfTurnThreeWrongDI$\pm$\eFourteenHalfTurnThreeWrongDStdI}{\eFourteenHalfTurnThreeWrongCiI} \\
    Final scene & \eFourteenHalfFinalAllArmE & \dci{\eFourteenHalfFinalAllDE$\pm$\eFourteenHalfFinalAllDStdE}{\eFourteenHalfFinalAllCiE} & \eFourteenHalfFinalWrongArmE & \dci{\eFourteenHalfFinalWrongDE$\pm$\eFourteenHalfFinalWrongDStdE}{\eFourteenHalfFinalWrongCiE} & \eFourteenHalfFinalAllArmI & \dci{\eFourteenHalfFinalAllDI$\pm$\eFourteenHalfFinalAllDStdI}{\eFourteenHalfFinalAllCiI} & \eFourteenHalfFinalWrongArmI & \dci{\eFourteenHalfFinalWrongDI$\pm$\eFourteenHalfFinalWrongDStdI}{\eFourteenHalfFinalWrongCiI} \\
    \midrule
    \multicolumn{9}{@{}l}{\emph{Attributes stated wrongly in turn 1, in the later scenes}} \\
    $p=0.25$, turn 3 & -- & -- & \dci{\eFourteenQuarterTOneTurnThreeArmE}{(\eFourteenQuarterTOneNIntE)} & \dci{\eFourteenQuarterTOneTurnThreeDE$\pm$\eFourteenQuarterTOneTurnThreeDStdE}{\eFourteenQuarterTOneTurnThreeCiE} & -- & -- & \dci{\eFourteenQuarterTOneTurnThreeArmI}{(\eFourteenQuarterTOneNIntI)} & \dci{\eFourteenQuarterTOneTurnThreeDI$\pm$\eFourteenQuarterTOneTurnThreeDStdI}{\eFourteenQuarterTOneTurnThreeCiI} \\
    $p=0.25$, final scene & -- & -- & \dci{\eFourteenQuarterTOneFinalArmE}{(\eFourteenQuarterTOneNIntE)} & \dci{\eFourteenQuarterTOneFinalDE$\pm$\eFourteenQuarterTOneFinalDStdE}{\eFourteenQuarterTOneFinalCiE} & -- & -- & \dci{\eFourteenQuarterTOneFinalArmI}{(\eFourteenQuarterTOneNIntI)} & \dci{\eFourteenQuarterTOneFinalDI$\pm$\eFourteenQuarterTOneFinalDStdI}{\eFourteenQuarterTOneFinalCiI} \\
    $p=0.5$, turn 3 & -- & -- & \dci{\eFourteenHalfTOneTurnThreeArmE}{(\eFourteenHalfTOneNIntE)} & \dci{\eFourteenHalfTOneTurnThreeDE$\pm$\eFourteenHalfTOneTurnThreeDStdE}{\eFourteenHalfTOneTurnThreeCiE} & -- & -- & \dci{\eFourteenHalfTOneTurnThreeArmI}{(\eFourteenHalfTOneNIntI)} & \dci{\eFourteenHalfTOneTurnThreeDI$\pm$\eFourteenHalfTOneTurnThreeDStdI}{\eFourteenHalfTOneTurnThreeCiI} \\
    $p=0.5$, final scene & -- & -- & \dci{\eFourteenHalfTOneFinalArmE}{(\eFourteenHalfTOneNIntE)} & \dci{\eFourteenHalfTOneFinalDE$\pm$\eFourteenHalfTOneFinalDStdE}{\eFourteenHalfTOneFinalCiE} & -- & -- & \dci{\eFourteenHalfTOneFinalArmI}{(\eFourteenHalfTOneNIntI)} & \dci{\eFourteenHalfTOneFinalDI$\pm$\eFourteenHalfTOneFinalDStdI}{\eFourteenHalfTOneFinalCiI} \\
    \bottomrule
  \end{tabularx}
  \caption{pAMR (\%) of \methodname{} on the dining-room and office test instances when the feedback of each turn is rewritten with probability $p$ to ask for a wrong value of one attribute, and its paired difference from the clean interactions of \cref{tab:main} of the main paper ($\Delta$, percentage points); mean $\pm$ standard deviation over three seeds, with the 95\% bootstrap interval over instances below.
    Wrongly stated attributes: those drawn in a noisy turn before the scene; this set grows from turn 2 to the final scene.
    Bottom block: only the attributes drawn in a noisy turn 1, with the number of interactions below the pAMR; at turn 2, this set is that of the rows ``Turn 2''.}
  \label{S-tab:supp_noisy}
\end{table*}

The simulated user writes its feedback from its taste narrative (\cref{S-sec:supp_user}), but a real user may also state a preference that they do not hold, \eg by asking for a wall color that they do not want.
We therefore test how \methodname{} responds to feedback that contains such wrong statements.

\paragraph{Setup.}
In each turn, the feedback that \methodname{} receives is rewritten independently with probability $p$.
One of the ten attributes of the room and one value of this attribute other than the hidden value are drawn at random with a fixed seed.
Gemma-4-31B, the LLM of all systems, rewrites the feedback at temperature 0 so that it asks for the drawn value and no longer states the hidden value of this attribute, and keeps the rest of the feedback.
If the feedback does not mention the attribute, the rewrite adds a request for the drawn value.
The simulated user is unchanged: it reads the same taste narrative, and the history of its earlier turns (\cref{S-sec:supp_user}) holds the feedback that it wrote, not the rewritten feedback.
We use $p=0.25$ and $p=0.5$, the nonzero noise levels of a study of LLM-based preference elicitation~\cite{austin2024bayesian}, in which a fixed share of the simulated user's answers is replaced with random answers.
For example, in one noisy turn in the dining room, the drawn attribute is the table setting, with the hidden value \emph{Bare Table} and the drawn value \emph{Flower Centrepiece} (\cref{S-sec:supp_schemas}).
The sentence ``Also, please remove the decorative items from the center of the table; I really need the surface to stay completely clear \ldots'' is replaced by ``Also, please place a large vase of fresh flowers in the middle of the table.'', and the two sentences before it are kept word for word.\looseness=1

We run \methodname{} at both levels on the dining-room and office test instances (\eFourteenNinst{} per form) with seeds 0--2 and $K=3$, \eFourteenNinter{} interactions in all.
Of the \eFourteenNturnsRun{} turns of one form at one noise level, the random draws make \eFourteenNoisyQuarterE{} (\vE{} form) and \eFourteenNoisyQuarterI{} (\vI{} form) noisy at $p=0.25$, and \eFourteenNoisyHalfE{} and \eFourteenNoisyHalfI{} at $p=0.5$.
The control consists of the interactions of \methodname{} in \cref{tab:main} of the main paper on the same instances and seeds, whose feedback is not rewritten; we call them the clean interactions.
As in \cref{S-fig:supp_rounds}, every scene is rendered and scored by the evaluator (\cref{S-sec:supp_judge}) after the furnishing step (\cref{S-sec:supp_furnish}).

\paragraph{Check of the rewritten feedback.}
In every noisy turn, Gemma-4-31B reads the feedback that \methodname{} received and answers whether this feedback asks for the hidden value, for the drawn value or for neither.
It finds the drawn value in all \eFourteenNoisyAll{} noisy turns and the hidden value in none.
In the original feedback of the same turns, it finds the hidden value in \eFourteenOrigTrueAll{} turns and neither value in the other \eFourteenOrigNeitherAll{}, in which the rewrite adds the wrong statement.
In a second question, it reads the original and the rewritten feedback side by side and lists every request, praise or criticism outside the drawn attribute that the rewrite removed, added or changed.
It lists none in \eFourteenChkKeptAll{} of the \eFourteenNoisyAll{} noisy turns (\eFourteenChkKeptPct\%).

\paragraph{Comparison.}
Each difference is computed on the triples of instance, seed and attribute that the noisy and the clean interactions share.
It is averaged over the triples of each seed and reported as the mean $\pm$ standard deviation over the three seeds, with a 95\% percentile bootstrap interval over instances (\suppBootB{} resamples).
We report the match on all attributes and on the wrongly stated attributes, \ie the attributes drawn in a noisy turn before the scene: in turn 1 for the scene of turn 2, in turns 1 and 2 for the scene of turn 3, and in turns 1 to 3 for the final scene.
On the wrongly stated attributes, the clean interactions are scored on the same attributes.
These sets differ between the scenes, so we also follow one fixed set, the attributes drawn in a noisy turn 1, through the scenes of turns 2 and 3 and the final scene (bottom block of \cref{S-tab:supp_noisy}).
On this set, we also compute the change of the difference from the scene of turn 2 to the final scene on the same triples, with the same estimator.

\paragraph{Result.}
\Cref{S-tab:supp_noisy} gives the match.
The scenes of turn 1, which precede any feedback, differ from those of the clean interactions by \eFourteenRngTurnOneAllDLo{} points on all attributes at both levels and in both forms, and every interval contains 0.
On all attributes, the pAMR of the final scenes with noisy feedback lies \eFourteenRngFinalAllLossLo{} to \eFourteenRngFinalAllLossHi{} points below that of the clean final scenes (\eFourteenCleanFinalAllE{}\% in the \vE{} form and \eFourteenCleanFinalAllI{}\% in the \vI{} form).
The interval of this difference lies below 0 only in the \vE{} form at $p=0.5$.
The wrongly stated attributes make up \eFourteenRngFinalWrongShareLo{} to \eFourteenRngFinalWrongShareHi{}\% of the triples of the final scenes.

On the wrongly stated attributes, the loss is large directly after the noisy turn.
In the scene of turn 2, which \methodname{} builds from the plan updated with the noisy feedback of turn 1, the difference lies between \eFourteenRngTurnTwoWrongDLo{} and \eFourteenRngTurnTwoWrongDHi{} points.
In the final scenes, whose wrongly stated attributes are those drawn in turns 1 to 3, the difference lies between \eFourteenRngFinalWrongDLo{} and \eFourteenRngFinalWrongDHi{} points, and every interval of these two rows lies below 0.
The bottom block of \cref{S-tab:supp_noisy} follows the attributes drawn in turn 1, with \eFourteenRngTOneNIntLo{} to \eFourteenRngTOneNIntHi{} interactions per combination of level and form.
On this set, the difference lies between \eFourteenRngTOneTurnTwoDLo{} and \eFourteenRngTOneTurnTwoDHi{} points in the scene of turn 2, between \eFourteenRngTOneTurnThreeDLo{} and \eFourteenRngTOneTurnThreeDHi{} points in the scene of turn 3, and between \eFourteenRngTOneFinalDLo{} and \eFourteenRngTOneFinalDHi{} points in the final scene; in the final scene, the interval lies below 0 only in the \vE{} form at $p=0.5$.
From the scene of turn 2 to the final scene, the difference on these triples changes by \eFourteenRngTOneChgFinalDLo{} to \eFourteenRngTOneChgFinalDHi{} points; the interval of this change lies above 0 in three of the four combinations of level and form and contains 0 in the \vI{} form at $p=0.25$ (\eFourteenQuarterTOneChgFinalDI{} points, \mbox{\eFourteenQuarterTOneChgFinalCiI{}}).
On the attributes stated wrongly in turn 1, the later turns thus undo part of the loss of the scene of turn 2.
The simulated user keeps its own feedback in its history and writes the feedback on each scene from its taste narrative.
It can therefore object to the drawn value and ask again for its hidden value.
This setup thus models a user who misstates a preference in one turn and can object to the result in a later turn.

\begin{table*}[t]
  \centering
  \footnotesize
  \setlength{\tabcolsep}{4pt}
  \begin{tabularx}{\textwidth}{@{}l*{6}{Y}@{}}
    \toprule
    & \multicolumn{3}{c}{\vE{}} & \multicolumn{3}{c}{\vI{}} \\
    \cmidrule(lr){2-4}\cmidrule(l){5-7}
    Variant & pAMR\,$\uparrow$ & \hdtwo{$\Delta$ all}{attributes} & \hdtwo{$\Delta$ missed}{in turn 1} & pAMR\,$\uparrow$ & \hdtwo{$\Delta$ all}{attributes} & \hdtwo{$\Delta$ missed}{in turn 1} \\
    \midrule
    \methodname{} (full) & \trAblFullE & -- & -- & \trAblFullI & -- & -- \\
    \midrule
    w/o explicit plan & \anAblArmfcAllArmE & \dci{\anAblArmfcAllDE$\pm$\anAblArmfcAllDstdE}{\anAblArmfcAllCiE} & \dci{\anAblArmfcTmissedDE$\pm$\anAblArmfcTmissedDstdE}{\anAblArmfcTmissedCiE} & \anAblArmfcAllArmI & \dci{\anAblArmfcAllDI$\pm$\anAblArmfcAllDstdI}{\anAblArmfcAllCiI} & \dci{\anAblArmfcTmissedDI$\pm$\anAblArmfcTmissedDstdI}{\anAblArmfcTmissedCiI} \\
    w/o scene descriptions in the history & \anAblNoddAllArmE & \dci{\anAblNoddAllDE$\pm$\anAblNoddAllDstdE}{\anAblNoddAllCiE} & \dci{\anAblNoddTmissedDE$\pm$\anAblNoddTmissedDstdE}{\anAblNoddTmissedCiE} & \anAblNoddAllArmI & \dci{\anAblNoddAllDI$\pm$\anAblNoddAllDstdI}{\anAblNoddAllCiI} & \dci{\anAblNoddTmissedDI$\pm$\anAblNoddTmissedDstdI}{\anAblNoddTmissedCiI} \\
    w/o instantiation rules & \anAblNodrwAllArmE & \dci{\anAblNodrwAllDE$\pm$\anAblNodrwAllDstdE}{\anAblNodrwAllCiE} & \dci{\anAblNodrwTmissedDE$\pm$\anAblNodrwTmissedDstdE}{\anAblNodrwTmissedCiE} & \anAblNodrwAllArmI & \dci{\anAblNodrwAllDI$\pm$\anAblNodrwAllDstdI}{\anAblNodrwAllCiI} & \dci{\anAblNodrwTmissedDI$\pm$\anAblNodrwTmissedDstdI}{\anAblNodrwTmissedCiI} \\
    \bottomrule
  \end{tabularx}
  \caption{Further ablations on the dining-room and office test instances, final scenes after the furnishing step: pAMR (\%) on all attributes, and the difference between the variant and the full method ($\Delta$, percentage points) on all attributes and on the attributes missed in turn 1; mean $\pm$ standard deviation over four seeds, with the 95\% bootstrap interval over instances below.
    Missed in turn 1: the attributes whose hidden value is shown neither by the turn-1 scene of the full method nor by that of the variant, \ie in each of the two scenes, at least one of the two answers of the evaluator does not name it; this set differs between the rows.}
  \label{S-tab:supp_more_ablations}
\end{table*}

In summary, a wrong statement strongly lowers the match of its attribute in the next scene.
On the attributes stated wrongly in turn 1, the later turns undo part of this loss, with an interval of the change above 0 in three of the four combinations of level and form.
The match on all attributes changes much less.

\subsection{Further ablations}
\label{S-sec:supp_more_ablations}

We remove further components of \methodname{}: the explicit plan, the scene descriptions in the history of the plan update, and the instantiation rules.
As in \cref{tab:ablation} of the main paper, each variant runs on the dining-room and office test instances (24 per form) with $K=3$ and seeds 0--3 and is scored, as the full method is, on its final scene after the furnishing step (\cref{S-sec:supp_furnish}).
The row of the full method uses the interactions of \methodname{} in \cref{tab:main} of the main paper on the same instances and seeds, as in \cref{tab:ablation}.
\begin{itemize}
  \item \emph{Without the explicit plan}, the LLM writes a free-text description of the room wherever the full method writes a plan. A separate LLM call turns this description into one value per attribute (a candidate, a new value or the default), which the same scene instantiation builds. This call receives the description, the attributes with their candidate values and the rules for a value of the plan update (\cref{S-sec:supp_start}), but not the input or the history. The input, the candidate values, the scene descriptions in the history and the furnishing step are the same as in the full method.
  \item \emph{Without the scene descriptions in the history}, the history of the plan update (\cref{eq:update} of the main paper) lists for every earlier turn the value shown for each attribute, the feedback and the score, but not the scene~description~$D_{k'}$~(\cref{S-sec:supp_descr}).
  \item \emph{Without the instantiation rules}, scene instantiation and the furnishing step build each plan without the rules of \cref{S-sec:supp_rules}, \ie the rules on lighting, materials, object counts and placement, and colors.
\end{itemize}
An attribute of an interaction counts as missed in turn 1 when neither the turn-1 scene that the full method showed in the interaction nor that of the variant shows its hidden value, \ie when in each of the two scenes at least one of the two answers of the evaluator (\cref{S-sec:supp_judge}) does not name the hidden value.
This set differs between the variants.
Each difference from the full method in \cref{S-tab:supp_more_ablations} is computed on the triples of instance, seed and attribute that both rows have, and given as the mean $\pm$ standard deviation over the four seeds with a 95\% percentile bootstrap interval over instances (\suppBootB{} resamples).

Without the explicit plan, the final scenes match fewer hidden values than those of the full method in both forms, although both use the same candidate values.
The difference is \anAblArmfcAllDE{} points in the \vE{} form and \anAblArmfcAllDI{} points in the \vI{} form on all attributes and \anAblArmfcTmissedDE{} and \anAblArmfcTmissedDI{} points on the attributes missed in turn 1, and all four intervals lie below 0.
With the same candidate values, the explicit plan therefore raises the match.

Without the scene descriptions in the history, the difference is \anAblNoddAllDE{} points in the \vE{} form and \anAblNoddAllDI{} points in the \vI{} form on all attributes, and \anAblNoddTmissedDE{} and \anAblNoddTmissedDI{} points on the attributes missed in turn 1.
Without the instantiation rules, the difference is \anAblNodrwAllDE{} and \anAblNodrwAllDI{} points on all attributes, and \anAblNodrwTmissedDE{} and \anAblNodrwTmissedDI{} points on the attributes missed in turn 1.
All eight intervals of these two variants contain 0.

\section{User study}
\label{S-sec:supp_human}

Both studies of this section, the pairwise study and the study with real users (\cref{S-sec:supp_real_users}), were approved by an institutional review board, and all participants gave informed consent before the session.

\paragraph{Participants.}
\hsN{} participants took part, aged \hsAgeMin{}--\hsAgeMax{} (mean \hsAgeMean{}).
We analyze one completed session per participant; five incomplete sessions and one session on an earlier set of trials are excluded.

\paragraph{Comparisons.}
Each participant completed 36 trials, six for each of the six baselines, so that each comparison has \hsPerSys{} answers per question.
For each comparison, every test instance of both forms enters a pool with one of the four seeds of \cref{tab:main} of the main paper, drawn at random (for LayoutGPT, the bedrooms and living rooms only), and the trials of each participant are drawn at random from these pools.
A trial showed the instruction, the persona and the taste narrative of a test instance above the final scenes of \methodname{} and of the baseline, with the system names hidden and the sides randomized.
The questions were ``Which room better matches this person's taste?'' and ``Regardless of this person's taste, which room is better designed overall?''
\Cref{S-tab:supp_human} gives the share of answers that chose \methodname{}, with 95\% Wilson intervals and two-sided binomial tests against 50\% with the Benjamini-Hochberg correction over the 12 tests, as~in~\mbox{SceneSmith}~\cite{pfaff2026scenesmith}.
When the answers are grouped by participant, every share stays above 50\%: a bootstrap over the \hsN{} participants gives 95\% intervals whose lower ends are at least 61.7\%, and a logistic mixed model with a random intercept per participant gives Benjamini-Hochberg-corrected $p<0.001$ in every comparison.

\paragraph{Ratings.}
Each participant then rated 24 single scenes in three blocks of eight, one block for each criterion (realism, functionality and instruction following), on a five-point scale with the rubric of R3L~\cite{gu2026r3l}.
The \hsRatedScenes{} rated scenes are drawn at random from the final scenes of all seven systems.
We map the ratings to 1--10, as R3L does, and compare them with the scores of the evaluator of \cref{tab:quality} of the main paper on the same scenes (\cref{S-tab:supp_human_rate}).
Kendall's $\tau_b$ between the mean rating of a scene and its evaluator score is \hsTauReal{}, \hsTauFunc{} and \hsTauInstr{} for the three criteria, and \hsTauUUReal{}, \hsTauUUFunc{} and \hsTauUUInstr{} between two random halves of the raters of each scene.

\begin{table*}[t]
  \centering
  \footnotesize
  \setlength{\tabcolsep}{6pt}
  \begin{tabularx}{\linewidth}{@{}X *{3}{c} *{3}{c}@{}}
\toprule
 & \multicolumn{3}{c}{Preferences} & \multicolumn{3}{c}{Design} \\
\cmidrule(lr){2-4}\cmidrule(l){5-7}
\methodname{} vs. & Win (\%) & 95\% CI & $p_{\text{BH}}$ & Win (\%) & 95\% CI & $p_{\text{BH}}$ \\
\midrule
Holodeck & \hsWinHolodeckQOne & \hsCIHolodeckQOne & \hsPHolodeckQOne & \hsWinHolodeckQTwo & \hsCIHolodeckQTwo & \hsPHolodeckQTwo \\
LayoutGPT & \hsWinLayoutgptQOne & \hsCILayoutgptQOne & \hsPLayoutgptQOne & \hsWinLayoutgptQTwo & \hsCILayoutgptQTwo & \hsPLayoutgptQTwo \\
I-Design & \hsWinIdesignQOne & \hsCIIdesignQOne & \hsPIdesignQOne & \hsWinIdesignQTwo & \hsCIIdesignQTwo & \hsPIdesignQTwo \\
LayoutVLM & \hsWinLayoutvlmQOne & \hsCILayoutvlmQOne & \hsPLayoutvlmQOne & \hsWinLayoutvlmQTwo & \hsCILayoutvlmQTwo & \hsPLayoutvlmQTwo \\
R3L & \hsWinRThreelQOne & \hsCIRThreelQOne & \hsPRThreelQOne & \hsWinRThreelQTwo & \hsCIRThreelQTwo & \hsPRThreelQTwo \\
Holodeck-Edit & \hsWinHolodeckEditQOne & \hsCIHolodeckEditQOne & \hsPHolodeckEditQOne & \hsWinHolodeckEditQTwo & \hsCIHolodeckEditQTwo & \hsPHolodeckEditQTwo \\
\midrule
All & \hsWinAllQOne & 76.6--82.5 & & \hsWinAllQTwo & 77.8--83.5 & \\
\bottomrule
\end{tabularx}

  \caption{User study, comparisons: share of the answers that chose \methodname{} over the baseline (\%, $\uparrow$), 95\% Wilson interval and Benjamini-Hochberg-corrected $p$ of a two-sided binomial test against 50\%, for the question on the person's preferences and the question on the design; \hsPerSys{} answers per comparison and question.}
  \label{S-tab:supp_human}
\end{table*}

\begin{table*}[t]
  \centering
  \footnotesize
  \setlength{\tabcolsep}{6pt}
  \begin{tabularx}{\linewidth}{@{}X *{3}{c} *{3}{c}@{}}
\toprule
 & \multicolumn{3}{c}{Participants} & \multicolumn{3}{c}{Evaluator (same scenes)} \\
\cmidrule(lr){2-4}\cmidrule(l){5-7}
System & Real. & Func. & Instr. & Real. & Func. & Instr. \\
\midrule
Holodeck & 4.8 & 5.1 & 4.4 & 7.8 & 7.2 & 7.4 \\
LayoutGPT & 3.2 & 3.7 & 3.6 & 7.1 & 6.8 & 7.0 \\
I-Design & 3.8 & 3.6 & 4.9 & 7.3 & 6.8 & 7.2 \\
LayoutVLM & 3.2 & 3.5 & 4.1 & 7.8 & 7.1 & 7.7 \\
R3L & \underline{7.3} & \underline{7.4} & 5.9 & \textbf{8.5} & \textbf{7.7} & \underline{8.4} \\
Holodeck-Edit & 7.2 & 6.6 & \underline{7.0} & 7.8 & \underline{7.3} & 7.6 \\
\methodname{} (Ours) & \textbf{8.2} & \textbf{7.6} & \textbf{8.6} & \underline{8.3} & \textbf{7.7} & \textbf{8.8} \\
\midrule
Kendall's $\tau_b$ with the evaluator & \hsTauReal & \hsTauFunc & \hsTauInstr & & & \\
Two halves of the participants & \hsTauUUReal & \hsTauUUFunc & \hsTauUUInstr & & & \\
One half of the participants and the evaluator & \hsTauULReal & \hsTauULFunc & \hsTauULInstr & & & \\
\bottomrule
\end{tabularx}

  \caption{User study, ratings: mean rating of the participants (five-point scale mapped to 1--10, $\uparrow$) and mean score of the evaluator of \cref{tab:quality} of the main paper on the same \hsRatedScenes{} rated scenes, per system and criterion. Bottom: Kendall's $\tau_b$ over the scenes between the mean rating and the evaluator score, between two random halves of the raters of each scene (mean of 200 splits), and between one half and the evaluator. Bold: best; underlined: second best.}
  \label{S-tab:supp_human_rate}
\end{table*}

\subsection{Interaction with real users}
\label{S-sec:supp_real_users}
In all other experiments, the feedback comes from the simulated user, which writes it from a taste narrative.
Here we test whether people can direct the interaction with their own feedback.
Participants design two rooms for themselves with \methodname{} and write the instruction and the feedback in their own words.

\paragraph{Participants.}
In total, \ruN{} participants took part online.
We analyze one completed session per participant; \ruIncomplete{} incomplete sessions are excluded.
The session pages are in Korean and English, and all participants wrote their texts in Korean.

\paragraph{Session.}
The participant first chooses two of eight room types of the benchmark: the bathroom, the bedroom, the children's room, the gallery, the kitchen, the library, the living room and the waiting room.
For each room, the participant writes an instruction in their own words and fills in a short form on their age range, their household and the use of the room.
This form takes the place of the persona of a benchmark instance.
\methodname{} then runs an interaction of $K=3$ turns.
In each turn, the participant sees the scene in the cutaway and the eye-level view (\cref{S-sec:supp_render}), writes free-text feedback as they would give it to a designer, and answers ``How satisfied are you with this room?'' with a score from 1 (``not at all'') to 10 (``completely'').
From turn 2 on, the participant also states whether the room became worse, stayed about the same or became better than in the previous turn.
\methodname{} receives the feedback and the score as it receives those of the simulated user.

\paragraph{System.}
The participants interact only with \methodname{}, which uses Gemma-4-31B as its LLM, Jev as its decision model and the attribute schema of the chosen room type, as in the main paper.
As in the interaction with the simulated user, the scene of each turn is shown before the furnishing step, which follows the last turn (\cref{S-sec:supp_furnish}).
The consent page and the instruction page state that the participant's texts are sent to an external AI service to generate the rooms, and ask the participant not to write personal information such as names or contact details.
To keep the waiting time short, the scenes that the participants see are rendered with the preview settings of the renderer ($768\times576$ pixels, at most 64 samples per pixel), from the same two cameras as in the main experiments.

\begin{figure*}[t]
  \centering
  \footnotesize
  \setlength{\tabcolsep}{5pt}
  \begin{tabular}{@{}m{0.285\textwidth}m{0.37\textwidth}m{0.285\textwidth}@{}}
    \includegraphics[width=\linewidth]{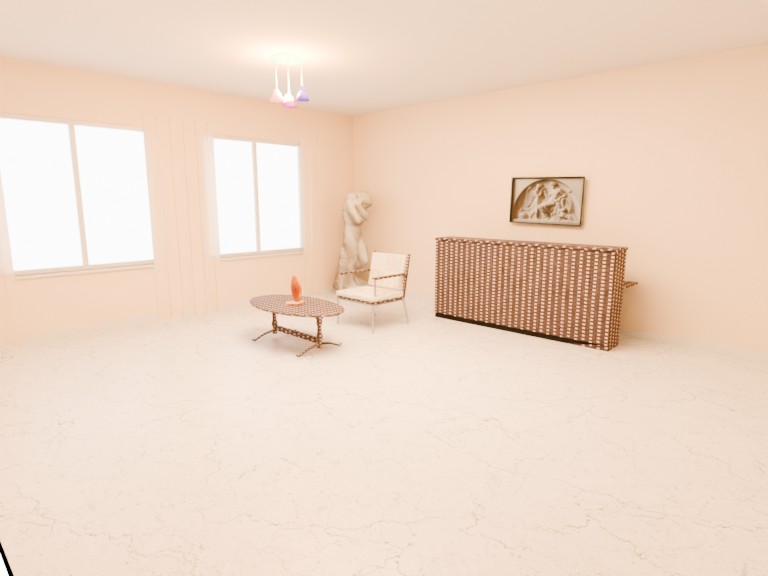}\par\centering Turn 1, satisfaction 4 &
    \textbf{Waiting room.} Instruction: ``A luxurious waiting room in the style of a Michelin-starred restaurant.''\par\smallskip
    Feedback on turn 1: ``Not bad for a start. There is quite a lot of empty space, so how it gets filled will matter from here on. It must not feel bare.''\par\smallskip
    Feedback on turn 2: ``Oh, it has become much better. Let us just refine the details in this direction.'' &
    \includegraphics[width=\linewidth]{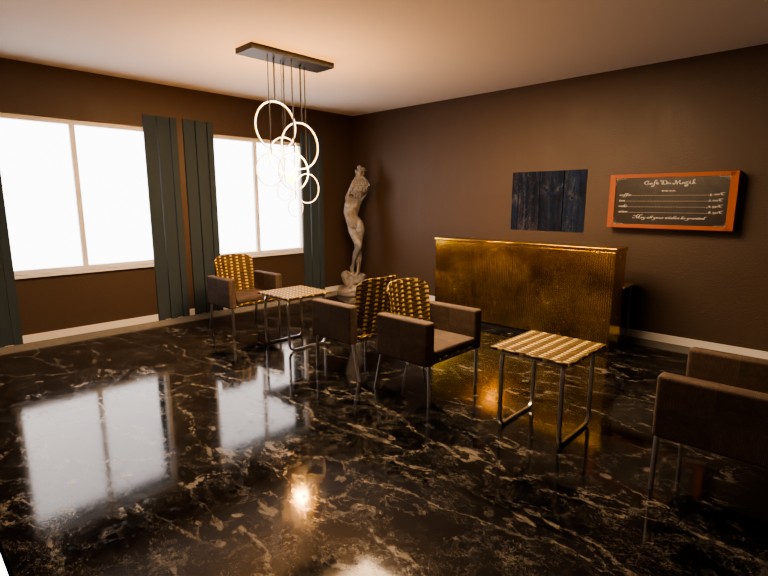}\par\centering Turn 2, satisfaction 8, ``better'' \tabularnewline[9pt]
    \includegraphics[width=\linewidth]{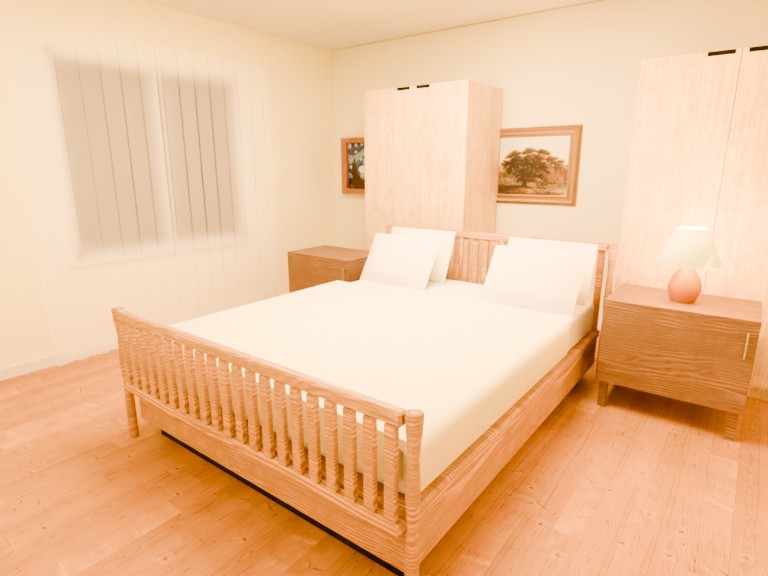}\par\centering Turn 1, satisfaction 6 &
    \textbf{Bedroom.} Instruction: ``I would like a wide bed and a mood lamp. I want the room to feel cozy. I prefer wooden furniture.''\par\smallskip
    Feedback on turn 1: ``I like the atmosphere, but if furniture stands in front of the wardrobe, it cannot be used, can it? Please adjust the positions. The wardrobe also hides the painting. [\dots] I would like the bedding dark brown instead of white. Please put one plush toy in the room.''\par\smallskip
    Feedback on turn 2: ``Good, clearing the front of the wardrobe and moving the painting seem good. [\dots]'' &
    \includegraphics[width=\linewidth]{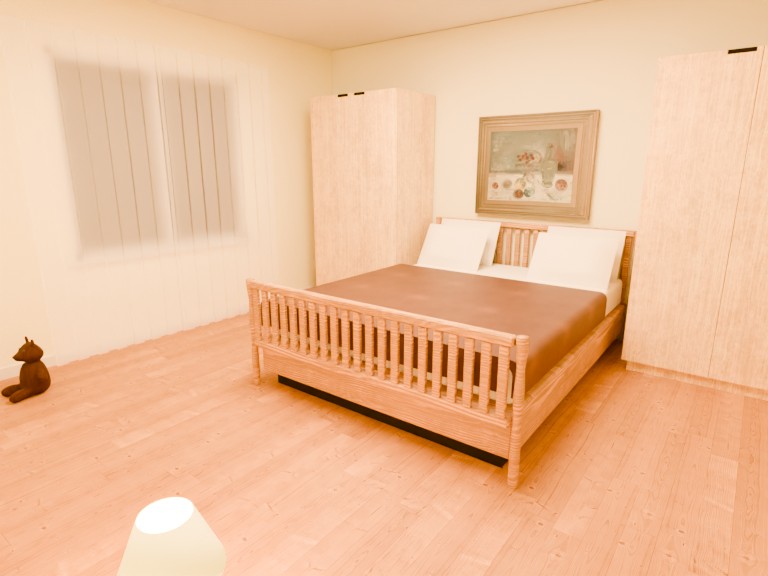}\par\centering Turn 2, satisfaction 7, ``better'' \tabularnewline[9pt]
    \includegraphics[width=\linewidth]{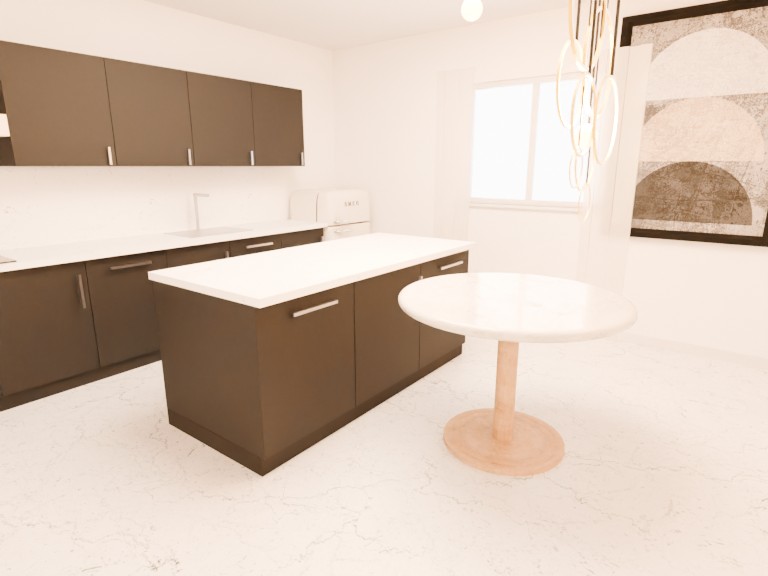}\par\centering Turn 2, satisfaction 6 &
    \textbf{Kitchen.} Instruction: ``A luxurious, modern design. A combination of black and white marble. Furniture with clean, modern finishes.''\par\smallskip
    Feedback on turn 2: ``It seems the chairs were not added? Also, the table should stand a little apart, but it is too close to the other furniture. Then no chairs can be added around the table, can they?''\par\smallskip
    Feedback on turn 3: ``Not bad. Now the chairs only need to be placed so that they surround the table well.'' &
    \includegraphics[width=\linewidth]{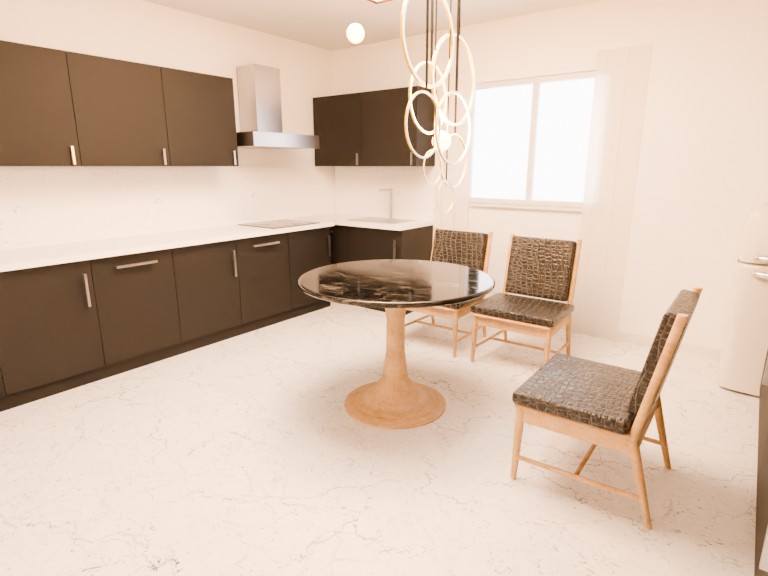}\par\centering Turn 3, satisfaction 7, ``better'' \tabularnewline
  \end{tabular}
  \caption{Interaction with real users: three turns of three rooms designed by two participants. Left: the scene that the participant saw (eye-level view, preview render); middle: the participant's instruction and feedback on both scenes, translated from Korean, with [\dots] marking omitted sentences; right: the scene of the next turn. Satisfaction is the participant's score from 1 to 10, and ``better'' is the participant's comparison with the previous turn.}
  \label{S-fig:supp_real_users}
\end{figure*}

\paragraph{Result.}
Of the \ruTrans{} judgments from turn 2 on, \ruBetter{} state that the room became better than in the previous turn, \ruSame{} that it stayed about the same and \ruWorse{} that it became worse.
The mean satisfaction score is \ruSa{} in turn 1, \ruSb{} in turn 2 and \ruSc{} in turn 3, and it is higher in turn 3 than in turn 1 in \ruRoomsUp{} of the \ruRooms{} rooms.
\Cref{S-fig:supp_real_users} shows three turns in which the next scene carries out the participant's feedback, together with the participant's feedback on that next scene.
With their own feedback, the participants raised the mean satisfaction in every turn, and they judged a turn worse than the previous one in only \ruWorse{} of \ruTrans{} cases.

\section{Qualitative results and failure cases}
\label{S-sec:supp_qual}

\begin{figure*}[t]
  \centering
  \includegraphics[width=\textwidth]{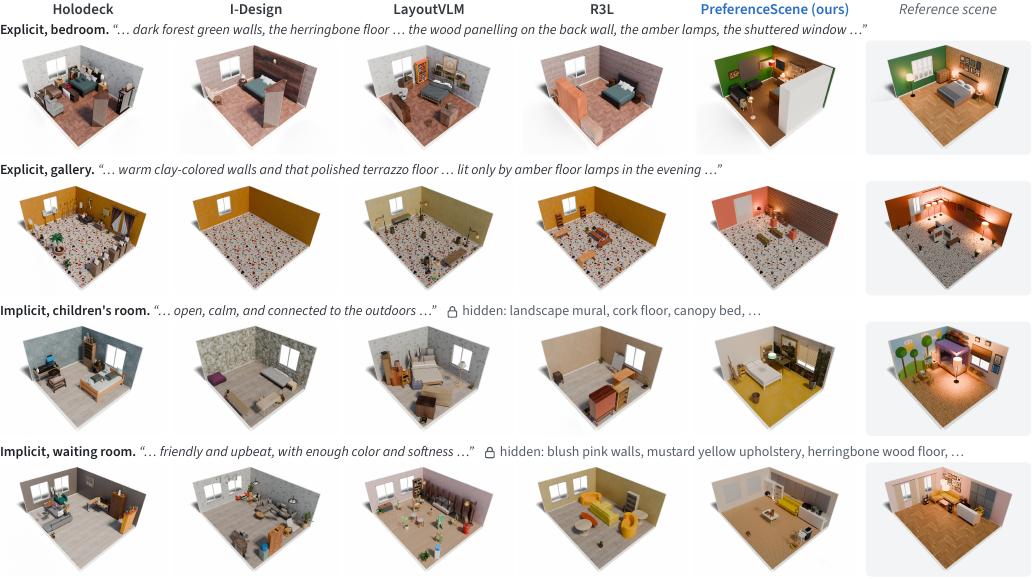}
  \caption{Final scenes after $K=3$ turns on four further test instances, in the layout of \cref{fig:qual} of the main paper.
    pAMR (\%) of the shown scenes, in column order: bedroom 40, 45, 50, 35, 90, 95; gallery 35, 25, 35, 55, 70, 100; children's room 15, 20, 25, 20, 50, 85; waiting room 20, 25, 20, 30, 55, 85.}
  \label{S-fig:supp_qual}
\end{figure*}

\Cref{S-fig:supp_qual} shows four further test instances in the layout of \cref{fig:qual} of the main paper.
LayoutGPT is not shown because it covers only bedrooms and living rooms, and three of the four instances are of other room types.

\begin{figure*}[t]
  \centering
  \newcommand{\failpanel}[2]{\begin{minipage}[t]{0.32\textwidth}\centering
    #1\\[3pt]
    \begin{minipage}[t]{\linewidth}\raggedright\scriptsize #2\end{minipage}\end{minipage}}%
  \newcommand{\failkey}[1]{\textcolor{black!60}{#1}}%
  \failpanel{\includegraphics[width=\linewidth]{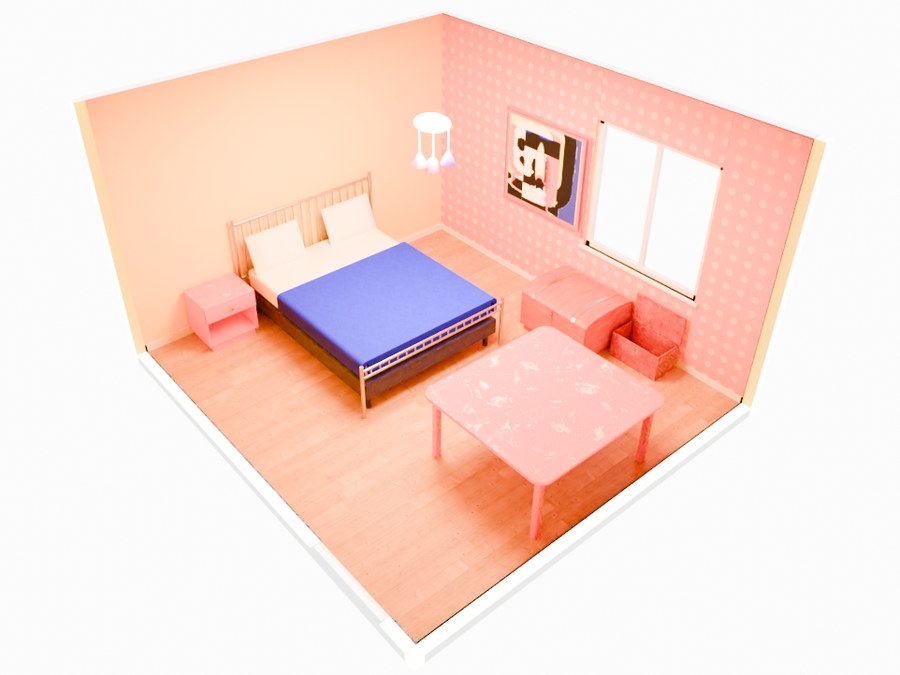}}{%
    \textbf{(a) Value not shown in the scene.} \vE{}, children's room, wall color.\\[1pt]
    \failkey{Hidden value:} Sunny Yellow.\\
    \failkey{Plan, all four scenes:} ``Sunny Yellow''.\\
    \failkey{Evaluator:} Blush Pink.}\hfill
  \failpanel{\includegraphics[width=\linewidth]{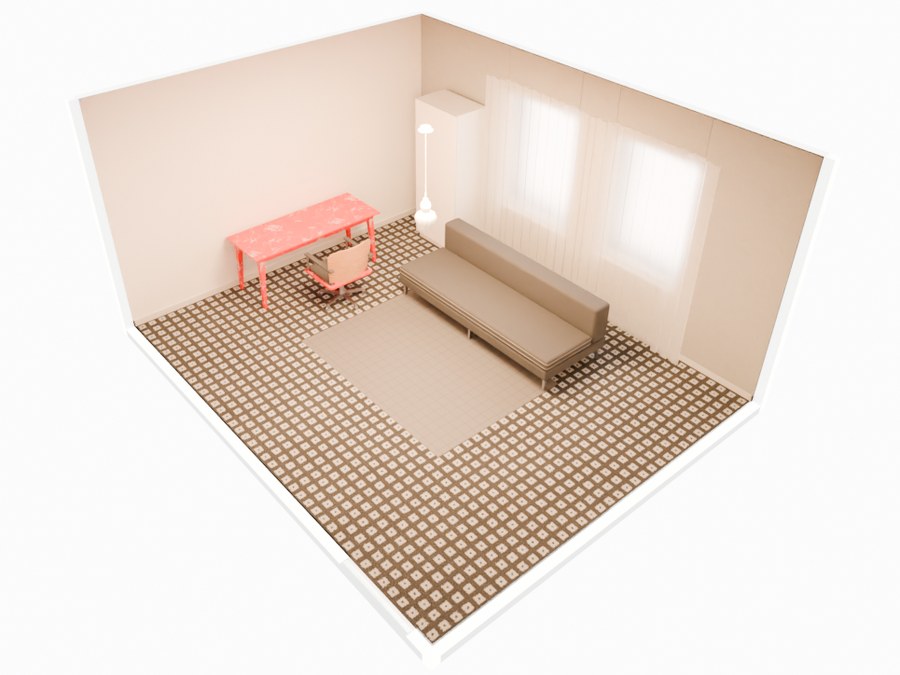}}{%
    \textbf{(b) Shown value replaced.} \vI{}, office, wall color.\\[1pt]
    \failkey{Hidden value:} Blush Pink.\\
    \failkey{Plan, turn 2:} ``Pale Blush''. \failkey{Evaluator, turn 2:} Blush Pink.\\
    \failkey{Feedback, turn 2:} no sentence on the wall color.\\
    \failkey{Plan, turn 3 and final:} ``Bright White''.\\
    \failkey{Evaluator:} Warm White.}\hfill
  \failpanel{\includegraphics[width=\linewidth]{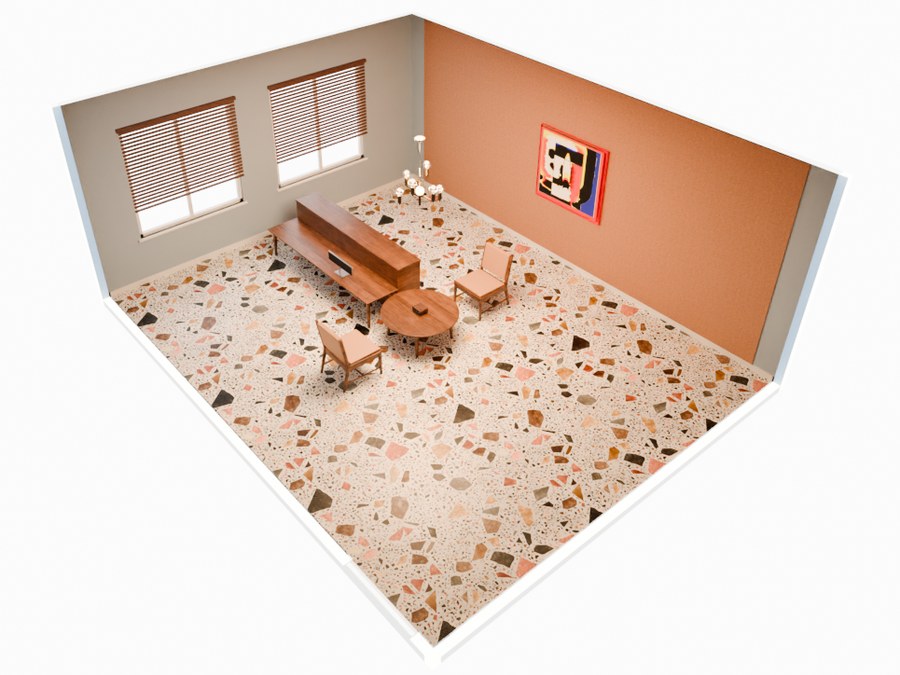}}{%
    \textbf{(c) Attribute never mentioned.} \vE{}, waiting room, window treatment.\\[1pt]
    \failkey{Hidden value:} Bamboo Roman Shades.\\
    \failkey{Plan, all four scenes:} ``Wooden Venetian''.\\
    \failkey{Instruction, persona and feedback:} no sentence on the windows.\\
    \failkey{Evaluator:} Venetian Blinds.}
  \caption{Representative failure cases of \methodname{}.
    Each panel shows the final scene of a test interaction and, for one attribute, the hidden value, the plan values, the feedback and the answer of the evaluator on the final scene; this answer is the same in both option orders.
    (a) The plans hold the hidden value, but the scene does not show it.
    (b) A plan update replaces a value for which the answer of the evaluator was the hidden value.
    (c) No feedback mentions the attribute, and the plans keep a value that differs from the hidden value.}
  \label{S-fig:supp_fail}
\end{figure*}

\begin{figure*}[t]
  \centering
  \includegraphics[width=\textwidth]{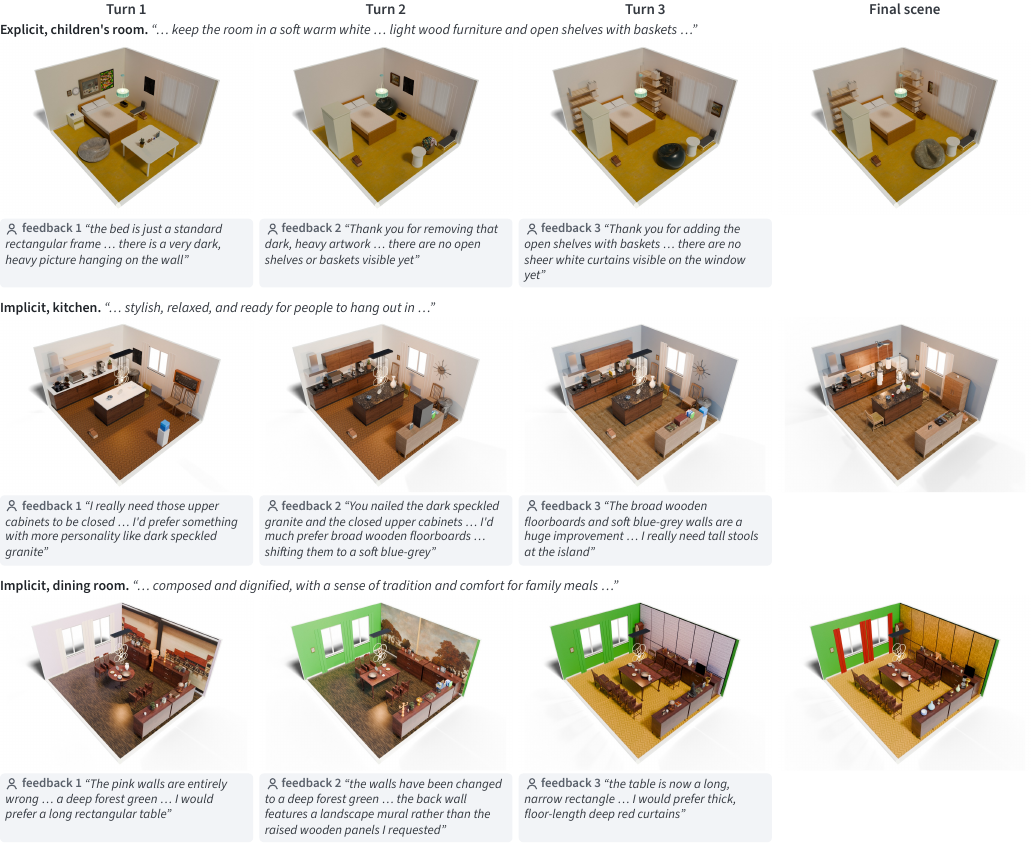}
  \caption{Three test interactions of \methodname{} over the $K=3$ turns.
    Each row gives the form, the room type and an excerpt of the instruction.
    Below the scene of each turn is an excerpt of the feedback that the simulated user wrote on that scene; ``\dots'' marks omitted text.
    The final scene is built after the third feedback.
    As in \cref{S-fig:supp_rounds}, every scene is shown and scored after the furnishing step (\cref{S-sec:supp_furnish}).
    The simulated user saw the scenes of turns 1--3 without this step.
    pAMR (\%) of the shown scenes, in column order: children's room 40, 50, 65, 75; kitchen 25, 50, 65, 80; dining room 25, 35, 60, 75.}
  \label{S-fig:supp_turns}
\end{figure*}

\begin{figure*}[!t]
  \centering
  \includegraphics[width=\textwidth]{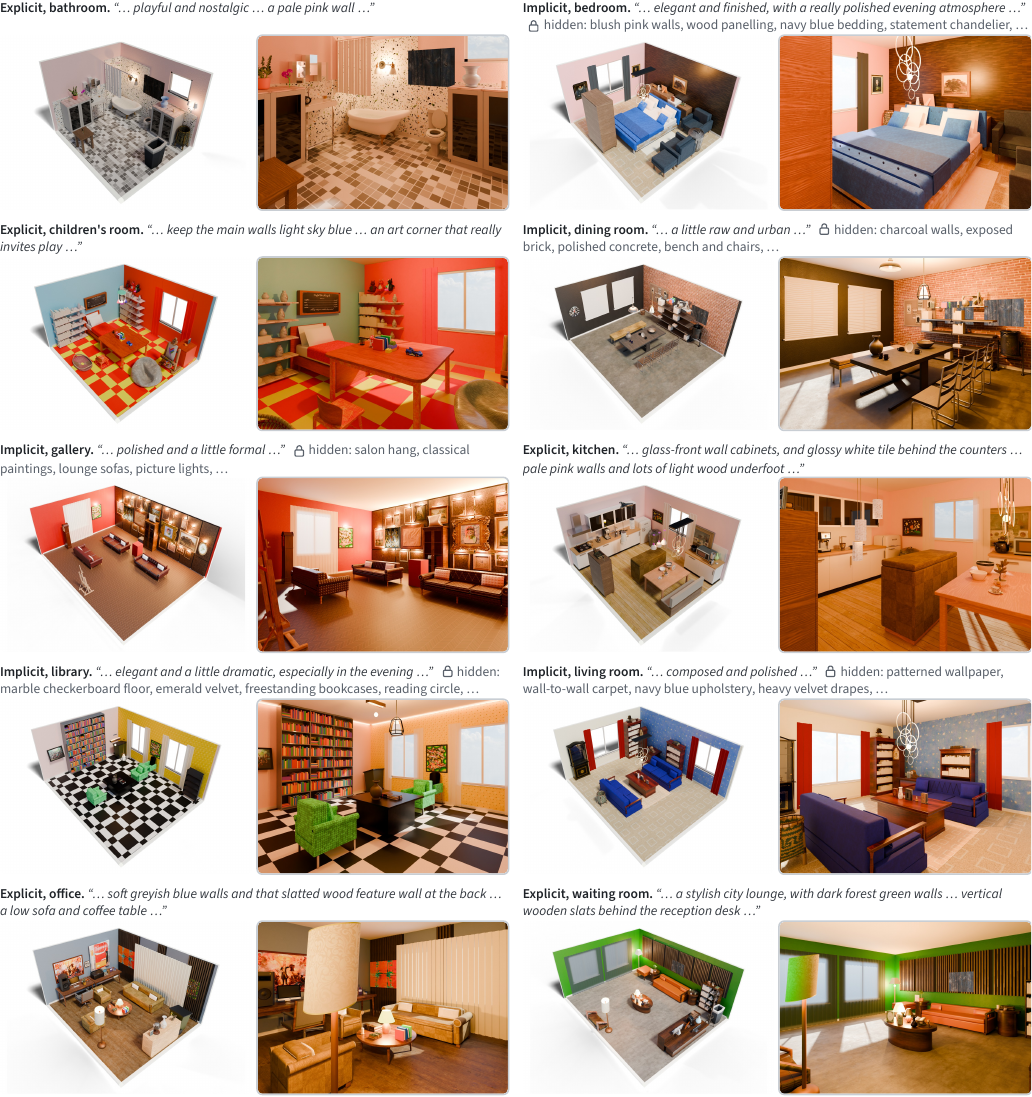}
  \caption{Final scenes of \methodname{} after $K=3$ turns, one test instance per room type, in the cutaway and the eye-level view.
    Each panel gives the form, the room type and an excerpt of the instruction, and each \vI{} panel also hidden values that the instruction does not state.
    pAMR (\%) of the shown scenes: bathroom 50, bedroom 85, children's room 45, dining room 65, gallery 75, kitchen 90, library 80, living room 70, office 75, waiting room 50.}
  \label{S-fig:supp_gallery}
\end{figure*}

\Cref{S-fig:supp_turns} follows three test interactions of \methodname{} over the $K=3$ turns, with an excerpt of the feedback of the simulated user (\cref{S-sec:supp_user}) in every turn.
In each of the three rows, the feedback of turns 2 and 3 acknowledges changes that the feedback of the previous turn asked for and then asks for further changes.
\Cref{S-fig:supp_gallery} shows the final scene of one test interaction for each of the ten room types.

\subsection{Failure cases}
\label{S-sec:supp_fail}

\Cref{S-fig:supp_fail} shows three representative failure cases of \methodname{}, each on one attribute of a test interaction.
In (a), all four plans hold the hidden value of the wall color, and the simulated user asks for yellow walls in every turn.
The walls carry the yellow paint of the plan, but under the bright warm light of this scene they appear pale peach when rendered.
In (b), the evaluator identifies the hidden wall color in the scene of turn 2.
The feedback of turn 2 does not mention the wall color, and the plan of turn 3 changes the wall color to ``Bright White''.
In (c), the instruction, the persona and the feedback of the three turns do not mention the windows.
The three later plans keep the window treatment of the initial plan, which differs from the hidden value.

\section{Limitations}
\label{S-sec:supp_limitations}

We restrict our study to one session of $K=3$ turns per user and to single rooms.
Longer sessions, preferences that change from one session to the next, whole homes and households of several people remain future work.

\section{Related work (full version)}
\label{S-sec:supp_related}

This section extends \cref{sec:related} of the main paper.

\paragraph{Indoor scene synthesis.}
Early methods optimize design-guideline costs~\cite{merrell2011interactive,yu2011makeithome}, and data-driven methods~\cite{paschalidou2021atiss,tang2024diffuscene,lin2024instructscene,yang2024physcene} learn layouts from 3D-FRONT~\cite{fu20213dfront}.
Recent methods use an LLM to predict a layout~\cite{feng2023layoutgpt}, to propose constraints for an optimizer~\cite{yang2024holodeck,littlefair2025flairgpt,sun2025layoutvlm}, or to reason about spatial relations~\cite{gu2026r3l}. Other methods combine such steps in multi-step pipelines~\cite{celen2025idesign,pun2026hsm,jiang2026hoglayout}, use generated images to guide the layout~\cite{wang2024architect,ling2026scenethesis}, or iterate with their own critic~\cite{hu2024scenecraft,xia2026sage,pfaff2026scenesmith,yang2025sceneweaver,ran2025directlayout}.
In the text-driven methods, the input text defines the target scene, and the system decides every unspecified aspect.\looseness=-1

\paragraph{Scene editing and personalization.}
Editing methods change a scene as instructed, through language commands~\cite{zheng2025editroom,bucher2025respace,noh2026editasact}, a multi-turn chat~\cite{wang2026chat2layout} or requests that an agent clarifies~\cite{liu2026designagent3d}.
I-Design~\cite{celen2025idesign} conditions on the design preferences that the user writes in the request.
OptiScene~\cite{yang2025optiscene} and ReSpace~\cite{bucher2025respace} train a single model for all users on preference pairs built from annotator-approved layouts and from automatically computed rewards, respectively.
Zhang~\etal~\cite{zhang2023interactive} recommend interior decoration designs, which are images from a database, and revise the recommendation from textual feedback with a learned policy.
SPHERE~\cite{lee2026sphere} learns spatial layout preferences in virtual reality from the user's speech and from the user's own edits of the generated scenes.
In contrast, our system infers preferences that one user does not state in the input, from free-form feedback on rendered images of full 3D scenes, and the user never edits a scene.

\paragraph{Inferring preferences from feedback.}
Interactive image retrieval and generation approach a target that the user has in mind using relative attributes~\cite{kovashka2012whittlesearch}, natural-language feedback~\cite{guo2018dialog,wu2021fashioniq}, queries~\cite{tan2019drilldown}, answers to questions~\cite{levy2023chatir} or choices among generated images~\cite{nabati2025pasta}.
Critiquing-based recommenders revise recommendations from user critiques~\cite{chen2012critiquing,wu2019deep}, and dialogue-state tracking updates an estimate of the user's goal at each turn~\cite{williams2013dialog}.
Collaborative LLM agents act on the interaction history over multiple turns~\cite{zhou2025sweetrl}.
Preference learning fits a reward model or a policy to comparisons~\cite{christiano2017deep,rafailov2023direct}.
It also infers a user's preferences from actively selected comparison queries~\cite{sadigh2017active,biyik2018batch,biyik2022aprel,gonzalez2017preferential,astudillo2023qeubo,handa2024bayesian}, choices among generated options~\cite{koyama2020sequential,rajagopalan2026personalized}, edits or demonstrations~\cite{gao2024aligning,arocaouellette2024predict}, and natural-language feedback or answers to questions~\cite{sumers2021learning,li2025eliciting,hahn2025proactive}. Reward-rational choice interprets such signals as choices that reveal a reward~\cite{jeon2020reward}.
Interactive evolutionary computation optimizes designs with the user's evaluations of candidate designs as the fitness~\cite{takagi2001interactive}; for 3D rooms, it optimizes the placement of the furniture from the user's evaluations of candidate layouts~\cite{akase2013automatic,akase2014web}.
Self-Refine~\cite{madaan2023selfrefine} and MINT~\cite{wang2024mint} pass natural-language feedback to an LLM that revises its output. Our plan serves as a dialogue state over the design attributes of a room, and every plan is instantiated as a 3D scene before the user gives feedback on it.

\paragraph{Simulated users and evaluators.}
Work that learns or elicits user preferences evaluates with simulated users that hold a hidden target~\cite{gao2024aligning,hahn2025proactive,zhang2023interactive}. User simulation for evaluation is surveyed in~\cite{balog2025user}.
LLM evaluators favor their own generations~\cite{panickssery2024llm}, and the MT-Bench study reports signs of this bias for some evaluators~\cite{zheng2023judging}.
In the main evaluation, we therefore use a different model family for each role.
Scene synthesis papers score rendered images with vision-language model (VLM) evaluators~\cite{sun2025layoutvlm,yang2025sceneweaver,yang2025optiscene,gu2026r3l}, and SceneEval checks each requirement of the input text against the scene~\cite{tam2026sceneeval}.
Like the question-answering metrics TIFA~\cite{hu2023tifa} and DSG~\cite{cho2024davidsonian}, our metric instead asks a VLM a multiple-choice question about each attribute on the rendered images, with the hidden preferences as the answer key.

\lstdefinestyle{promptP}{style=prompt, aboveskip=0pt, belowskip=0pt}
\lstdefinestyle{jsonP}{style=json, aboveskip=0pt, belowskip=0pt}
\lstdefinestyle{jsoncol}{style=jsonP, frame=none, backgroundcolor={}, xleftmargin=0pt, xrightmargin=0pt}
\clearpage\onecolumn\raggedbottom
\phantomsection\addcontentsline{toc}{section}{\texorpdfstring{\protect\tocnonum}{}Prompts and example outputs}
\section*{Prompts and example outputs}
The listings give the prompts of the benchmark construction, the simulated user and the evaluator (\cref{S-sec:supp_bench}), the prompts of \methodname{} (\cref{S-sec:supp_method}) and examples of specifications, scene descriptions and furnishing lists (\cref{S-sec:supp_examples}), in the order in which these sections cite them.
The prompts use the following words for terms of the paper: \emph{brief} for the room type, the instruction and the persona, \emph{query} for instruction, \emph{part} and \emph{topic} for attribute, \emph{option} for value, \emph{realizer} for scene instantiation, ``what the realizer drew'' for scene description, \emph{round} for turn and \emph{critique} for feedback.

\noindent\begin{minipage}{\linewidth}
\begin{lstlisting}[style=promptP,caption={Prompt for writing the profiles of a room type.},label={S-lst:profile}]
You are designing a benchmark for personalized indoor scene synthesis.

For the scene type "{scene_type}", generate 8 distinct preference profiles (G1-G8) using the SCHEMA below. SCHEMA lists {n} discrete preference attributes; for each, what it is about and its options as "label: description".

Each profile is one person's coherent interior taste: the room a real person with a clear, consistent taste would want, put together the way a competent interior designer would design it for that person. All values of a profile must belong to the same design.

For each profile, work in this order:
1. First choose the person's overall design style and target mood for this {scene_type}. Give each a short label and one sentence of description.
2. Then choose one option for every attribute, in the order listed in SCHEMA. Every attribute is part of this person's taste, including layout, arrangement, seating, fixtures, storage and decoration: each value must follow from the style and mood and fit the values already chosen for this profile; this holds for colours too (choose the wall colour for this style and mood). For each value write one sentence: why it fits this person's style, mood and other choices.

Constraint: any two profiles must differ in at least 50% of the attributes (at least {min_dist} of {n}).
Make the profiles represent genuinely different design preferences — not just random combinations, but coherent, realistic design visions. Profiles differ because their styles and moods differ: do not change a value only to make two profiles differ.
Use only option labels from SCHEMA, copied exactly (the text before the colon).

SCHEMA: {schema}

Output ONLY valid JSON in the following format (no extra text):
{"scene_type": "{scene_type}", "g_candidates": [{"g_id": "{room}_g1", "design_style": {"label": "...", "description": "..."}, "target_mood": {"label": "...", "description": "..."}, "attributes": {"<attribute>": {"value": "<option label>", "reason": "..."}, ...}}, ...]} with exactly 8 profiles, g_id {room}_g1 to {room}_g8, and every attribute of SCHEMA in each profile.
\end{lstlisting}
\end{minipage}\par\medskip

\noindent\begin{minipage}{\linewidth}
\begin{lstlisting}[style=promptP,caption={Prompt for writing the personas of a room type.},label={S-lst:persona}]
You are creating personas for a personalized indoor scene synthesis benchmark.

Scene type: {room}

Available preference profiles (PROFILES). Each profile is one person's complete design for this room:
{profiles}

Generate {n} distinct personas for this scene type.

Each persona should:
1. Be a realistic, detailed character with name, age, occupation, lifestyle, personality traits, living situation, desires and dislikes.
2. Be plausibly compatible with 2-4 of the PROFILES (not just one) and fit one of them best. The persona should hint at some preferences and leave others open.
3. Feel like a real person with a coherent backstory, not an artificial combination of attributes. A persona may strongly imply certain preferences (for example, a minimalist designer implies little decoration) and leave other attributes open.
4. Be meaningfully different from the other personas in background, lifestyle and aesthetic sensibility. Vary age (20 to 75), gender, occupation, family situation and cultural background.
5. Coverage: every profile of kind "coherent" must be the best fit of at least {cap_c} personas, and every profile of kind "mixed" of at least {cap_m}.
6. No brand names. Do not mention benchmarks, schemas, options, attributes or profile ids. Never write any of the FORBIDDEN phrases (case-insensitive, also inside a longer phrase or with hyphens); express the idea in other words.

FORBIDDEN: {forbidden}

Return JSON: {"personas": [{"name": "Full Name", "age": 30, "occupation": "Job title and context", "lifestyle": "2-3 sentences about daily life and habits", "personality_traits": ["trait1", "trait2", "trait3"], "living_situation": "1 sentence", "desires": ["three sentences about the ideal room"], "dislikes": ["three sentences about rooms"]}, ...]} with exactly {n} entries.
\end{lstlisting}
\end{minipage}\par\medskip

\noindent\begin{minipage}{\linewidth}
\begin{lstlisting}[style=promptP,caption={Prompt for writing the desires, the dislikes and the instruction of the \vE{} form.},label={S-lst:explicit}]
You finish the test cases of a research benchmark on personalized indoor scene synthesis. In each test case a person asks an interior designer for a {room}. Each entry of ITEMS gives the person (PERSONA), all of the person's design preferences (PREFERENCES; each gives the topic, the part of the room it is about and the preference), the overall style and mood the person likes (OVERALL), and the topics this person says out loud (STATE).

For each item write:
- "desires": 3 sentences about the room;
- "dislikes": 3 sentences about rooms;
- "query": the request the person says to the designer: first person, natural spoken style, 2-3 sentences.

Rules:
1. Together, the query, the desires and the dislikes must make the preference of each STATE topic clear to an attentive reader, and make clear which part of the room it is about. The query must carry at least two of them; the others may sit in the desires or dislikes.
2. Everything in PREFERENCES and OVERALL is true of this person. The text may hint at the other preferences and at the overall style, the way people do when they talk about their home and their life. The text must not contradict any entry of PREFERENCES.
3. Stay consistent with the PERSONA: the person's background, reasons and way of talking come from it.
4. Use the usual name of a thing when that is the clearest way to say it (for example "a single bed"); design terms are fine when this person would use them.
5. No brand names. Do not mention benchmarks, schemas, options, attributes or topics as categories.

ITEMS: {items}

Return JSON: {"items": [{"item_id": "...", "desires": ["...", "...", "..."], "dislikes": ["...", "...", "..."], "query": "..."}, ...]} with one entry per item, in the same order.
\end{lstlisting}
\end{minipage}\par\medskip

\noindent\begin{minipage}{\linewidth}
\begin{lstlisting}[style=promptP,caption={Prompt for writing the instruction of the \vI{} form.},label={S-lst:implicit}]
You are generating natural user instructions for a personalized indoor scene synthesis benchmark.

For each item in ITEMS, given the persona and scene type, write a short, natural instruction that this person would give when asking someone to design their room. The instruction should:

1. Be 1-2 sentences, casual and natural (as if spoken to a designer)
2. Reference the scene type but NOT explicitly reveal the persona's hidden preferences
3. Optionally hint at the persona's lifestyle or needs, but stay vague enough that the exact preference profile cannot be determined
4. Feel unique to this persona, not generic

Scene type: {room}

ITEMS: {items}

Return JSON: {"items": [{"item_id": "...", "query": "..."}, ...]} with one entry per item, in the same order.
\end{lstlisting}
\end{minipage}\par\medskip

\noindent\begin{minipage}{\linewidth}
\begin{lstlisting}[style=promptP,caption={Prompt for writing the taste narratives.},label={S-lst:narrative}]
You write "taste narratives" for a research benchmark on personalized indoor scene synthesis. A taste narrative is what a person says, in their own voice, about the {room} they would love. A simulated user who later judges rendered rooms reads it, so it must contain all of the person's preferences.

For each item in ITEMS write "narrative":
- first person, in the voice of the given persona, 90-140 words, one paragraph;
- convey EVERY entry of PREFERENCES, each clearly enough that a reader could pick it from a list of alternatives for that topic; each entry gives the part of the room it is about ("about"); make clear which part of the room each preference applies to (several entries are colours or materials of different parts);
- let OVERALL, the overall style and mood the person likes, show in how the person talks about the room;
- describe what the person wants to see (materials, colours, light, objects, arrangement); design terms are fine when this person would use them;
- never write any of the FORBIDDEN phrases (case-insensitive, also inside a longer phrase or with hyphens); do not mention options, lists or benchmarks;
- stay consistent with the persona (age, job, habits).

FORBIDDEN: {forbidden}

ITEMS: {items}

Return JSON: {"items": [{"item_id": "...", "narrative": "..."}, ...]} with one entry per item, in the same order.
\end{lstlisting}
\end{minipage}\par\medskip

\noindent\begin{minipage}{\linewidth}
\begin{lstlisting}[style=promptP,caption={Prompt of the simulated user. From the second turn on, the block HISTORY replaces the placeholder of the same name in the user message, and its placeholder ROUNDS holds one HISTORY\_ROUND line per earlier turn.},label={S-lst:user}]
### SYSTEM
You are role-playing a real person whose home is being furnished by an interior designer. You are the client, not an assistant. Stay in character.

Who you are:
<<PERSONA>>

What you like (private: the designer has not read this, and you never quote it or recite it):
<<TASTE>>

The designer will show you computer renderings of the room they propose. All pictures show the same room from different camera angles. The designer reworks the room after each of your comments.

How to react:
- React the way this person would when looking at the pictures with the designer. Mention specific elements you see: furniture, layout, lighting, materials, colours, mood.
- Judge the room only by your own taste as described above: welcome what it asks for and object to what goes against it. Never ask for something that your taste does not mention or that it rules out.
- Give honest, actionable feedback that the designer can use in the next iteration: say what works, what should change and in which direction, with the most important issue first.
- What you say about the present room must come from what you can see in the pictures. If you cannot tell what something is, what colour it is or what it is made of, leave it out. What you ask for instead comes from your taste.
- Speak in the first person, in your own words and in character. You may use design terms and style names when they say what you mean.
- Ignore rendering artifacts such as a missing ceiling, a plain background outside the room or small glitches; react to the room itself.
- Say it in 3 to 5 sentences.

Then rate how satisfied you are with this room as a whole, as an integer from 1 to 10:
1-2: nothing or almost nothing feels like what you want.
3-4: a few things feel right, but most do not.
5-6: some things feel right and others clearly do not.
7-8: most things feel right, with one or two clear misses.
9-10: it is what you want, or very close to it.
Base the score on how well the room matches your taste, not on the quality of the pictures.

Reply with a JSON object with two fields: "critique" (what you say to the designer) and "score" (your satisfaction from 1 to 10).
### USER
<<HISTORY>>Here <<N_IMAGES_PHRASE>> of the room the designer is proposing now. What do you think of it?
### HISTORY
Your earlier rounds with this designer:
<<ROUNDS>>

Say whether the things you asked to change have improved, and do not repeat praise for things that stayed the same.
### HISTORY_ROUND
Round <<ROUND>>: you gave <<SCORE>> out of 10 and said: "<<CRITIQUE>>"
\end{lstlisting}
\end{minipage}\par\medskip

\noindent\begin{minipage}{\linewidth}
\begin{lstlisting}[style=promptP,caption={Question of the evaluator on one attribute, the mirror type of the bathroom, with the options in the given order.},label={S-lst:judge}]
The 2 images show the same room. Image 1 is a cutaway view from above (the walls and the ceiling nearest the camera are removed so that the inside of the room is visible). Image 2 is a view from inside the room at eye level.
Question: which option best describes the room's mirror type (the mirror above the washbasin)?
A. Oversized Full Wall: One large frameless mirror covers the wall above the basin.
B. Antique Ornate Frame: A mirror in a carved, gilded frame.
C. Tri-Fold Medicine Cabinet: Three side-by-side mirrored doors on a shallow box above the basin.
D. Asymmetrical Cluster: Three mirrors of different sizes hung together off-centre.
Answer with the letter of the best option.
\end{lstlisting}
\end{minipage}\par\medskip

\noindent\begin{minipage}{\linewidth}
\begin{lstlisting}[style=promptP,caption={Prompt of the LLM call that writes the candidate values. \texttt{\{brief\}} holds the room type, the instruction and the persona.},label={S-lst:supp_cand}]
You are an interior designer. The brief (the same text the other systems receive):
{brief}

First write 4 design directions this person might want, each a style and a mood in one sentence. Make the 4 directions clearly different from each other.

Then name the 16 visible design aspects of this room that matter most for this person: things a viewer of the finished room can see. Start with the aspects that shape the look of the whole room, then smaller details. Together the aspects must cover what the room is made of (the finishes of the floor and the walls), its colours, its furniture and how the furniture is arranged, its light and its decoration. The aspects must not overlap.

For every aspect, write 12 candidate choices for this person:
- The first 6 candidates are what this person most likely wants, given the request, the description and the 4 directions, ordered from the most to the least likely.
- The last 6 candidates span the range of what different people commonly choose for this aspect, whatever this person's taste: each comes from a different family of choices (a different colour family, material, shape or arrangement), ordered from the most to the least likely for this person.
- Every candidate must look different in the room from every other candidate of the same aspect; no two candidates may be described with the same words.
- Each candidate has a short label (2-5 words), one sentence that describes only what a viewer sees in the room, and the hex code of its main colour if it names a colour (otherwise null).

Output only JSON: {"directions": [{"style": "...", "mood": "..."}, ...4 items], "aspects": [{"aspect": "...", "scope": "what this aspect decides, in one sentence", "candidates": [{"label": "...", "look": "...", "hex": "#rrggbb or null"}, ...12 items]}, ...16 items]}
\end{lstlisting}
\end{minipage}\par\medskip

\noindent\begin{minipage}{\linewidth}
\begin{lstlisting}[style=promptP,caption={System and user messages of the prompt for the initial plan. \texttt{\{parts\}} lists every attribute as ``- name: description'' with its candidates as ``- \texttt{"label"}: description (hex)'', grouped under ``Candidates from the design aspect \texttt{"aspect"}:''. \texttt{\{form\}} is \texttt{\{"name": VALUE, ...\}} with every attribute name.},label={S-lst:supp_start}]
[system] You are an interior designer. You choose a design plan for one room: exactly one value for each part of the room, either one of the part's listed candidates or a new value that you write. You answer with a JSON object only.

[user] The brief (the same text the other systems receive):
{brief}

A realizer builds the {room} from exactly one value for each of its parts. The parts, each with what it decides and the candidate values written for this person (label: what a viewer sees in the room; main colour). The candidates of one part can come from several design aspects; the aspects only group the candidates, and the part still takes one value:
{parts}

Propose the plan for round 1. The person will see a room built from this plan and will critique it.

Answer with a JSON object with two fields:
- "analysis": at most 80 words on the person's preferences as shown by the brief;
- "plan": an object with one key for each part, in this form: {form}
Each VALUE is one value for the whole part, never an object with one value per design aspect. A value is the label of one of the part's candidates, written exactly; or a new value that you write for the whole part, as {"label": "2-5 words", "look": "one sentence that describes only what a viewer sees in the room", "hex": "#rrggbb of its main colour, or null"}; or null, which keeps the realizer's default for the part.
\end{lstlisting}
\end{minipage}\par\medskip

\noindent\begin{minipage}{\linewidth}
\begin{lstlisting}[style=promptP,caption={Additions to \cref{S-lst:supp_start} for the plan update. The history follows \texttt{\{parts\}}, the task replaces that of the first turn, and ``as shown by the brief'' in the answer format becomes ``as shown by the brief, the critiques and the scores''. A shown value is written as \texttt{"label"} for a candidate, as \texttt{your new value "label" (description)} for a new value, or as \texttt{the realizer's default}.},label={S-lst:supp_update}]
[history] The person has preferences about the room that the brief may not reveal. In each earlier round, the person saw a room built from one plan, wrote a critique of the room and gave a satisfaction score from 1 (not satisfied at all) to 10 (fully satisfied). For each part, a round lists the value shown and what the realizer drew for it. The rounds so far:

Round {j}
Design shown:
  - {part}: {shown value}. What the realizer drew: {scene description}
  ... (one line per part)
Critique: "{feedback of round j}"
Score: {score of round j} out of 10
... (one block per earlier round)

[task, rounds k = 2, ..., K] Infer the person's preferences from the critiques and the scores. Then propose the plan for round {k}. The person will see a room built from this plan and will critique it.

[task, final plan] Infer the person's preferences from the critiques and the scores. Then propose your final plan. The person will not critique it, so choose the plan that you believe matches the person's preferences best.
\end{lstlisting}
\end{minipage}\par\medskip

\noindent\begin{minipage}{\linewidth}
\begin{lstlisting}[style=promptP,caption={Prompt of the LLM call that furnishes the final scene, verbatim for one test interaction of the \vE{} form in a living room. The lists of values, present pieces, furniture tops, judged values, targets and allowed types are generated for each scene.},label={S-lst:supp_furnish},lastline=55]
You are an experienced interior designer. You furnish one living room fully for one person, whose taste the decided values below describe. The room is too empty now: large parts of the floor, of the furniture tops and of the walls are bare. Add furniture, everyday objects and decoration until the room looks lived-in and complete, not too crowded and not empty.

The design of the room has these decided values (attribute: "name: look"):
- furniture_style: "Dark Mahogany: Deep reddish-brown polished hardwood."
- wall_color: "Warm Cream: A soft, buttery off-white that reflects light gently."
- feature_wall: "Damask Pattern: Intricate symmetrical floral motifs in gold and deep red."
- flooring: "Beige Wool Pile: A thick, soft natural wool in a sandy tone."
- upholstery: "Deep Blue Chenille: A soft, textured fabric with a matte finish."
- seating_config: "English Roll-Arm: Classic curved arms with a relaxed, plump back."
- layout_type: "Conversational Circle: All pieces angled inward toward the coffee table."
- lighting_scheme: "Brass Floor Lamp: A tall, curved lamp with a fabric shade."
- window_treatment: "Heavy Velvet Curtains: Thick, floor-length drapes in a deep gold."
- decor_theme: "Densely Packed Archive: Floor-to-ceiling mahogany shelves overflowing with leather-bound books, multiple antique clocks, and various vintage globes with no empty space."

The room measures 6.2 x 5.0 m.
The room already holds these pieces:
- on the floor: 1 coffee table, 2 sofa, 1 side table, 2 shelf, 1 floor lamp
- on furniture tops: 2 books, 2 globe
- on the walls: nothing
- at the ceiling: nothing

Furniture tops that can carry objects (no other piece of the room has such a top):
- coffee_table#1: top 0.85 x 0.85 m at height 0.42 m, holds 1 globe; target: 3 more objects (at most 5)
- side_table#1: top 0.87 x 0.72 m at height 0.72 m, holds 1 globe; target: 3 more objects (at most 5)
- shelf#1: top 1.01 x 0.38 m at height 1.54 m, holds 1 books; target: 3 more objects (at most 3); too high for devices

Small objects are of two kinds:
- everyday objects, which a person uses: books, cookware, tableware, toy; and the devices: computer, printer, speaker, tv.
- ornaments, which are placed for looks: candle, flowers, globe, plant, sculpture, tabletop_decor, vase; and every wall piece.

First judge each of these values: decor_theme. Give each one of three states:
- "empty": the value asks for empty, bare, clear or uncluttered surfaces or walls, or for no ornaments at all. A value that asks for empty surfaces and also states a number is "empty".
- "complete": the value names concrete pieces: one piece in the singular ("one sketch", "a single vase", "a giant screen"), a list of two or more kinds of pieces ("a trunk, a clock and a tea service", "pothos, journals and a bonsai"), or a number of pieces ("three prints and two vases", "ten frames").
- "open": the value names no concrete piece (it describes a theme, a style, a mood or a pattern), or it names only one kind of piece in the plural without a number ("prints", "plants").
What a state means for your additions:
- "empty" for decor_theme: add nothing on any furniture top, no wall piece and no ornament. Furniture and other floor pieces that a person uses are still added.
- "complete" and "open" forbid nothing else: the pieces of the value stay as they are, and pieces of other kinds are added beside them.

Targets. Reach each target; stay below it only where a state "empty" forbids the pieces or no allowed type fits the function of the room:
A. Floor: about 10 floor pieces in the room after your additions (it holds 7 now). Add the furniture and the other floor pieces that this person's living room still lacks, at most 10 added pieces in total.
B. Tops: each top of the list above gets the number of objects named as its target. The top of a piece that you add gets 2 to 5 objects by its size.
C. Walls: at least 6 wall pieces in the room after your additions (it holds 0 now); add at most 6.
D. Variety: at least 8 different types among all pieces that you add.

Rules:
1. Every added piece must fit the function of the room and the style of the values.
2. Never add a piece that a value excludes.
3. Do not add another piece of a kind whose exact number a value states (for example a value with two armchairs allows no third armchair).
4. A device stands only on the top that you name for it, and only on a top lower than 1.2 m. A tv stands only on a tv_stand, sideboard or console_table (one of the list of tops, or one that you add in "room_pieces"), never on a coffee table, a dining table or a desk.
5. A sculpture stands on a top or on a pedestal that you add, never on the floor.
6. Give a quantity for every piece.
7. Write each type exactly as it is written in the lists of allowed types (for example "books", also for one book).

Objects on tops. Each top gets objects that belong to its use. A table, island or bar where people eat gets place settings: tableware in pairs by the number of seats (up to 6 pieces) and at most one centrepiece (a vase, flowers or a candle); no device, appliance, book stack or further ornament on it. A desk gets work objects. A coffee table gets a few books and one ornament. Shelves, sideboards and cabinets take books and ornaments. The target of 8 different types counts over the whole room, not per top.
\end{lstlisting}
\end{minipage}\par\medskip

\noindent\begin{minipage}{\linewidth}
\begin{lstlisting}[style=promptP,title={\lstlistingname~\ref*{S-lst:supp_furnish} (continued)},firstline=56]
Value consistency. The design is complete for the things its values decide. Each decided value is the full statement about one kind of thing in the room; the attribute's name and the value's text say which kind (for example the seats, the tables, the storage furniture, the fixtures, the decoration, the wall art, or how the furniture is arranged). For each attribute, first write in "governs" the kind of thing its value decides. Then go through every candidate type (the allowed types of "room_pieces", "on_tops" and "wall_pieces" named below) and put into "types" every type that is of that kind, except the types that the value's own text names or describes: those express the value, stay off the list, and more of them are welcome. A value that asks for none, for empty tops, for a single piece or for very few puts every type of its kind on the list. An attribute about how the furniture is arranged or grouped lists the seat types and the table types: one more seat or table would join or break the arrangement. An attribute about a surface, a colour, a material, a fabric, a window covering or the light decides no candidate type: its list is empty. Propose no piece of a type on a list, on the floor, on a top or on a wall. A target may stay unmet when the lists leave no fitting type.

Answer with one JSON object with these keys:
"avoid": one key for each decided value above (the attribute names, in that order); each key holds {"governs", "types"}: "governs" names, in at most 60 characters, the kind of thing that the value decides; "types" is the list of the candidate types to avoid for that attribute, each written exactly as in the lists of allowed types; an empty list when there is none.
"value_states": one entry for each value that you judge (an empty list when there is none): {"attribute", "state", "reason"}; "reason" has at most 5 words.
"strategy": one or two sentences: what the room still lacks for this person and how your additions reach the targets.
"room_pieces": pieces that stand on the floor, each {"type", "count"}; the counts sum to at most 10. Allowed types with a top that can carry objects: bookshelf, cabinet, chest_trunk, console_table, display_case, dresser, file_cabinet, nightstand, pedestal, piano, shelf, shoe_rack, side_table, sideboard, toy_box, tv_stand, washing_machine (a bookshelf or cabinet or display_case or shelf only when it is lower than 1.6 m). Allowed types without such a top: armchair, basket, bathroom_sink, bathtub, bean_bag, bench, chair, coat_rack, easel, fan, fridge, grandfather_clock, high_chair, kids_chair, kitchen_cabinet, kitchen_island, ladder, laundry_hamper, magazine_rack, office_chair, ottoman, radiator, range_hood, rocking_chair, rocking_horse, room_divider, sofa, stool, stove, stroller, toilet, towel_rack, trash_can, vanity, wardrobe, water_dispenser.
"on_tops": one entry for each furniture top that gets objects: {"top", "objects"}. "top" is a label of the list of tops above, or "<type>#<n>" for the top of a room piece with a top that you add; <n> continues the numbers of that type in the list (the first added piece of a type that the list does not hold is "<type>#1"). "objects" is a list of {"type", "text", "count"}; "text" describes the object for retrieval in at most 8 words, in the style of the values. Allowed types: the everyday objects, the devices and the ornaments named above.
"wall_pieces": pictures or other wall pieces to add, each {"type", "text", "count"}; the counts sum to at most 6. Allowed types: painting, board, wall_clock.
An empty list means that nothing is added.
\end{lstlisting}
\end{minipage}\par\medskip

\noindent\begin{minipage}{\linewidth}
\vspace{2pt}{\captionsetup{position=top, skip=0.6pt, hypcap=false}\captionof{lstlisting}{Specifications of three values of the turn-1 plan in the interaction of \cref{fig:overview} of the main paper. The retrieval queries, asset identifiers, empty fields and probabilities of the placement answers are omitted.}\label{S-lst:ex_spec}}
\begin{tcolorbox}[colback=black!3, boxsep=2pt, left=0.6pt, right=0.6pt, top=0pt, bottom=0pt, before skip=0pt, after skip=0pt]
\begin{minipage}[t]{0.325\linewidth}\vspace{0pt}
\begin{lstlisting}[style=jsoncol,lastline=16]
{
 "flooring_type": {
  "attribute": "flooring_type",
  "category": "floor",
  "label": "Black Marquina Marble",
  "look": "Highly polished black stone with stark white veining.",
  "colors": {
   "floor": "#000000"
  },
  "surfaces": {
   "floor": {
    "acg": "acg:Marble006",
    "cover": "material"
   }
  }
 },
\end{lstlisting}\end{minipage}\hfill
\begin{minipage}[t]{0.325\linewidth}\vspace{0pt}
\begin{lstlisting}[style=jsoncol,firstline=17,lastline=37]
 "lighting_scheme": {
  "attribute": "lighting_scheme",
  "category": "light",
  "label": "Crystal Sputnik Chandelier",
  "look": "An explosive array of gold arms and sparkling crystals.",
  "light": {
   "cct": 3000,
   "power": "high",
   "lamps": false,
   "daylight": "low",
   "time": "day",
   "source": "electric",
   "fixtures": [
    {
     "type": "ceiling_lamp",
     "count": 1,
     "surface": "ceiling"
    }
   ]
  }
 },
\end{lstlisting}\end{minipage}\hfill
\begin{minipage}[t]{0.325\linewidth}\vspace{0pt}
\begin{lstlisting}[style=jsoncol,firstline=38]
 "seating_arrangement": {
  "attribute": "seating_arrangement",
  "category": "seating",
  "label": "Circular Halo",
  "look": "Chairs spaced equally around the perimeter of a round table.",
  "objects": [
   {
    "type": "chair",
    "count": 2,
    "surface": "floor",
    "size_cm": [46.9, 52.1, 92.0]
   },
   {
    "type": "dining_table",
    "count": 1,
    "surface": "floor",
    "size_cm": [156.2, 100.5, 75.6]
   }
  ],
  "placement": {
   "anchor": "center",
   "along": 0.5,
   "grouping": "middle",
   "k": 1,
   "symmetry": true,
   "facing": "focal",
   "spacing": "normal"
  }
 }
}
\end{lstlisting}\end{minipage}
\end{tcolorbox}
\end{minipage}\par\medskip

\noindent\begin{minipage}{\linewidth}
\begin{lstlisting}[style=promptP,caption={Scene descriptions $D_1$ of three attributes in the interaction of \cref{fig:overview} of the main paper. The description of the seating arrangement is cut after the placement, before the probabilities of the placement answers.},label={S-lst:ex_descr}]
flooring_type: Floor: Marble 006 (black, dark, gray, grey, marble, stone; black). Floor black (#000000).

lighting_scheme: Light 3000 K (warm); shown in the daytime; fixtures: 1 ceiling lamp; lamps off; daylight through the window on.

seating_arrangement: 2 chairs: built with one-piece moulded shell, fully padded seat, thin metal tube legs, open arms, carved curves, tufting, decorative mouldings, fully rounded, curvy shapes; frame and legs Metal 007 (bumpy, gold, metal, scratches, silver, steel; brown); fronts Metal 007 (bumpy, gold, metal, scratches, silver, steel; brown); upholstery Fabric 026 (carpet, cloth, fabric, floor, red, tiled; brown); about 50 x 53 x 100 cm, free-standing in the room, along 0.5, grouped middle, k 1, arranged symmetrically, facing the focal, normal spacing. [...]
\end{lstlisting}
\end{minipage}\par\medskip

\noindent\begin{minipage}{\linewidth}
\begin{lstlisting}[style=jsonP,caption={Excerpt of the reply of the LLM call that furnishes the final scene of the second user of \cref{fig:teaser} of the main paper (\vE{} form, seed 3), with the values of three attributes, the object types that these values rule out and the proposed pieces that are dropped. The lists of ruled-out types of the other seven attributes are empty.},label={S-lst:ex_avoid}]
{
 "values": {
  "seating_arrangement": "Symmetrical Square",
  "storage_solution": "Black Lacquer Vitrine",
  "lighting_scheme": "Art Deco Globe"
 },
 "avoid": {
  "seating_arrangement": {
   "governs": "seating layout",
   "types": ["armchair", "bench", "bean_bag", "high_chair", "kids_chair", "office_chair", "ottoman", "rocking_chair", "stool"]
  },
  "storage_solution": {
   "governs": "storage furniture",
   "types": ["bookshelf", "chest_trunk", "dresser", "file_cabinet", "nightstand", "shelf", "shoe_rack", "sideboard", "toy_box", "tv_stand", "wardrobe"]
  },
  "lighting_scheme": {
   "governs": "ceiling fixtures",
   "types": []
  }
 },
 "dropped": [
  {
   "part": "room_pieces",
   "type": "sideboard",
   "count": 1,
   "by": "storage_solution"
  },
  {
   "part": "on_tops",
   "type": "speaker",
   "count": 1,
   "by": "storage_solution",
   "top": "sideboard#1"
  },
  {
   "part": "on_tops",
   "type": "tabletop_decor",
   "count": 1,
   "by": "storage_solution",
   "top": "sideboard#1"
  },
  {
   "part": "on_tops",
   "type": "candle",
   "count": 2,
   "by": "storage_solution",
   "top": "sideboard#1"
  }
 ]
}
\end{lstlisting}
\end{minipage}\par\medskip

\clearpage\flushbottom

\clearpage\onecolumn\makeatletter\if@twocolumn\onecolumn\fi\makeatother
\section{Full schemas}
\label{S-sec:supp_full}
The tables of this section give the schema of every room type (\cref{S-sec:supp_schemas}).
Each row gives one attribute: its name, its group, its description in the attribute list and the names of its values, separated by semicolons.
The definitions of the attributes and the descriptions of the values are not listed; \cref{S-lst:judge} shows both for one attribute.

\footnotesize
\begin{longtable}{@{}>{\raggedright\arraybackslash}p{2.5cm}>{\raggedright\arraybackslash}p{1.7cm}>{\raggedright\arraybackslash}p{\dimexpr0.5\textwidth-2.1cm-3\tabcolsep\relax}>{\raggedright\arraybackslash}p{\dimexpr0.5\textwidth-2.1cm-3\tabcolsep\relax}@{}}
\caption{Schema of the room type bathroom: 10 attributes, 50 values.}\label{S-tab:schema_bathroom}\\
\toprule
Attribute & Group & Description in the attribute list & Values \\
\midrule\endfirsthead
\toprule
Attribute & Group & Description in the attribute list & Values \\
\midrule\endhead
\bottomrule\endfoot
\textbf{wall tile} & surfaces & This attribute determines the lower section of the walls and the shower walls. & White Subway Tile; Large Beige Porcelain; White Marble; Patterned Cement Tile; Smooth Concrete \\
\textbf{wall color} & surfaces & The attribute determines the upper walls above the tiles. & Warm White; Dusty Blue; Terracotta; Deep Green; Charcoal; Blush Pink \\
\textbf{flooring type} & surfaces & The flooring type determines the floor surface of the bathroom. & Large Pale Tiles; Black-and-White Checkerboard; Small Hexagon Mosaic; Wood-look Planks; Polished Concrete; Terrazzo \\
\textbf{vanity layout} & furniture & The vanity layout determines the arrangement of the washbasin unit. & Floating Double Vanity; Washstand on Legs; Wall-Hung Console; Built-In Vanity Cabinet; Pedestal Sink \\
\textbf{shower configuration} & furniture & The shower configuration decides the shower or bath and its enclosure. & Frameless Glass Walk-in; Black-Framed Glass Enclosure; Bathtub with Curtain; Open Shower with Bench \\
\textbf{storage solution} & furniture & This attribute determines the storage features located elsewhere in the room. & Tall Closed Cabinet; Open Towel Shelves; Wall Niches; Over-Toilet Cabinet; No Extra Storage \\
\textbf{mirror type} & accessories & The mirror type decides the mirror above the washbasin. & Oversized Full Wall; Antique Ornate Frame; Tri-Fold Medicine Cabinet; Asymmetrical Cluster \\
\textbf{lighting scheme} & lighting & The lighting scheme decides the illumination of the room. & Bright Daylight; Cool Bright Overhead; Mirror Sconces; Bare Bulb Pendants; Hidden Strip Glow \\
\textbf{window treatment} & textiles & Window treatment decides the window. & Frosted Glass; Stained Glass; Bamboo Roman Shade; Venetian Blinds; Wooden Shutters \\
\textbf{decor theme} & accessories & The decor theme decides the decorative objects on the walls, counter, and floor, as well as the bath mat. & Bare Surfaces; Potted Plants; Framed Prints; Woven Baskets; Ceramic Vessels \\
\end{longtable}

\begin{longtable}{@{}>{\raggedright\arraybackslash}p{2.5cm}>{\raggedright\arraybackslash}p{1.7cm}>{\raggedright\arraybackslash}p{\dimexpr0.5\textwidth-2.1cm-3\tabcolsep\relax}>{\raggedright\arraybackslash}p{\dimexpr0.5\textwidth-2.1cm-3\tabcolsep\relax}@{}}
\caption{Schema of the room type bedroom: 10 attributes, 47 values.}\label{S-tab:schema_bedroom}\\
\toprule
Attribute & Group & Description in the attribute list & Values \\
\midrule\endfirsthead
\toprule
Attribute & Group & Description in the attribute list & Values \\
\midrule\endhead
\bottomrule\endfoot
\textbf{wall color} & surfaces & The wall color determines the finish of the walls, excluding any feature wall with its own finish. & Warm White; Dusty Blue; Terracotta; Deep Green; Charcoal; Blush Pink \\
\textbf{feature wall} & surfaces & The feature wall determines the finish of the wall opposite the door. & Plain Paint; Patterned Wallpaper; Wood Panelling; Exposed Brick; Raw Concrete \\
\textbf{flooring type} & surfaces & The flooring type decides the floor surface of the room. & Hardwood; Plush Carpet; Polished Concrete; Herringbone Parquet \\
\textbf{bed frame material} & furniture & The bed frame and headboard. & Solid Wood; Upholstered Fabric; Metal Frame \\
\textbf{bedding} & textiles & Bedding determines the appearance of the bed. & Crisp White; Sage Green; Slate Grey; Navy Blue; Mustard Yellow; Patterned Quilt \\
\textbf{storage solution} & furniture & The storage solution determines the arrangement of furniture for storing clothes and objects. & Built-in Wardrobe Wall; Freestanding Dresser; Open Bookshelves; Open Clothes Rail; Under-Bed Drawers \\
\textbf{furniture layout} & spatial arrangement & The furniture layout determines the placement of the bed and other furniture within the room. & Centred Symmetrical; Corner Bed; Sideways by the Window; Raised Platform \\
\textbf{lighting scheme} & lighting & The lighting scheme decides the illumination of the room. & Bright Daylight; Warm Lamp Glow; Bedside Reading Lights; Statement Chandelier; Hidden Cove Glow \\
\textbf{window treatment} & textiles & Window treatment decides the window. & Sheer Curtains; Linen Side Panels; Venetian Blinds; Bamboo Roman Shades; Wooden Shutters \\
\textbf{decor theme} & accessories & The decor theme decides the decorative objects on the walls, tables, and floor. & Bare Surfaces; Statement Art; Gallery Wall; Indoor Jungle \\
\end{longtable}

\begin{longtable}{@{}>{\raggedright\arraybackslash}p{2.5cm}>{\raggedright\arraybackslash}p{1.7cm}>{\raggedright\arraybackslash}p{\dimexpr0.5\textwidth-2.1cm-3\tabcolsep\relax}>{\raggedright\arraybackslash}p{\dimexpr0.5\textwidth-2.1cm-3\tabcolsep\relax}@{}}
\caption{Schema of the room type children's room: 10 attributes, 44 values.}\label{S-tab:schema_children_room}\\
\toprule
Attribute & Group & Description in the attribute list & Values \\
\midrule\endfirsthead
\toprule
Attribute & Group & Description in the attribute list & Values \\
\midrule\endhead
\bottomrule\endfoot
\textbf{feature wall} & surfaces & The feature wall attribute determines the finish of the long wall on the left as you enter. & Plain Paint; Polka Dots; Geometric Shapes; Landscape Mural; Floral Wallpaper \\
\textbf{wall color} & surfaces & The wall color determines the appearance of the walls. & Warm White; Blush Pink; Sage Green; Sunny Yellow; Sky Blue \\
\textbf{flooring type} & surfaces & The attribute determines the floor surface of the room. & Hardwood with Rug; Wall-to-Wall Carpet; Cork Floor; Foam Play Mats \\
\textbf{bed type} & furniture & The bed type determines the shape and structure of the bed. & Single Bed; Bunk Bed; House-Frame Bed; Canopy Bed; Low Floor Bed \\
\textbf{furniture finish} & furniture & The furniture finish determines the appearance of the bed frame, storage furniture, and activity furniture. & Natural Light Wood; White Painted; Bright Primary Colours; Rattan \& Cane \\
\textbf{storage solution} & furniture & It determines the storage area. & Wall Shelves; Closed Cupboards; Cubby Grid; Toy Chest \\
\textbf{activity zone} & spatial arrangement & The activity zone determines the activity corner of the room. & Art \& Craft Corner; Reading Nook; Music Corner; Study Desk Station \\
\textbf{lighting scheme} & lighting & The lighting scheme decides the illumination of the room. & Bright Daylight; Warm Lamp Glow; Fairy Lights; Playful Pendant \\
\textbf{window treatment} & textiles & Window treatment decides the window. & Sheer Curtains; Patterned Curtains; Blackout Roller Blind; Wooden Shutters; Bamboo Roman Shade \\
\textbf{decor theme} & accessories & The decor theme decides the walls, bed, shelves, and floor. & Bare Surfaces; Framed Prints; Plush Toys; Learning Props \\
\end{longtable}

\begin{longtable}{@{}>{\raggedright\arraybackslash}p{2.5cm}>{\raggedright\arraybackslash}p{1.7cm}>{\raggedright\arraybackslash}p{\dimexpr0.5\textwidth-2.1cm-3\tabcolsep\relax}>{\raggedright\arraybackslash}p{\dimexpr0.5\textwidth-2.1cm-3\tabcolsep\relax}@{}}
\caption{Schema of the room type dining room: 10 attributes, 48 values.}\label{S-tab:schema_dining_room}\\
\toprule
Attribute & Group & Description in the attribute list & Values \\
\midrule\endfirsthead
\toprule
Attribute & Group & Description in the attribute list & Values \\
\midrule\endhead
\bottomrule\endfoot
\textbf{wall color} & surfaces & The wall color determines the finish of the walls, excluding any feature wall with its own finish. & Warm White; Dusty Blue; Terracotta; Deep Green; Charcoal; Blush Pink \\
\textbf{feature wall} & surfaces & The feature wall attribute determines the finish of the wall opposite the door. & Plain Paint; Patterned Wallpaper; Wood Panelling; Exposed Brick; Raw Concrete \\
\textbf{flooring type} & surfaces & The flooring type decides the floor surface of the room. & Wide Plank Hardwood; Herringbone Parquet; Marble Checkerboard; Terracotta Tiles; Polished Concrete \\
\textbf{lighting scheme} & lighting & The lighting scheme decides the illumination of the room. & Bright Daylight; Statement Chandelier; Pendant Row; Wall Sconces; Warm Lamp Glow \\
\textbf{furniture style} & furniture & The furniture style determines the design of the table, chairs, bench, built-in seat bases, and storage furniture. & Scandinavian; Industrial; Mid-Century Modern; Traditional; Art Deco; Contemporary \\
\textbf{table shape} & furniture & The table shape determines the dining table. & Long Rectangular Table; Oval Table; Round Table; Square Table \\
\textbf{seating arrangement} & spatial arrangement & The seating arrangement determines the placement of the table and its seats within the room. & Chairs All Around; Bench and Chairs; Corner Banquette; Window Banquette \\
\textbf{storage solution} & furniture & The storage solution determines the location of the storage piece within the room. & Low Sideboard; Glass-Front Cabinet; Open Plate Shelves; Wine Rack Wall \\
\textbf{window treatment} & textiles & Window treatment decides the window area. & Sheer Curtains; Linen Side Panels; Heavy Velvet Drapes; Venetian Blinds; Wooden Shutters \\
\textbf{table setting} & accessories & Table setting determines the arrangement of items on the dining table. & Bare Table; White Tablecloth; Full Place Settings; Flower Centrepiece \\
\end{longtable}

\begin{longtable}{@{}>{\raggedright\arraybackslash}p{2.5cm}>{\raggedright\arraybackslash}p{1.7cm}>{\raggedright\arraybackslash}p{\dimexpr0.5\textwidth-2.1cm-3\tabcolsep\relax}>{\raggedright\arraybackslash}p{\dimexpr0.5\textwidth-2.1cm-3\tabcolsep\relax}@{}}
\caption{Schema of the room type gallery: 10 attributes, 40 values.}\label{S-tab:schema_gallery}\\
\toprule
Attribute & Group & Description in the attribute list & Values \\
\midrule\endfirsthead
\toprule
Attribute & Group & Description in the attribute list & Values \\
\midrule\endhead
\bottomrule\endfoot
\textbf{hanging style} & spatial arrangement & Hanging style determines the arrangement of artworks on the walls. & Sparse Single Row; Salon Hang; Symmetrical Grid; Hung from Rail \\
\textbf{art type} & accessories & The art type attribute decides the artworks and their framing. & Abstract Paintings; Black-and-White Photographs; Classical Paintings; Textile Hangings \\
\textbf{wall color} & surfaces & The wall color determines the finish of the walls, excluding any feature wall with its own finish. & Gallery White; Terracotta; Deep Red; Navy Blue; Charcoal; Blush Pink \\
\textbf{feature wall} & surfaces & The feature wall determines the finish of the wall opposite the entrance. & Plain Paint; Exposed Brick; Wood Panelling; Raw Concrete \\
\textbf{flooring type} & surfaces & The flooring type decides the floor surface of the room. & Polished Concrete; Herringbone Parquet; Large Format Tile; Terrazzo \\
\textbf{display fixtures} & furniture & Display fixtures decide the floor-level stands and cases. & White Plinths; Glass Vitrines; Freestanding Partitions \\
\textbf{seating integration} & furniture & The seating integration attribute decides the seating area of the room. & No Seating; Simple Benches; Lounge Sofas \\
\textbf{furniture arrangement} & spatial arrangement & Furniture arrangement decides where the display fixtures and seats stand on the floor. & Along the Walls; One Central Group; Two Separate Groups \\
\textbf{lighting scheme} & lighting & The lighting scheme decides the illumination of the room. & Bright Daylight; Warm Lamp Glow; Picture Lights in the Dark; Cool Bright Overhead \\
\textbf{decor theme} & accessories & The decor theme determines the decorative objects other than wall artworks and display fixtures. & No Decor; Floor Plants; Floor Sculptures; Clock, Trunk \& Globe; Video Screens \\
\end{longtable}

\begin{longtable}{@{}>{\raggedright\arraybackslash}p{2.5cm}>{\raggedright\arraybackslash}p{1.7cm}>{\raggedright\arraybackslash}p{\dimexpr0.5\textwidth-2.1cm-3\tabcolsep\relax}>{\raggedright\arraybackslash}p{\dimexpr0.5\textwidth-2.1cm-3\tabcolsep\relax}@{}}
\caption{Schema of the room type kitchen: 10 attributes, 47 values.}\label{S-tab:schema_kitchen}\\
\toprule
Attribute & Group & Description in the attribute list & Values \\
\midrule\endfirsthead
\toprule
Attribute & Group & Description in the attribute list & Values \\
\midrule\endhead
\bottomrule\endfoot
\textbf{cabinet color} & furniture & This attribute determines the appearance of the cabinet doors and drawer fronts. & White; Light Oak; Sage Green; Navy Blue; Matte Black \\
\textbf{upper cabinets} & furniture & Upper cabinets determine the wall space above the counters. & Closed Wall Cabinets; Glass-Front Cabinets; Open Shelves; No Upper Storage \\
\textbf{countertop material} & furniture & The attribute determines the material of the countertops and island top. & Engineered Quartz; Butcher Block Wood; Polished Concrete; Natural Marble; Classic Granite; Stainless Steel \\
\textbf{backsplash type} & surfaces & It determines the wall surface between the counters and the wall cabinets. & Full Slab Extension; Exposed Brick; White Subway Tile; Hexagon Mosaic; Patterned Cement Tile \\
\textbf{wall color} & surfaces & The wall color determines the appearance of the wall surfaces not covered by cabinets or the backsplash. & Warm White; Dusty Blue; Terracotta; Deep Green; Charcoal; Blush Pink \\
\textbf{flooring type} & surfaces & The attribute decides the floor surface of the kitchen. & Wide Plank Wood; Black-and-White Checkerboard; Terracotta Tiles; Polished Concrete; Large Pale Tiles \\
\textbf{spatial layout} & spatial arrangement & The spatial layout determines the placement of the counter within the room. & Island Centered; L-Shape Corner; Straight Single-Wall \\
\textbf{lighting scheme} & lighting & The lighting scheme decides the illumination of the room. & Bright Daylight; Cool Bright Overhead; Pendant Row; Under-Cabinet Glow; Wall Sconces \\
\textbf{dining seating} & furniture & The dining seating determines the seating area for eating in the kitchen. & Bar Stools at Counter; Small Dining Table; Built-in Banquette; No Seating \\
\textbf{decor theme} & accessories & The decor theme decides the decorative and everyday objects on the counters, the window sill, and the walls. & Bare Counters; Cookware on Display; Potted Herbs; Art \& Ceramics \\
\end{longtable}

\begin{longtable}{@{}>{\raggedright\arraybackslash}p{2.5cm}>{\raggedright\arraybackslash}p{1.7cm}>{\raggedright\arraybackslash}p{\dimexpr0.5\textwidth-2.1cm-3\tabcolsep\relax}>{\raggedright\arraybackslash}p{\dimexpr0.5\textwidth-2.1cm-3\tabcolsep\relax}@{}}
\caption{Schema of the room type library: 10 attributes, 50 values.}\label{S-tab:schema_library}\\
\toprule
Attribute & Group & Description in the attribute list & Values \\
\midrule\endfirsthead
\toprule
Attribute & Group & Description in the attribute list & Values \\
\midrule\endhead
\bottomrule\endfoot
\textbf{furniture style} & furniture & The furniture style determines the design of the bookcases, tables, desk and chair frames. & Scandinavian; Industrial; Mid-Century Modern; Traditional; Art Deco; Contemporary \\
\textbf{wall color} & surfaces & The wall color determines the finish of the walls, excluding any feature wall with its own finish. & Warm White; Dusty Blue; Terracotta; Deep Green; Charcoal; Blush Pink \\
\textbf{feature wall} & surfaces & The feature wall determines the finish of the back wall opposite the door. & Plain Paint; Patterned Wallpaper; Exposed Brick; Wood Panelling; Fabric Panels \\
\textbf{flooring type} & surfaces & The attribute determines the floor surface of the room. & Polished Hardwood; Large Patterned Rug; Polished Concrete; Marble Checkerboard; Cork Tiles \\
\textbf{upholstery} & textiles & Upholstery determines the covering of the armchairs, sofas, and cushions. & Cognac Leather; Emerald Velvet; Oatmeal Linen; Slate Grey; Mustard Yellow; Navy Blue \\
\textbf{shelving configuration} & furniture & The shelving configuration determines the placement of bookshelves within the room. & Floor-to-Ceiling Built-ins; Freestanding Bookcases; Floating Wall Shelves; Central Book Stacks \\
\textbf{seating arrangement} & spatial arrangement & The seating arrangement determines the placement and orientation of the seats. & Reading Circle; Window Seats; Lounge Groups; Reading Corner \\
\textbf{workspace integration} & furniture & Workspace integration determines the placement of desks and work tables. & Writing Desk; Long Shared Table; Study Carrels; No Desk \\
\textbf{lighting scheme} & lighting & The lighting scheme decides the illumination of the room. & Bright Daylight; Warm Lamp Glow; Green Banker's Lamps; Lit Bookshelves; Hidden Cove Glow \\
\textbf{decor theme} & accessories & The decor theme decides the decorative objects on the shelves, tables, walls, and floor, excluding books. & Books Only; Indoor Plants; Clock, Trunk \& Globe; Abstract Paintings; Classical Busts \\
\end{longtable}

\begin{longtable}{@{}>{\raggedright\arraybackslash}p{2.5cm}>{\raggedright\arraybackslash}p{1.7cm}>{\raggedright\arraybackslash}p{\dimexpr0.5\textwidth-2.1cm-3\tabcolsep\relax}>{\raggedright\arraybackslash}p{\dimexpr0.5\textwidth-2.1cm-3\tabcolsep\relax}@{}}
\caption{Schema of the room type living room: 10 attributes, 55 values.}\label{S-tab:schema_living_room}\\
\toprule
Attribute & Group & Description in the attribute list & Values \\
\midrule\endfirsthead
\toprule
Attribute & Group & Description in the attribute list & Values \\
\midrule\endhead
\bottomrule\endfoot
\textbf{furniture style} & furniture & Furniture style determines the design of the seating, tables, shelving, sideboard, and TV stand. & Scandinavian; Industrial; Mid-Century Modern; Traditional; Rattan \& Cane; Contemporary \\
\textbf{wall color} & surfaces & The wall color determines the appearance of the walls. & Warm White; Dusty Blue; Terracotta; Deep Green; Charcoal; Blush Pink \\
\textbf{feature wall} & surfaces & The feature wall determines the finish of the wall opposite the room's entrance. & Plain Paint; Exposed Brick; Wood Slats; Patterned Wallpaper; Raw Concrete \\
\textbf{flooring} & surfaces & Flooring determines the floor surface of the room. & Wide Plank Hardwood; Herringbone Parquet; Polished Concrete; Large Format Tile; Wall-to-Wall Carpet; Large Patterned Rug \\
\textbf{upholstery} & textiles & Upholstery determines the covering of the seating furniture. & Cognac Leather; Emerald Velvet; Oatmeal Linen; Slate Grey; Mustard Yellow; Navy Blue \\
\textbf{seating config} & furniture & The seating config attribute decides the seat pieces in the room. & Large U-Sectional; Sofa \& Accent Chairs; Classic 3-Piece Set; Low Floor Cushions; Paired Loveseats; Bench \& Ottoman Cluster \\
\textbf{layout type} & spatial arrangement & The layout type determines the placement and orientation of the seats. & Conversation Circle; TV-Facing; Wall-Lined; Lounge + Dining \\
\textbf{lighting scheme} & lighting & The lighting scheme decides the illumination of the room. & Bright Daylight; Warm Lamp Glow; Large Pendant; Cool Bright Overhead; Hidden Cove Glow \\
\textbf{window treatment} & textiles & Window treatment decides the window area. & Sheer Curtains; Linen Side Panels; Heavy Velvet Drapes; Venetian Blinds; Bamboo Roman Shades; Wooden Shutters \\
\textbf{decor theme} & accessories & The decor theme decides the decorative objects on the walls, shelves, tables, and floor. & Bare Surfaces; Statement Pieces; Gallery Wall; Indoor Jungle; Collector Shelves \\
\end{longtable}

\begin{longtable}{@{}>{\raggedright\arraybackslash}p{2.5cm}>{\raggedright\arraybackslash}p{1.7cm}>{\raggedright\arraybackslash}p{\dimexpr0.5\textwidth-2.1cm-3\tabcolsep\relax}>{\raggedright\arraybackslash}p{\dimexpr0.5\textwidth-2.1cm-3\tabcolsep\relax}@{}}
\caption{Schema of the room type office: 10 attributes, 49 values.}\label{S-tab:schema_office}\\
\toprule
Attribute & Group & Description in the attribute list & Values \\
\midrule\endfirsthead
\toprule
Attribute & Group & Description in the attribute list & Values \\
\midrule\endhead
\bottomrule\endfoot
\textbf{wall color} & surfaces & The wall color determines the finish of the walls, excluding any feature wall with its own finish. & Warm White; Dusty Blue; Terracotta; Deep Green; Charcoal; Blush Pink \\
\textbf{feature wall} & surfaces & The feature wall determines the finish of the back wall opposite the door. & Plain Paint; Exposed Brick; Wood Slats; Fabric Panels; Raw Concrete \\
\textbf{flooring type} & surfaces & The attribute determines the floor of the room. & Wide Plank Hardwood; Herringbone Parquet; Carpet Tiles; Polished Concrete; Cork Tiles \\
\textbf{lighting scheme} & lighting & The lighting scheme decides the illumination of the room. & Bright Daylight; Cool Bright Overhead; Warm Lamp Glow; Green Banker's Lamps; Hidden Cove Glow \\
\textbf{furniture style} & furniture & The furniture style determines the design of the desk, chair and seat frames, tables, and storage furniture. & Scandinavian; Industrial; Mid-Century Modern; Traditional; Art Deco; Contemporary \\
\textbf{desk type} & furniture & The desk type determines the work desk and its accompanying seating. & Executive Desk; L-Shaped Desk; Twin Facing Desks; Standing Desk \\
\textbf{storage solution} & furniture & The storage solution determines the placement of storage furniture within the room. & Filing Cabinets; Floor-to-Ceiling Bookshelves; Cube Shelving; Low Credenza \\
\textbf{seating area} & spatial arrangement & The seating area determines which part of the room is designated for seating. & Meeting Table; Lounge Sofa; Reading Armchair; No Extra Seating \\
\textbf{window treatment} & textiles & Window treatment decides the window area. & Sheer Curtains; Linen Side Panels; Venetian Blinds; Blackout Roller Blind; Wooden Shutters \\
\textbf{decor theme} & accessories & The decor theme decides the decorative objects on the walls, shelves, desk, and floor. & Bare Surfaces; Indoor Plants; Abstract Paintings; Whiteboard \& Pinboard; Framed Photographs \\
\end{longtable}

\begin{longtable}{@{}>{\raggedright\arraybackslash}p{2.5cm}>{\raggedright\arraybackslash}p{1.7cm}>{\raggedright\arraybackslash}p{\dimexpr0.5\textwidth-2.1cm-3\tabcolsep\relax}>{\raggedright\arraybackslash}p{\dimexpr0.5\textwidth-2.1cm-3\tabcolsep\relax}@{}}
\caption{Schema of the room type waiting room: 10 attributes, 53 values.}\label{S-tab:schema_waiting_room}\\
\toprule
Attribute & Group & Description in the attribute list & Values \\
\midrule\endfirsthead
\toprule
Attribute & Group & Description in the attribute list & Values \\
\midrule\endhead
\bottomrule\endfoot
\textbf{furniture style} & furniture & The attribute determines the design of the chairs' frames, tables, and reception desk. & Scandinavian; Industrial; Mid-Century Modern; Traditional; Rattan \& Cane; Contemporary \\
\textbf{wall color} & surfaces & The wall color determines the finish of the walls, excluding the section behind the reception desk. & Warm White; Dusty Blue; Terracotta; Deep Green; Charcoal; Blush Pink \\
\textbf{upholstery} & textiles & Upholstery determines the covering of the seating. & Cognac Leather; Emerald Velvet; Oatmeal Linen; Slate Grey; Mustard Yellow; Navy Blue \\
\textbf{seating layout} & spatial arrangement & The seating layout determines the placement and orientation of the room's seating. & Circle Around a Table; Rows Along the Walls; Small Clusters; Bench Rows; L-Shaped Sofa Corner \\
\textbf{flooring type} & surfaces & The attribute determines the floor of the room. & Wood-look Vinyl Planks; Carpet Tiles; Polished Concrete; Terrazzo; Large Ceramic Tiles; Herringbone Wood \\
\textbf{lighting scheme} & lighting & The lighting scheme decides the illumination of the room. & Bright Daylight; Cool Bright Overhead; Large Pendants; Wall Sconces; Hidden Cove Glow \\
\textbf{wall decoration} & accessories & It decides the free part of the back wall. & One Large Painting; Landscape Photo Mural; Geometric Colour Blocks; Living Green Wall; Gallery Cluster \\
\textbf{accessory theme} & accessories & The accessory theme determines the objects placed among the seats and on the side walls. & Magazine Racks; Statement Sculpture; Side Tables; Digital Info Screen; Pottery \& Books Shelf \\
\textbf{acoustic treatment} & surfaces & It decides the wall section behind the reception desk. & Fabric Panels; Wood Slats; Plain Wall; Cork Wall \\
\textbf{window treatment} & textiles & Window treatment decides the window area. & Sheer Curtains; Linen Side Panels; Venetian Blinds; Bamboo Roman Shades; Wooden Shutters \\
\end{longtable}

\end{document}